%% file: main.tex
\documentclass{article}

\usepackage[nonatbib,main,final]{neurips_2026}

\usepackage[utf8]{inputenc}
\usepackage[T1]{fontenc}
\usepackage{hyperref}
\hypersetup{
  pdftitle={FocusBranch: Combinatorial Branch-and-Bound for ℓ0 Neural Network Robustness Verification},
  pdfauthor={Renwei Deng, Yang He, Linyi Li, Yuepeng Wang},
  colorlinks=true,
  linkcolor=red!60!black,
  citecolor=green!40!black,
  urlcolor=blue!60!black,
}
\usepackage{url}
\usepackage{booktabs}
\usepackage{multirow}
\usepackage{amsfonts}
\usepackage{nicefrac}
\usepackage{microtype}
\usepackage{xcolor}
\usepackage{algorithm}
\usepackage[noend]{algpseudocode}
\usepackage{macros}
\usepackage{thm-restate}
\usepackage{graphicx}
\usepackage[sort,numbers]{natbib}
\usepackage{soul}

\title{\textsc{FocusBranch}: Combinatorial Branch-and-Bound for {\boldmath$\ell_0$} Neural Network Robustness Verification}

\author{
  Renwei Deng \quad
  Yang He \quad
  Linyi Li\thanks{Equal supervision.} \quad
  Yuepeng Wang\footnotemark[1] \\
  Simon Fraser University%
  \\
  \texttt{\{rda102, yha244, linyi\_li, yuepeng\}@sfu.ca}
}

\begin{document}

\maketitle

\input{sections/abstract}

\input{sections/intro}
\input{sections/prelim}

\input{sections/verification}
\input{sections/eval}
\input{sections/related}
\input{sections/concl}
\input{sections/ack}

\appendix
\crefalias{section}{appendix}
\crefalias{subsection}{appendix}
\input{appendices/extended-related}
\input{appendices/limitations}
\input{appendices/impacts}
\input{appendices/Formula}
\input{appendices/proofs}
\input{appendices/impl}

\input{appendices/hyperparameters}
\input{appendices/extended-eval}

\end{document}

%% file: sections/abstract.tex
\begin{abstract}
Verifying the local robustness of neural networks under $\ell_0$ perturbations is inherently combinatorial, as it must reason over all sparse subsets of $\epsilon$ perturbed pixels. This paper presents \textsc{FocusBranch}, a combinatorial branch-and-bound framework for $\ell_0$ robustness verification. The key insight is to make the $\ell_0$-ball decomposable and branchable using a node abstraction, which consists of a focus pixel set $\mathcal{S}$ and a level $l$, permitting at most $\epsilon-l$ perturbed pixels outside $\mathcal{S}$.
This abstraction unifies two branching mechanisms: concrete nodes branch through upper shadows to fix additional perturbed locations, while super nodes branch through covering designs to cover many perturbation cases with fewer child nodes. \textsc{FocusBranch} proves node safety by concretizing linear relaxations over node-constrained $\ell_0$-balls and recursively branches on unknown nodes. Rather than relying on a fixed verification plan, it samples nodes across levels to estimate safe rates and costs, then selects a starting level and branching strategy predicted to minimize verification time. Experiments on fully connected and convolutional classifiers for MNIST, Fashion-MNIST, and CIFAR-10 show that \textsc{FocusBranch} verifies $\ell_0$ robustness effectively across architectures and outperforms state-of-the-art verifiers.
The code is available at: \mbox{\url{https://github.com/simpl-group/FocusBranch}}.
\end{abstract}

%% file: sections/intro.tex
\section{Introduction} \label{sec:intro}

Local robustness provides formal guarantees that a neural network's prediction is stable under a class of input perturbations~\cite{lirpa,crown,deeppoly}. For image classification, robustness to sparse perturbations is especially important, because modifying only a few carefully-chosen pixels can cause misclassification~\cite{vargas2019understanding,pomponi2022pixle,sigma0}. Therefore, we study local robustness verification against an \lo-bounded perturbation. In particular, given a neural network, an input image, and a perturbation budget $\epsilon$, we aim to prove that the prediction of the network remains unchanged for every image obtained by modifying at most $\epsilon$ pixels.
Unlike the widely studied $\ell_\infty$-norm setting~\cite{li2023sok,wang2021beta,deeppoly,katz2019marabou,kaulen20256th}, \lo-norm verification presents unique challenges.
The verifier must navigate
through as large as $\binom{V}{\eps}$ sets of perturbed pixels (about $2.4\times10^{12}$ for a $28\times28$ image at $\eps=5$)
while keeping the affine bounds tight enough.

Existing \lo robustness verifiers offer different solutions. On the one hand, auto\_LiRPA~\cite{lirpa} derives a closed-form concretization solution by selecting the top-$\epsilon$ most-contributing pixels, which can be integrated with LiRPA-style relaxation propagation pipeline (DeepPoly~\cite{deeppoly}/CROWN~\cite{crown}) to compute bounds, but the bounds are too loose to verify under large $\eps$.
On the other hand, \calzone~\cite{calzone} and \coverd~\cite{coverd} run branch-and-bound by recursively decomposing the perturbable pixel sets via covering design~\cite{Schonheim1964} and verify subproblems with smaller perturbable sets using a box verifier~\cite{gpupoly}.
A more recent verifier \cvdtt~\cite{shapira2026tight} combines \coverd with auto\_LiRPA's \lo concretization (the top-$\eps$ method) such that those decomposed subproblems are verified over \lo-balls instead of boxes.
Because every subproblem must contain all perturbed pixels in one set, these verifiers are effective when large sets can be verified, so that few sets cover all perturbation cases.

\input{figures/workflow}

We introduce \tool, a combinatorial branch-and-bound framework to verify local robustness of neural networks under \lo perturbations. The key of \tool is the \emph{node} abstraction, consisting of a \emph{focus pixel set} and its \emph{level}, restricting at most how many perturbed pixels outside focus set. This abstraction naturally captures two different kinds of nodes: \emph{super nodes}, where the focus set size is larger than the level, and \emph{concrete nodes}, where the focus set size equals its level.
\tool reduces \lo robustness verification to proving the robustness of certain nodes.
When a node cannot be verified directly by margin lower bounds, we can branch on it to decompose it into several child nodes that are easier to verify.
Specifically, \emph{concrete branching} fixes one additional perturbation location, while \emph{super branching} uses covering designs~\cite{GordonKuperbergPatashnik1995,Schonheim1964,Todorov1985} to refine~\cite{calzone} the focus set and compactly cover many perturbation cases with fewer generated nodes.

As shown in \Cref{fig:workflow}, \tool does not commit in advance to a fixed verification plan. Instead, it samples nodes across levels, estimates their safe rates and costs, and selects the best starting level and super branching strategy with the smallest predicted running time to initialize the branch-and-bound process. In fact, starting at a small level can substantially reduce the number of nodes to be checked: if a smaller-level node is verified safe, all larger-level nodes it covers can be safely pruned. Restricting to start at level $\epsilon$ recovers the decomposition used in prior work~\cite{shapira2026tight}.

We evaluated \tool on 13 neural networks for image classification, using 100 images collected from MNIST~\cite{lecun1998mnist}, Fashion-MNIST~\cite{xiao2017fashionmnist}, and 10 images from CIFAR-10~\cite{krizhevsky2009cifar}. The results show that \tool is effective and efficient for \lo robustness verification.
In particular, at $\epsilon=5$ it verifies 87.2\% of the candidate instances on fully connected networks (75\,s on average) and 85.9\% on convolutional ones (151\,s), versus 42.5\% and 61.4\% for the strongest baseline \cite{shapira2026tight}.

\vspace{-5pt}
\bfpara{Contributions.}
To summarize, we make the following main contributions in this paper:
\vspace{-5pt}
\begin{itemize}[leftmargin=*]
\item We introduce a node-based combinatorial decomposition of the \lo-ball, using upper shadows for concrete nodes and covering designs for super nodes to generate node-constrained robustness verification subproblems.
\item We derive closed-form concretization solutions over node-constrained \lo-balls, allowing linear-relaxation-based approaches to bound over nodes without enumerating perturbation subsets.
\item
We extend the sampling-based planning technique~\cite{calzone,coverd} across \emph{levels}, which estimates safe rates and costs of all-level nodes and selects the best plan predicted to minimize verification time.
\item We present \tool, a branch-and-bound verification framework for \lo robustness verification. Our evaluation on 13 neural networks and 100 images from various datasets shows that \tool outperforms state-of-the-art verifiers.
\end{itemize}

%% file: figures/workflow.tex
\begin{figure}[!t]
\centering
\resizebox{0.9\linewidth}{!}{
\begin{tikzpicture}[
  >=Latex,
  node distance=6mm and 10mm,
  box/.style={draw, rounded corners=2mm, align=center, inner sep=3pt, minimum height=8mm},
  note/.style={draw, dashed, rounded corners=2mm, align=center, inner sep=3pt},
  line/.style={->, thick},
  lab/.style={font=\small, inner sep=0.5pt}
]

\definecolor{planblue}{RGB}{232,240,254}
\definecolor{procgold}{RGB}{255,243,224}
\definecolor{branchgreen}{RGB}{232,245,233}
\definecolor{milpred}{RGB}{252,228,236}
\definecolor{okgreen}{RGB}{232,245,233}

\node[align=left] (in) {$f,~\nomvx,~\epsilon$};

\node[box, fill=planblue, minimum width=26mm, below=4mm of in] (plan)
  {Plan\\sample nodes\\choose $l_s,\gR^{(l_s)}$};

\node[box, fill=procgold, minimum width=24mm, right=25mm of plan] (proc)
  {Dequeue $o$;\\ $\Bound{o}$};

\node[box, fill=branchgreen, minimum width=24mm, below=7mm of proc] (branch)
  {$\branch(o)$\\
super node: covering design\\
concrete node: upper shadow};

\node[box, fill=milpred, minimum width=24mm, right=27mm of proc] (milp)
  {MILP fallback\\(rare)};

\node[below=6mm of milp, align=center] (adv)
  {Adversarial\\Example};

\node[above=4mm of milp] (ver)
  {Verified};

\node[note, left=4mm of branch, yshift=2pt] (node)
  {Node $o:=(\gS,l)$: at most $\eps-l$\\ perturbed pixels outside $\gS$};

\draw[line] (in) -- (plan);
\draw[line] (plan) -- node[lab, above, align=center]
  {init worklist} (proc);

\path (proc) edge [line, loop above, looseness=5]
    node [lab, yshift=1pt] {safe, continue} (proc);

\draw[line] (proc.north east) -- node[lab, above, xshift=-10pt] {worklist empty} (ver.west);

\draw[line] ([xshift=5]proc.south) to[out=330,in=60] node[lab, right,xshift=2pt,align=left] {unknown,\\[-2pt]branchable} ([xshift=5]branch.north);

\draw[line] ([xshift=-5]branch.north) to[out=145,in=240]
  node[lab, left, xshift=-2pt] {enqueue child nodes} ([xshift=-5]proc.south);

\draw[line] ([yshift=-2pt]proc.east) -- node[lab, below, align=center] {unknown,\\[-2pt]unbranchable} ([yshift=-2pt]milp.west);

\draw[line] (milp) -- node[lab, right] {feasible} (adv);
\draw[line, dotted] ([yshift=2pt]milp.west) -- node[lab, above] {infeasible} ([yshift=2pt]proc.east);

\end{tikzpicture}
}
\caption{Overview of \tool. After choosing $l_s$ and $\gR^{(l_s)}$ by sampling, \tool runs branch-and-bound over node-constrained $\ell_0$-balls, branching via covering designs (super node) or upper shadows (concrete node). Unknown, unbranchable nodes fall back to MILP.}
\label{fig:workflow}
\vspace{-15pt}
\end{figure}
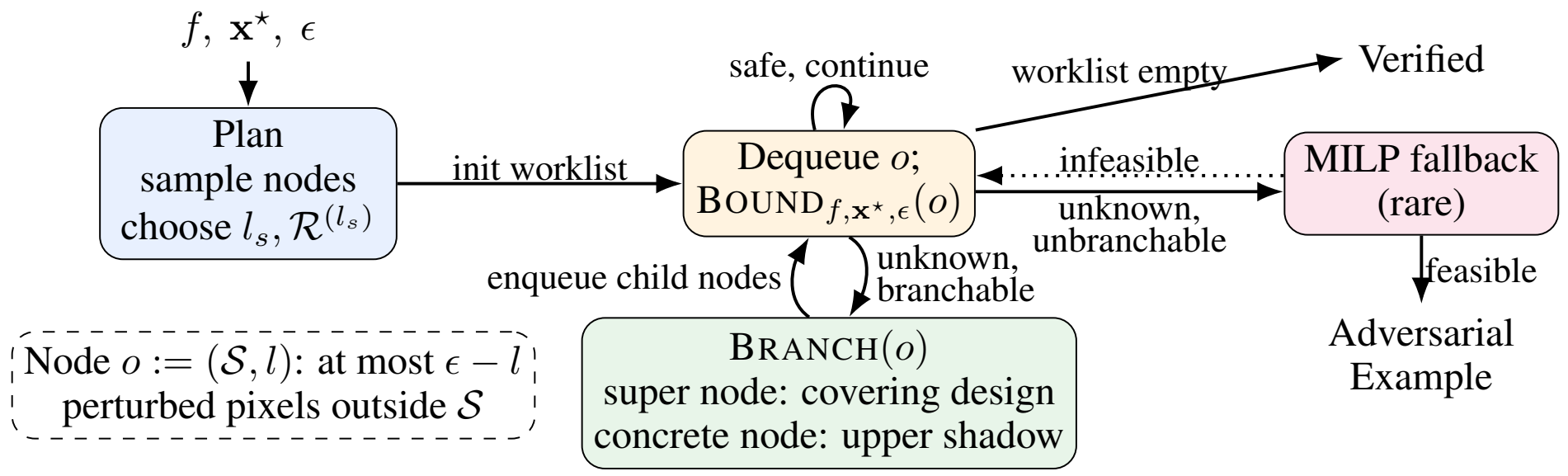

%% file: sections/prelim.tex
\section{Preliminaries} \label{sec:prelim}

\vspace{-5pt}
\paragraph{Neural networks.}
Let $f : \gD \to \sR^N$ be a neural network for image classification that maps an input $\vx \in \gD$ to a vector of class scores.
Here, the domain $\gD$ is a box $[\lvx, \uvx] \subseteq \sR^{VC}$, where $V$ denotes the number of pixels and $C$ the number of channels.
In general, we write $\vx = [\vx_1; \dots; \vx_V]$, where each $\vx_v \in \sR^C$ corresponds to pixel $v$. For example, a normalized $28 \times 28$ grayscale image lies in $[0,1]^{784}$, while a normalized $32 \times 32$ RGB image lies in $[0,1]^{3072}$.
The predicted label of image $\vx$ by network $f$ is defined as $\argmax_j f_j(\vx)$, where $f_j(\vx)$ is the $j$-th entry of $f(\vx)$.

\vspace{-4pt}
\paragraph{Adversarial attacks and perturbation.}
An adversarial attack considers a nominal input $\nomvx \in \gD$ and seeks a perturbation $\vdelta$ such that the perturbed input $\nomvx + \vdelta$ is misclassified, i.e. $\argmax_j f_j(\nomvx) \neq \argmax_j f_j(\nomvx + \vdelta)$, while remaining within $\nomvx + \vdelta \in \gD$.
We consider local robustness against few-pixel attacks~\cite{sparsefool}, where the adversary may modify at most $\epsilon$ pixels. The integer $\epsilon$ is called the perturbation budget and $\gV = \set{1, \dots, V}$ the pixel index set.
The perturbation $\vdelta\in\sR^{VC}$ satisfies the pixel-wise \lo constraint $|\{v \in \gV : \vdelta_v \neq \mathbf{0}\}| \le \epsilon$, where $\vdelta_v \in \sR^C$ is the perturbation on pixel $v$.
The space of admissible perturbed inputs $\eball$ defines a pixel-wise \lo-ball centered at $\nomvx$ with radius $\epsilon$:
\begin{equation} \label{eq:eball}
    \eball := \left\{ \vx\in \gD  \;:\; \; |\{v \in \gV : \vx_v \neq \nomvx_v\}| \le \epsilon\right\}
\end{equation}

\vspace{-8pt}

\paragraph{{\boldmath\lo} local robustness and affine bounds.}\label{sec:affine-bounds}
A network $f$ is said to be \lo \emph{locally robust} w.r.t. $\nomvx$ and $\epsilon$ if it assigns the same label to all inputs over $\eball$, i.e., for all $\vx \in \eball,\; \argmax_j f_j(\vx) = \argmax_j f_j(\nomvx)$.
To verify this property, it suffices to ensure that the predicted class $\nomy := \argmax_j f_j(\nomvx)$ remains dominant over all other classes throughout $\eball$, i.e.,
\begin{equation} \label{eq:local-robust}
    \min_{j\neq\nomy}~\min_{\vx \in \eball} m_{\nomy,j}(\vx) > 0
\end{equation}
where $m_{\nomy,j}:=f_{\nomy} - f_j$ is the margin function.
We adopt Linear Relaxation-based Perturbation Analysis (LiRPA)~\cite{lirpa} to compute the concrete lower bounds of these margins. In particular, LiRPA can be applied directly to $m_{\nomy,j}$, yielding an affine margin lower bound $\lvm(\vx)$ such that for all $\vx \in \eball$, $\lvm(\vx) \le f_{\nomy}(\vx) - f_j(\vx)$.
Thus, a sufficient condition for \Eqref{eq:local-robust} is
\begin{equation} \label{eq:cert-local-robust}
\min_{j\neq\nomy}~\min_{\vx\in\eball}\lvm(\vx) > 0.
\end{equation}
More specifically, for any intermediate affine layer $i$ in the computation graph of $f$, let $h^{(i)}(\vx)$ denote its value. There exist affine lower and upper bounds of $h^{(i)}(\vx)$ w.r.t. input $\vx$, namely $\lhi$ and $\uhi$, such that for all $\vx \in \eball,\;\lhix\le h^{(i)}(\vx)\le \uhix$ holds. They are computed by back-propagating linear relaxations layer by layer until $\vx$~\cite{deeppoly,crown}. Moreover, $\lhi$ and $\uhi$ are then concretized over $\eball$~\cite{lirpa}~(see below) to obtain concrete lower and upper bounds of $h^{(i)}(\vx)$ before over-approximating its immediate nonlinear activation by linear relaxation~\cite{deeppoly,crown}.
Recursively repeating this process forward through every intermediate affine layer until the margins yields $\lvm$. Performing one more concretization gives the concrete lower bound needed to evaluate \Eqref{eq:cert-local-robust}.

\vspace{-4pt}
\paragraph{Concretization over the full {\boldmath\lo}-ball.}\label{sec:concretize-loball}
When a linear relaxation is back-propagated to the input, the resulting affine form $g(\vx)=\omega^\top \vx+b$ must be optimized over $\eball$. Since $\eball$ constrains the number of modified pixels rather than their magnitudes, the minimum (resp.\ maximum) of $g$ is $g(\vx^\star)$ minus (resp.\ plus) the sum of the top-$\epsilon$ pixel contributions, where a pixel's contribution is the largest decrease (resp.\ increase) it can make to $g$ within its range~\cite{chiang2020certified,lirpa}, aggregated over channels~\cite{shapira2026tight}.

\paragraph{Covering-design-based branch-and-bound.}\label{sec:covering-bab}

A covering design $\gC(\gS,k,t)$ is a collection of $k$-element subsets (blocks) of $\gS$ such that every $t$-element subset of $\gS$ is contained in at least one block~\cite{GordonKuperbergPatashnik1995,Schonheim1964,Todorov1985,gordon2023ljcr}.
Since every set of $\binom{V}{\eps}$ lies in some block of $\gC(\gV,k,\eps)$, \calzone~\cite{calzone} verifies $\eball$ by verifying the box
$\gD_B:=\set{\vx\in\gD: \forall v\notin B.\;\vx_v=\nomvx_v}$ of every block $B$ with an incomplete box verifier~\cite{gpupoly}.
Starting from $\gV$, a block $B$ that fails is refined by $\gC(B,k_{|B|},\eps)$ at $\eps\le k_{|B|}<|B|$, recursively, and a failing block of exactly $\eps$ pixels is solved by MILP~\cite{tjeng2019mip}. The refinement size $k_{|B|}$ for each block size $|B|$ is fixed in advance by dynamic programming that minimizes the expected verification time, using safe rates and costs estimated on sampled blocks of each size~\cite{calzone}.
Instead of starting from $\gC(\gV,k_V,\eps)$, \coverd~\cite{coverd} introduces covering verification design (\cvd), whose blocks may differ in size, chosen among candidate \cvd{}s by estimating the cost over the predicted block-size distributions.
\cvdtt\cite{shapira2026tight} inherits \coverd and upgrades the box verifier~\cite{gpupoly} with auto\_LiRPA's \lo-ball concretization~\cite{lirpa} (\Cref{sec:concretize-loball}), such that it verifies each block $B$ over its corresponding \lo-ball $\{\vx\in\eball:\forall v\notin B.\;\vx_v=\nomvx_v\}$, instead of its box~$\mathcal{D}_B$.

%% file: sections/verification.tex
\section{Verification Algorithm}\label{sec:verification}
\vspace{-5pt}
\subsection{Node Abstraction}\label{sec:node-abstraction}

\vspace{-5pt}

Our verification method uses a combinatorial branch-and-bound approach that leverages the \emph{node} abstraction to verify no \lo perturbation on the input image within the given budget $\epsilon$ can change the prediction.
We define a node $o:=(\gS,l)$ as a pair of a \textit{focus} pixel set $\gS\subseteq\gV$ and its level $l \in \{0, \ldots, \epsilon\}$, such that $|\gS|\ge l$.
We use $o.\gS$ and $o.l$ to specify the $\gS$ and $l$ elements of a given node $o$, respectively.
Node $o$ requires no more than $\eps-o.l$ pixels are perturbed outside $o.\gS$.
Formally, the admissible perturbed input space for $\vx\in\eball$ constrained by a node $o$ can be defined as follows:
\begin{equation} \label{eq:eoball}    \eoball:=\set{\vx\in\eball:~|\set{v:~\vx_v\neq\nomvx_v}\setminus o.\gS|\le\epsilon-o.l}
\end{equation}
\begin{figure}[t]
  \centering
  \includegraphics[width=\linewidth]{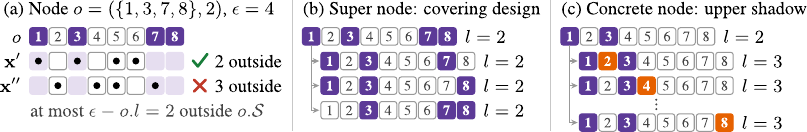}
  \caption{(a) A node $o$ admits at most $\epsilon - o.l$ perturbed pixels (dots) outside $o.\mathcal{S}$ (purple).
    (b) Every 2-subset of $o.\mathcal{S}$ lies in some child. (c) Each child adds one pixel $v \notin o.\mathcal{S}$ (orange).}
  \label{fig:node-branching}
  \vspace{-12pt}
\end{figure}
\vspace{-12pt}

Generally, if $\epsilon$ pixels are perturbed, node $o$ requires that at least $o.l$ perturbed pixels come from the focus set $o.\gS$. Note that perturbing more than $o.l$ pixels in $o.\gS$ is allowed.

\begin{example}
\Cref{fig:node-branching}(a) shows the node $o=(\set{1,3,7,8},2)$ with $V=8$ and $\eps=4$ admitting at most $\eps-o.l=2$ perturbed pixels outside $o.\gS$.
$\vx'\in\eoball$ perturbs $\set{1,3,5,6}$ with only $5$ and $6$ lie outside $o.\gS$, whereas $\vx''\notin\eoball$ perturbs $\set{2,4,5,7}$ with more than two pixels lie outside $o.\gS$.

\end{example}

\vspace{-5pt}
\subsection{Verifying Local Robustness using Branch-and-Bound}
\vspace{-5pt}

In order to verify $f$ is locally robust, we first define its local robustness on one node. Similar to \Eqref{eq:local-robust}, $f$ is said to be \emph{locally robust on $o$} (or simply called $o$ is \emph{safe}), denoted by $\safe(o)$, if $\nomy$ remains dominant over all other classes throughout $\eoball$, i.e.,
\begin{equation} \label{eq:o-local-robust}
\safe(o) := \min_{j\neq\nomy}~ \min_{\vx \in \eoball} m_{\nomy,j}(\vx) > 0
\end{equation}
To prove node $o$ is safe, we can compute the concrete lower bounds of the margins in \Eqref{eq:o-local-robust}
(see \Cref{sec:cert-o}).
We define the process as a function of nodes (\Eqref{eq:bound-o}).
Therefore, a sufficient condition for $o$ being safe is $\Bound{o} > 0$~(\Cref{lem:bound-soundness}).
By contrast, a node is said to be \emph{unknown} if $\Bound{o} \le 0$.
\begin{equation} \label{eq:bound-o}
\Bound{o} := \min_{j\neq\nomy}~\min_{\vx\in\eoball}\lvm(\vx)
\end{equation}
\vspace{-5pt}

We define $\Theta_l$ as the \emph{initial node} for each level $l$, where $\Theta_0 := (\emptyset,0)$ and $\Theta_l := (\gV,l)$ for $l>0$. To prove $f$ is robust over $\eball$, it suffices to prove the safety of any initial node $\Theta_l$ (\Cref{lem:Initial Node Safety}). However, directly proving the safety of $\Theta_l$ may be difficult. To address this issue, we use branch-and-bound to recursively decompose an unknown node $o$ into child nodes using a branching operator $\branch(o)$, which must be sound~(\Cref{cor:sound-branching}): if all nodes in $\branch(o)$ are safe, then $o$ is also safe. This decomposes a difficult node-constrained verification problem into smaller subproblems.

\input{algorithms/Verify}

Following this methodology, we design the verification procedure shown in \Cref{algo:verify}. Given a network $f$, a nominal input $\nomvx$, and a perturbation budget $\epsilon$, the algorithm first
selects a starting level $l_s$ and a super branching strategy $\gR^{(l_s)}$ predicted to minimize overall verification time using sampling statistics (Line 2, as detailed in \Cref{sec:planning}).
The algorithm then proceeds with a worklist procedure.
Specifically, the worklist $\gO$ is initialized with an initial node $\Theta_{l_s}$ (Line 3). For each node $o \in \gO$, the algorithm first checks whether $o$ is safe with $\Bound{o} > 0$. If so, $o$ is pruned and the algorithm proceeds to the
next node. Otherwise, $o$ is unknown and is decomposed according to its type (see \Cref{sec:branching}) and adds the generated nodes into $\gO$ (Lines 7 -- 10). If $o$ is unbranchable (level-$\eps$ concrete, see \Cref{sec:branching}), the corresponding safety verification problem is encoded as a Mixed-Integer Linear Programming (MILP) problem~\cite{tjeng2019mip} in order to search for an adversarial example within the input \emph{box} induced by $\epsilon$-level concrete $o$. If such an example $\vx^\texttt{adv}$ is found, then the network $f$ is unsafe (Lines 12 -- 13).
Otherwise, the algorithm continues until $\gO$ is empty, at which point the safety of $\Theta_{l_s}$ is proven and the robustness of $f$ over $\eball$ is verified~(\Cref{lem:Initial Node Safety}).

\begin{restatable}[Relative soundness]{theorem}{relsoundness}
\label{thm:soundness}
Suppose the MILP solver is sound. Given a network $f$, a nominal input $\nomvx$, and a maximum number of perturbed pixels $\epsilon$, if $\Tool{f, \nomvx, \epsilon}$ returns \texttt{Verified}, then $f$ is locally robust with respect to $\nomvx$ and $\epsilon$.
\end{restatable}

\begin{restatable}[Relative completeness]{theorem}{relcompleteness}
\label{thm:completeness}
Suppose the MILP solver is sound and complete. Given a network $f$, a nominal input $\nomvx$, and a maximum number of perturbed pixels $\epsilon$, if $f$ is locally robust with respect to $\nomvx$ and $\epsilon$, then $\Tool{f, \nomvx, \epsilon}$ returns \texttt{Verified}.
\end{restatable}

\subsection{Branching Policies} \label{sec:branching}
The node abstraction enables a combinatorial decomposition of the \lo-ball beyond explicit perturbable set branching: branching an unknown node $o$ replaces the node-constrained ball $\eoball$ by a collection of child node-constrained balls whose union equals exactly their parent's node-constrained ball (\Cref{cor:branching-exactness}), so proving every child safe proves $o$ safe (\Cref{cor:sound-branching}).
\begin{enumerate}[leftmargin=*]
\item A node $o$ is a \emph{super node} if its focus set size is larger than its level, i.e., $|o.\gS| > o.l$.
Our branching policy for super nodes decomposes $o$ by applying the covering design refinement (\Cref{sec:covering-bab}) to its focus set $o.\gS$:
\begin{equation} \label{eq:superbranch}
\sBranch{k}{o}:= \set{(\gS^\prime,o.l):~ \gS^\prime\in\gC(o.\gS,k,o.l)}
\end{equation}
Here $k$ is chosen by the strategy planner (see \Cref{sec:planning}) that satisfies $o.l\le k<|o.\gS|$. Thus, when $k>o.l$, one generated child super node can cover many concrete nodes at level $o.l$.

\item The node $o$ is a \emph{concrete node}, if the focus set size equals the level, i.e., $|o.\gS| = o.l$.
Our branching policy for concrete nodes decomposes $o$ by the upper shadow of $o.\gS$, i.e., its immediate supersets:
\begin{equation}\label{eq:concretebranch}
\cBranch{o}:=\set{(\gS^\prime,o.l+1):~ \gS^\prime\in\set{o.\gS\cup\set{v}:v\in\gV\setminus o.\gS}}
\end{equation}
Intuitively, each child node fixes one additional perturbed location $v\in\gV\setminus o.\gS$ to the focus set, which increases the level by one. In practice, concrete branching is more efficient, and a pruning technique can be applied afterwards (detailed in \Cref{sec:sap}).
Note that concrete nodes at level $\epsilon$ are \textbf{unbranchable} because the level cannot be increased further. Otherwise, we call $o$ \textbf{branchable} if $|o.\gS|>o.l$ or $|o.\gS|=o.l<\eps$.
\end{enumerate}

\begin{example}

\Cref{fig:node-branching}(b) branches the super node $o=(\set{1,3,7,8},2)$ with $k=3$.
The covering design $\gC(o.\gS,3,2)=\set{\set{1,3,7},\set{1,3,8},\set{3,7,8}}$ contains every 2-subset of $o.\gS$ in some block, so $\sBranch{3}{o}$ yields three children at level~2, and each child covers three concrete nodes, e.g., $(\set{1,3,7},2)$ covers $(\set{1,3},2)$, $(\set{1,7},2)$ and $(\set{3,7},2)$.
\Cref{fig:node-branching}(c) branches the concrete node $o=(\set{1,3},2)$, and $\cBranch{o}$ yields six level-3 children $(\set{1,3,v},3)$, one per pixel $v\in\gV\setminus o.\gS=\set{2,4,5,6,7,8}$.

\end{example}

Each branching step either strictly shrinks the focus set or strictly increases the level and focus set by one, so the recursion is finite and in the worst case terminates at level-$\eps$ concrete nodes (\Cref{lem:branching-progress,lem:finite-propagation}).

\subsection{Bounding a Node} \label{sec:cert-o}

A critical step in \Cref{algo:verify} is to prove the safety of each node $o$ with $\Bound{o}>0$ early, thereby avoiding unnecessary branching (Line 6). The concrete margin bounds in \Eqref{eq:bound-o} are obtained through the same LiRPA{-style} pipeline as in \Cref{sec:affine-bounds}, except that each concretization step, for the intermediate affine layers as well as for the margins, is performed over the node-constrained \lo-ball $\eoball$ rather than the full \lo-ball $\eball$. This change preserves soundness, i.e., $\Bound{o}$ remains a valid concrete lower bound of the margins over $\eoball$ (\Cref{lem:bound-validity}). What remains is to perform this concretization without enumerating the perturbation subsets that $o$ covers.

We now derive a closed-form solution for concretizing any affine function $g(\vx)=\omega^\T\vx+b$ over $\eoball$, starting with $\max_{\vx\in\eoball} g(\vx)$.
Write $\vx=[\vx_1;\ldots;\vx_V]$ and $\omega=[\omega_1;\ldots;\omega_V]$, where $\vx_v,\omega_v\in\sR^C$. Then $g(\vx)=g(\nomvx)+\sum_{v=1}^V \Delta_v$, where $\Delta_v:=\omega_v^\T(\vx_v-\nomvx_v)=\sum_{c=1}^C\omega_{v,c}(\vx_{v,c}-\nomvx_{v,c})$ is the contribution of pixel $v$ that aggregates the contribution of all its channels. Since $g(\nomvx)$ is fixed, maximizing $g(\vx)$ over $\eoball$ reduces to selecting the pixel contributions that can be made largest under the node constraint. For any pixel $v$, its largest possible contribution can be computed by
\begin{equation}\label{eq:pixel-max}
    \uD_v
    :=\max_{\vx_v\in[\lvx_v,\uvx_v]}\Delta_v
    =\sum_{c=1}^C \omega_{v,c}(\vx'_{v,c}-\nomvx_{v,c}),\quad
    \vx'_{v,c}:=
    \begin{cases}
        \uvx_{v,c}, & \omega_{v,c}\ge 0,\\[2pt]
        \lvx_{v,c}, & \omega_{v,c}<0.
    \end{cases}
\end{equation}
That is, each channel is pushed to its upper (resp.\ lower) bound when the coefficient is positive (resp.\ negative) to maximize its contribution before aggregation.

Since $\uD_v\ge0$, the optimal perturbation uses all budgets, i.e., perturbs exactly $\eps$ pixels, and the $o$-constrained ball becomes $\set{\vx\in\eball:~|\set{v:~\vx_v\neq\nomvx_v}\cap o.\gS|\ge o.l}$. By \Cref{lem:node-constrained-concretization}, the maximum of $g(\vx)$ over $\eoball$ admits the closed-form
\begin{equation}\label{eq:maxgx}
    \max_{\vx \in \eoball} g(\vx)
= g(\nomvx) + \max_{\vx\in\eoball}\sum_{v=1}^V\Delta_v
= g(\nomvx) + \sum_{v \in
\overline{\gA}\left(o.l,o.\gS\right)\cup\overline{\gA}\left(\epsilon-o.l,\gV\setminus\overline{\gA}\left(o.l,o.\gS\right)\right)}\uD_v,
\end{equation}
where $\overline{\gA}(a,A):=\arg{\operatorname{top}\text{-}a_{v \in A}\uD_v}$ is a set of $a$ ($a\in\set{0, \ldots, \epsilon}$) most positively contributed pixels from set $A\subseteq\gV$.
Similarly, the closed-form solution for $\min_{\vx\in\eoball}g(\vx)$ can be derived by \Eqref{eq:mingx} in \Cref{sec:Formulas}.

\begin{figure}[t]
  \centering
  \includegraphics[width=\linewidth]{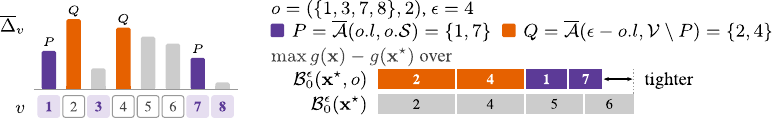}
  \caption{Closed-form node bound with illustrative contributions $\overline{\Delta}_v$. The node bound takes $P \cup Q$,
    whereas the full $\ell_0$-ball takes the top-$\epsilon$ pixels $\{2,4,5,6\}$, so the node bound is tighter.}
  \label{fig:node-bound}
  \vspace{-15pt}
\end{figure}
\vspace{5pt}
\begin{example}

\Cref{fig:node-bound} concretizes $g$ over $\eoball$ for $o=(\set{1,3,7,8},2)$ and $\eps=4$ with illustrative contributions $\uD_v$.
Let $P$ be the top-$o.l$ pixels of the focus set, $P=\overline{\gA}(o.l,o.\gS)=\set{1,7}$, and the remaining selected pixels are the top-$(\eps-o.l)$ of $\gV\setminus P$, i.e., $\overline{\gA}(\eps-o.l,\gV\setminus P)=\set{2,4}$, denoted by $Q$.
By \Eqref{eq:maxgx}, $\max_{\vx\in\eoball}g(\vx)=g(\nomvx)+\uD_1+\uD_2+\uD_4+\uD_7$.
Over the full ball $\eball$, the top-$\eps$ selection picks $\set{2,4,5,6}$ instead, but all four pixels lie outside $o.\gS$, whereas $o$ admits at most two.
Since $\uD_5+\uD_6>\uD_1+\uD_7$, the node bound is tighter.

\end{example}

Since $\eopball\subseteq\eoball$ (\Cref{lem:child-containment}), every concretization step performed for a child node $o'$ is at least as tight as the corresponding step for its parent node $o$. Tighter intermediate bounds in turn yield tighter relaxations downstream, so in practice $\Bound{o'}\ge\Bound{o}$. This is exactly what makes branching productive.

\textbf{Special cases.} At level~0, ~\Eqref{eq:maxgx} and~\Eqref{eq:mingx} reduce to the top-$\epsilon$ selection~(\Cref{sec:concretize-loball}), so \tool restricted to $\Theta_0$ computes auto\_LiRPA's \lo bounds~\cite{lirpa} (but pixel-wise rather than entry-wise). From $\Theta_\epsilon$, a node $(B,\epsilon)$ is exactly the subproblem \cvdtt~\cite{shapira2026tight} solves for block $B$, and super branching is its refinement.

\subsection{Verification Planner} \label{sec:planning}

\input{algorithms/ChooseBestStrategy}

\vspace{-2pt}
A verification plan is
a starting level $l_s \in \set{0, \ldots, \epsilon}$ and a super branching strategy $\Rls:\sN\to\sN$
mapping a focus-set size to its children's focus-set size.
\Cref{algo:plan} first samples nodes of every focus-set size across levels to estimate safe rates $\hmP$ and mean bounding times (the costs) $\hmT$, indexed by $(|o.\gS|,o.l)$ for a given node $o$ (Line~3).
For each level $l$ (Lines 4 -- 8), dynamic programming~\cite{calzone} over $\hmP_{:,l}$ and $\hmT_{:,l}$ gives $\gR^{(l)}$ and its estimated time $t_{est}^{(l)}$.
Then the total estimated time $t_{est}$ starting at level $l$ is obtained by adding up $t_{est}^{(l)}$ and $t_{tail}$, where $t_{tail}$ is the sum of estimated per-level bounding cost $\tvt_i$ (detailed in~\Cref{sec:plan-detail}) from level $l+1$ to $\epsilon$ (Line 6). The best estimation keeps updating in this way until the loop terminates.

%% file: algorithms/Verify.tex
\begin{figure}[!t]
\begin{algorithm}[H]
\caption{Verifying \lo local robustness for network $f$ over $\eball$ using \tool}
\label{algo:verify}
\begin{algorithmic}[1]
\Procedure{\tool}{$f,\nomvx,\epsilon$}
\Statex \textbf{Input:} A classifier $f$, a nominal input image $\nomvx$, and a perturbation budget $\epsilon$
\Statex \textbf{Output:} \texttt{Verified} if the verification is successful, an adversarial example $\vx^\texttt{adv}$ otherwise
\State $(l_s,\gR^{(l_s)}) \gets$ \Call{ChooseBestPlan}{$f,\nomvx,\epsilon$}
\State $\gO\gets\set{\Theta_{l_s}}$ \Comment{Enqueue an initial node at level $l_s$}
\While{$\gO \ne \emptyset$}
    \State $o \gets\gO.\texttt{dequeue}()$
    \If{$\Bound{o}\le0$}
        \If{$|o.\gS| > o.l$} \Comment{Branch on unknown super node}
            \State $\gO\gets\gO\cup\sBranch{\gR^{(l_s)}[|o.\gS|]}{o}$
        \ElsIf{$o.l < \epsilon$} \Comment{Branch on unknown concrete node}
            \State $\gO\gets\gO\cup\cBranch{o}$
        \Else \Comment{Use MILP to verify unbranchable unknown node}
            \State $(\texttt{feasible},\vx^{\texttt{adv}})\gets \texttt{MILP}\left(f, \eoball\right)$
            \If{$\texttt{feasible}$} \Return $\vx^{\texttt{adv}}$ \EndIf
        \EndIf
    \EndIf
\EndWhile
\State \Return \texttt{Verified}
\EndProcedure
\end{algorithmic}
\end{algorithm}
\vspace{-20pt}
\end{figure}

%% file: algorithms/ChooseBestStrategy.tex
\begin{figure}[!t]
\begin{algorithm}[H]
\caption{Choosing Best Verification Plan}
\label{algo:plan}
\begin{algorithmic}[1]
\Procedure{\textsc{ChooseBestPlan}}{$f,\nomvx,\epsilon$}
\Statex \textbf{Input:} A classifier $f$, a nominal input image $\nomvx$, and a perturbation budget $\epsilon$
\Statex \textbf{Output:} The level $l_s$ to start verification, and a super branching strategy $\gR^{(l_s)}$ at level $l_s$
\State $(l_s,\gR^{(l_s)}) \gets (0,\emptyset)$;~ $t_{min}\gets\infty$
\State $(\hmP,\hmT)\gets\texttt{SampleNodesOfAllShape}(f,\nomvx,\epsilon)$
\ForAll{$l\in\set{0, \ldots, \epsilon}$}
    \State $(\gR^{(l)},t_{est}^{(l)})\gets\texttt{GetSuperBranchingStrategyAndEstimate}(\hmP_{:,l}, \hmT_{:,l},l)$
    \State $t_{tail} \gets \sum_{i=l+1}^{\epsilon}{\tvt_i}$;~ $t_{est} \gets t_{est}^{(l)} + t_{tail}$
    \If{$t_{est} < t_{min}$}
        \State $t_{min} \gets t_{est}$;~ $(l_s,\gR^{(l_s)}) \gets (l,\gR^{(l)})$
    \EndIf
\EndFor
\State \Return{$(l_s, \gR^{(l_s)})$}
\EndProcedure
\end{algorithmic}
\end{algorithm}
\vspace{-22pt}
\end{figure}

%% file: sections/eval.tex
\section{Evaluation} \label{sec:eval}

\vspace{-5pt}
\subsection{Experimental Setup} \label{sec:setup}

\vspace{-2pt}
\tool is implemented using PyTorch~\cite{paszke2019pytorch} and CUDA~\cite{nvidia2023cuda122}. Its super branching strategy is partially adapted from the refinement strategy of prior work~\cite{coverd, calzone}; its relaxation techniques and backward affine bounds propagation are inspired by DeepPoly~\cite{deeppoly} and CROWN~\cite{crown}; and its MILP formulation comes from Tjeng et al.~\cite{tjeng2019mip}, using Gurobi~\cite{gurobi} as the solver. The experiments were conducted on a server with AMD EPYC 9454 (Zen 4) @ 2.75 GHz 48-core CPU and four NVIDIA H100 SXM (80 GB memory) GPUs.

We consider three standard image classification datasets with 10 classes each: MNIST~\cite{lecun1998mnist}, which contains 28$\times$28 grayscale handwritten digits; Fashion-MNIST~\cite{xiao2017fashionmnist}, a drop-in MNIST replacement with 28$\times$28 grayscale clothing images;
and CIFAR-10~\cite{krizhevsky2009cifar}, which consists of 32$\times$32 RGB natural images across objects and animals and is generally more challenging due to higher visual variability.

We evaluate \tool on 13 networks, where 8 are adopted from \coverd~\cite{coverd}: one fully connected MNIST network (bfc1), and seven convolutional neural networks (CNNs) (bcv1--bcv7), consisting of four MNIST networks (bcv1--bcv4), two Fashion-MNIST networks (bcv6 and bcv7), and one CIFAR-10 network (bcv5, originally from ERAN~\cite{eran}). The remaining five are fully connected MNIST networks (afc1--afc5) adversarially trained under the \lo setting following the auto\_LiRPA certified defense configuration~\cite{autolirpa_examples}. The network \texttt{mlp\_3layer} (afc4) is obtained by running the official auto\_LiRPA training script~\cite{autolirpa_simple_training}; the other four are derived from pretrained networks. Specifically, afc5 is adapted from bfc1, and afc1--afc3 are adapted from \texttt{mnist-net\_256x2}, \texttt{mnist-net\_256x4}, and \texttt{mnist-net\_256x6}, respectively, from the VNN-COMP 2020 OVAL benchmark~\cite{vnncomp2020mnistoval}.

Before verification, we first run adversarial example search on the MNIST and Fashion-MNIST test sets using \sigmao~\cite{sigma0}, which is
effective at finding $\ell_0$-norm adversarial examples. This preprocessing step reduces the number of instances that can potentially be verified. For all networks except the CIFAR-10 network bcv5, we record the minimum pixel-wise $\ell_0$ budget $\epsilon_{\mathrm{known}}$ at which an adversarial example is known for each image. For bcv5, we do not use \sigmao because \sigmao defines the $\ell_0$ ball over individual image entries for multi-channel inputs, whereas our verification target is pixel-wise $\ell_0$ robustness. Instead, we use the results for adversarial example reported in prior work~\cite{shapira2026tight}.
Since we prioritize verification rather than adversarial example discovery, we skip verification for an image whenever $\epsilon_{\mathrm{target}} \ge \epsilon_{\mathrm{known}}$, where $\epsilon_{\mathrm{target}}$ is the target perturbation budget. \Cref{app:qualitative} shows qualitative examples of verified and adversarial images.

\subsection{Main Results}

\begin{table}[t]
\centering
\caption{Main evaluation results. \textbf{Neurons} is the number of neurons in the network, \textbf{Img} is the number of evaluated images, and \textbf{CC} is the number of correctly classified images. \textbf{Cand.} denotes the number of verification candidate instances at the target budget $\epsilon$, after excluding images for which an adversarial example within the budget is already known. \textbf{R} is the number of verified images, \textbf{R/Cand} is the verified rate, and \textbf{Time} reports the mean verification time (in seconds) over each tool's verified instances. Refer to \Cref{fig:comp,fig:comp-e3,fig:comp-e4} for sample-wise comparison.}
\vspace{5pt}
\label{tab:main}
\input{figures/mainresults}
\vspace{-10pt}
\end{table}

We evaluate all networks at $\epsilon\in\{3,4,5,6\}$ except bcv4 (by far the largest) and bcv5 (CIFAR-10, where 3 pixels span 9 input features), because their bounds are loose enough that verification relies heavily on MILP even at $\epsilon=3$. So, as in prior works~\cite{calzone,coverd,shapira2026tight}, we evaluate them only at $\epsilon=3$ (see \Cref{sec:benchmark-details}). All tools use a timeout of 1800\,s
(3600\,s for bcv4), shorter than the 5\,h per instance used by~\cite{coverd}.
For bcv5, our machine cannot run \cvdtt with 50 MILP instances, so we obtained from their results table~\cite{shapira2026tight} and manually timed out those verified but above 1800s. The mean time is not calculated since we run on different machines.

The results are summarized in \Cref{tab:main}. Overall, \tool verifies 2799 out of 3189 candidate instances, achieving an aggregate verification rate of $87.8\%$. The method is particularly effective for moderate perturbation
budgets: at $\epsilon=3$ and $\epsilon=4$, \tool verifies $99.8\%$ and $98.8\%$ of the candidate instances, respectively. As expected, the verification problem becomes increasingly challenging as $\epsilon$ grows. Nevertheless, \tool still verifies $86.6\%$ of the instances at $\epsilon=5$ and verifies a non-trivial fraction of instances at $\epsilon=6$.

\subsection{Comparison with Baseline Tools}\label{sec:comparison}

Next, we compare \tool with state-of-the-art \lo robustness verifiers.
Since \coverd~\cite{coverd} outperforms \calzone~\cite{calzone}, we only compare \tool with \coverd and a more recent improvement \cvdtt~\cite{shapira2026tight}. The latter can be viewed as a special case of \tool that always starts from the initial node $\Theta_\epsilon$ at the largest level $\eps$, whereas \tool selects the starting level per instance from node-sampling statistics.
Note that \tool enables batching, an additional axis of parallelism, while \cvdtt cannot. \Cref{app:parallelization-ablation} further studies the impact of batching on all the networks.

\begin{figure}
    \centering
    \def\svgwidth{0.32\textwidth}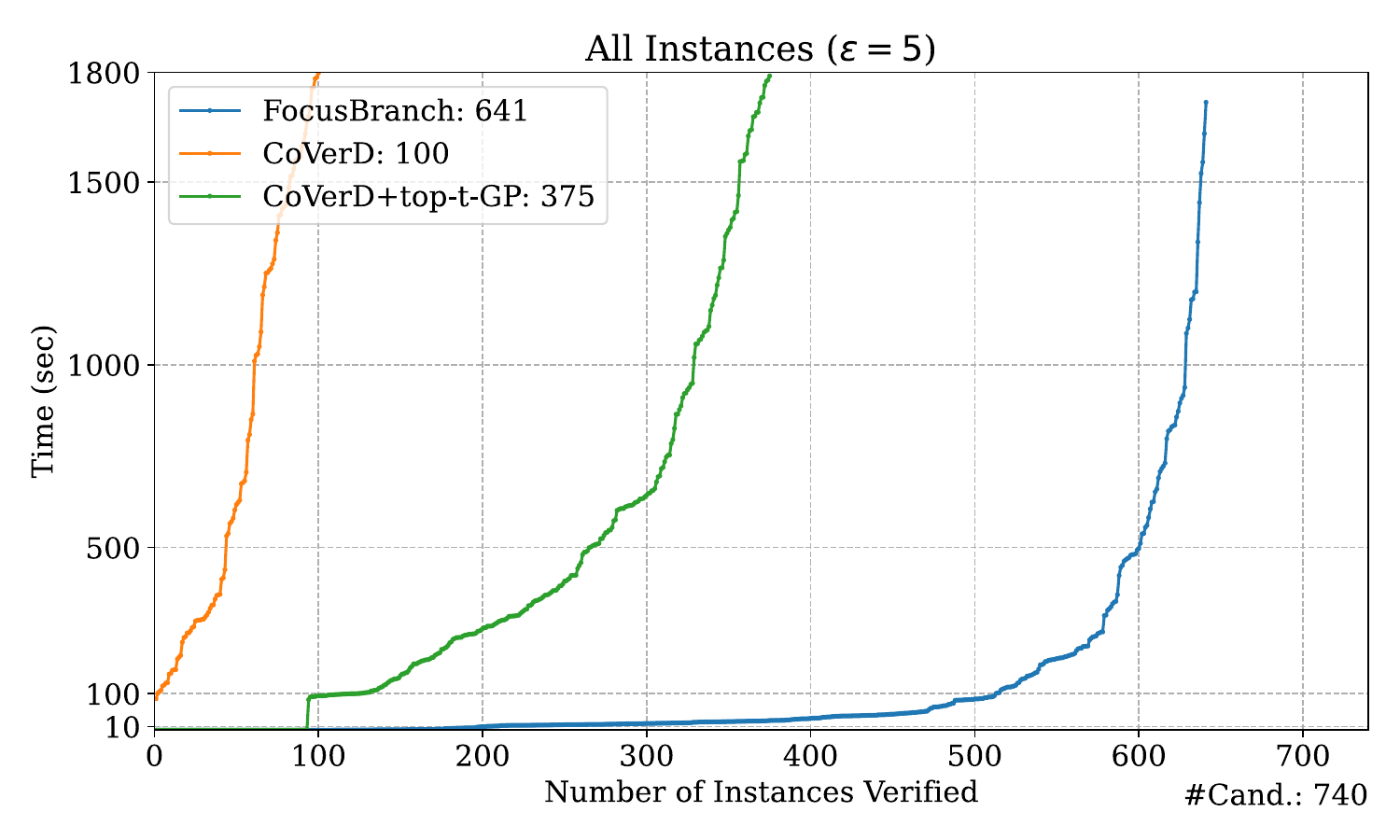
    \def\svgwidth{0.32\textwidth}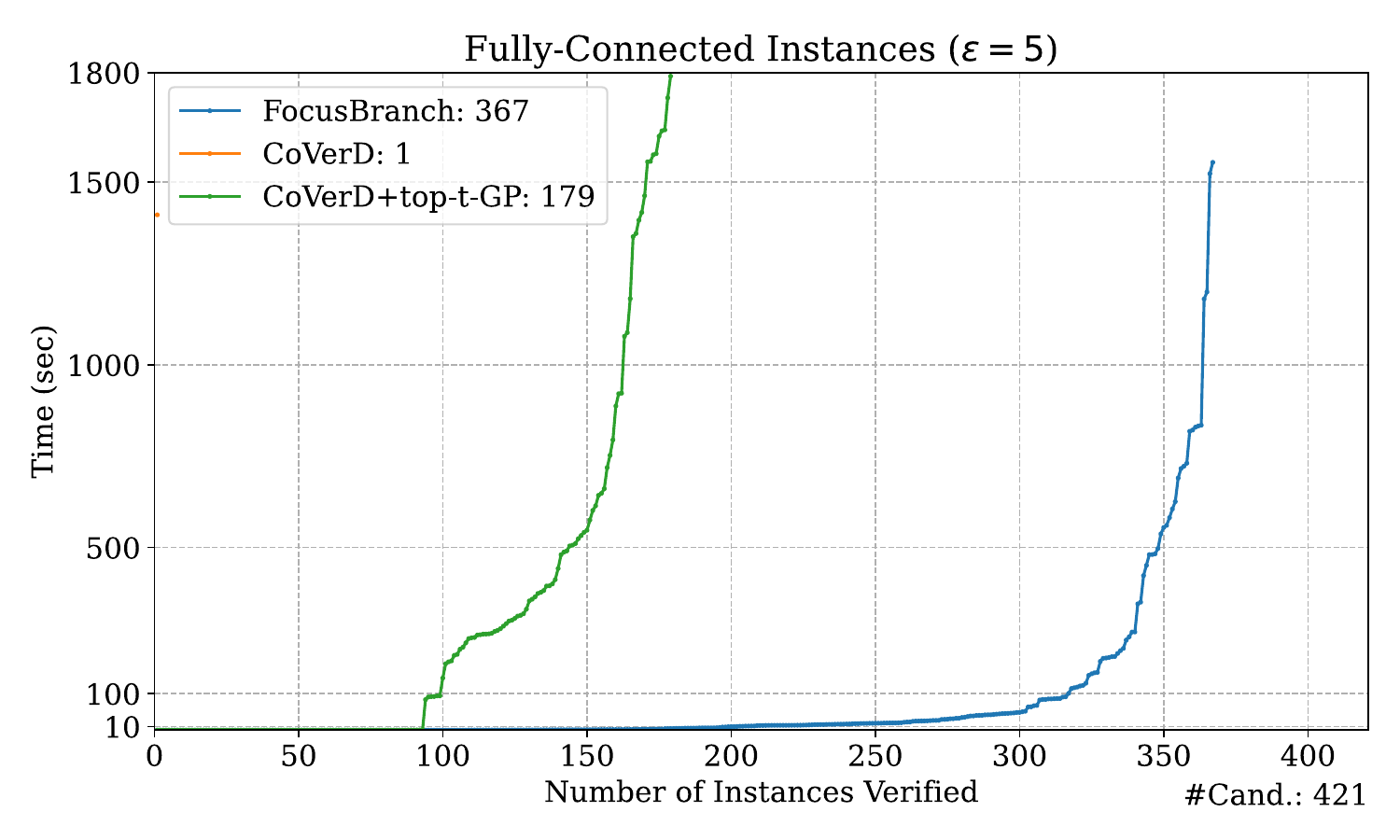
    \def\svgwidth{0.32\textwidth}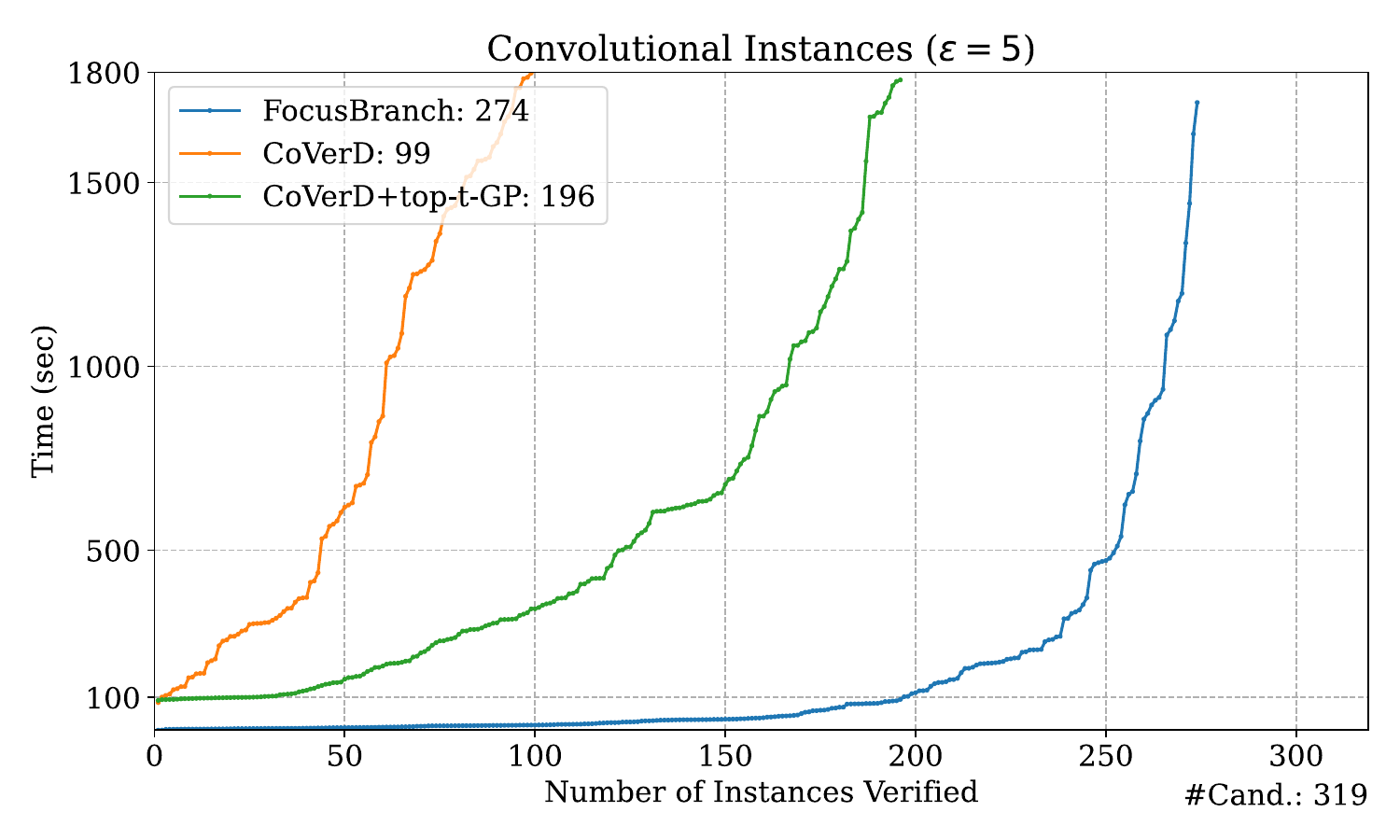
    \vspace{-5pt}
    \caption{Comparison between \tool, \coverd and \cvdtt at $\epsilon=5$.}
    \label{fig:comp}
    \vspace{-12pt}
\end{figure}

We evaluate all methods on the candidate instances at $\epsilon=5$, excluding bcv4 and bcv5,
since they are only evaluated at $\epsilon=3$ in the main results.
As shown in \Cref{fig:comp}, \tool significantly outperforms both baselines.
Across all instances, \tool verifies 641 out of 740 instances, compared with 100 by \coverd and 375 by \cvdtt.
For fully connected networks, \tool verifies 367 instances, while \coverd verifies only one and \cvdtt verifies 179.
On CNNs, \tool also achieves the best result, verifying 274 instances compared with 99 by \coverd and 196 by \cvdtt.
As \Cref{tab:extended-main} shows, the planner selects a starting level less than $\eps$ more often on fully connected networks, so they benefit from both branching policies. On CNNs, the sparse affine representation of convolutional operator (\Cref{sec:benchmark-details}) prevents this, but CNNs still benefit from batching.

The comparison with MILP-based verification is deferred to \Cref{sec:extended-eval-milp}.

\subsection{Ablation Study}\label{sec:abl}
\begin{figure}
    \centering
    \def\svgwidth{0.6\textwidth}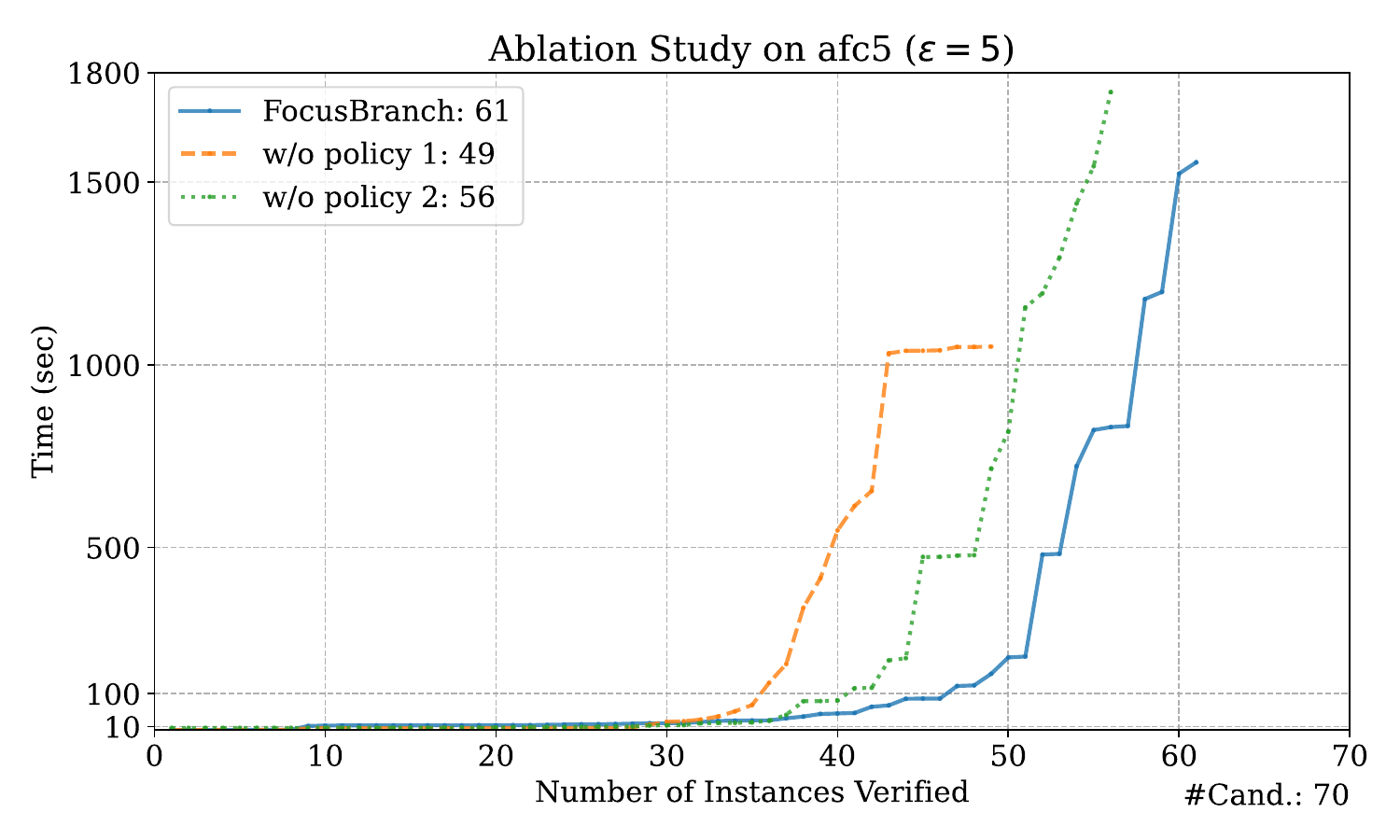
    \vspace{-7pt}
    \caption{Ablation study on afc5 at $\epsilon=5$. ``w/o policy 1'' disables super-node branching by always starting from $\Theta_0$, while ``w/o policy 2'' disables concrete-node branching by always starting from $\Theta_\epsilon$.}
    \label{fig:abl-main}
    \vspace{-10pt}
\end{figure}

To evaluate the contribution of super-node and concrete-node branching policies, we conduct an ablation study on afc5 at $\epsilon=5$, where the selected starting levels are relatively balanced between $l_s<\epsilon$ and $l_s=\epsilon$ (\Cref{tab:extended-main}).
Specifically, we consider two variants for the study. First, \emph{w/o policy 1} disables super-node branching by always starting verification from the initial node $\Theta_0$ and skipping the planning. Since $\Theta_0$ is a concrete node and concrete branching only generates concrete children~(\Cref{lem:concrete-propagation}), this variant performs only concrete branching throughout verification. Second, \emph{w/o policy 2} disables concrete-node branching by always starting from $\Theta_\epsilon$. This variant performs only super-node branching until nodes become concrete; since such concrete nodes are already at level $\epsilon$, they are unbranchable and cannot be further decomposed.

As shown in \Cref{fig:abl-main}, the full \tool verifies the most instances, solving 61 out of 70 instances compared with 49 for w/o policy 1 and 56 for w/o policy 2. The gap shows that neither branching policy alone is sufficient: concrete-node branching is effective for refining difficult concrete nodes, while super-node branching compactly covers many perturbation cases. By adaptively selecting the starting level and branching strategy, \tool combines both mechanisms and achieves the best overall verification performance.

%% file: figures/mainresults.tex
\resizebox{0.8\linewidth}{!}{
\begin{tabular}{ccccccccccccc}
\toprule
\multirow{2}{*}{\textbf{Dataset}} &
\multirow{2}{*}{\textbf{Network}} &
\multirow{2}{*}{\textbf{Neurons}} &
\multirow{2}{*}{\textbf{Img}} &
\multirow{2}{*}{\textbf{CC}} &
\multirow{2}{*}{{\bf $\epsilon$}} &
\multirow{2}{*}{\textbf{Cand.}} &
\multicolumn{3}{c}{\textbf{\tool}} &
\multicolumn{3}{c}{\textbf{\cvdtt}} \\
\cmidrule(lr){8-10}\cmidrule(lr){11-13}
& & & & & & & \textbf{R}$\uparrow$ & \textbf{R/Cand.}$\uparrow$ & \textbf{Time (s)}$\downarrow$ & \textbf{R}$\uparrow$ & \textbf{R/Cand.}$\uparrow$ & \textbf{Time (s)}$\downarrow$ \\
\midrule
\multirow{37}{*}{MNIST} & \multirow{4}{*}{afc1} & \multirow{4}{*}{512} & \multirow{4}{*}{100} & \multirow{4}{*}{100} & 3 & 90 & 90 & 100.0\% & 0.3 & 90 & 100.0\% & 26.6 \\
 &  &  &  &  & 4 & 72 & 70 & 97.2\% & 10.1 & 67 & 93.1\% & 165.7 \\
 &  &  &  &  & 5 & 57 & 52 & 91.2\% & 17.8 & 30 & 52.6\% & 203.3 \\
 &  &  &  &  & 6 & 46 & 29 & 63.0\% & 53.7 & 13 & 28.3\% & 257.1 \\
\cmidrule(lr){2-13}
 & \multirow{4}{*}{afc2} & \multirow{4}{*}{1024} & \multirow{4}{*}{100} & \multirow{4}{*}{100} & 3 & 94 & 94 & 100.0\% & 0.3 & 94 & 100.0\% & 37.2 \\
 &  &  &  &  & 4 & 88 & 87 & 98.9\% & 6.1 & 80 & 90.9\% & 165.0 \\
 &  &  &  &  & 5 & 79 & 70 & 88.6\% & 48.0 & 33 & 41.8\% & 126.9 \\
 &  &  &  &  & 6 & 68 & 38 & 55.9\% & 90.9 & 15 & 22.1\% & 235.8 \\
\cmidrule(lr){2-13}
 & \multirow{4}{*}{afc3} & \multirow{4}{*}{1536} & \multirow{4}{*}{100} & \multirow{4}{*}{100} & 3 & 92 & 92 & 100.0\% & 0.5 & 92 & 100.0\% & 57.3 \\
 &  &  &  &  & 4 & 84 & 84 & 100.0\% & 12.9 & 76 & 90.5\% & 191.9 \\
 &  &  &  &  & 5 & 74 & 67 & 90.5\% & 34.9 & 39 & 52.7\% & 308.5 \\
 &  &  &  &  & 6 & 66 & 42 & 63.6\% & 193.1 & 15 & 22.7\% & 253.7 \\
\cmidrule(lr){2-13}
 & \multirow{4}{*}{afc4} & \multirow{4}{*}{384} & \multirow{4}{*}{100} & \multirow{4}{*}{99} & 3 & 89 & 89 & 100.0\% & <0.1 & 89 & 100.0\% & 9.5 \\
 &  &  &  &  & 4 & 85 & 85 & 100.0\% & 0.5 & 78 & 91.8\% & 109.2 \\
 &  &  &  &  & 5 & 73 & 70 & 95.9\% & 5.4 & 37 & 50.7\% & 90.5 \\
 &  &  &  &  & 6 & 61 & 49 & 80.3\% & 53.5 & 19 & 31.1\% & 33.1 \\
\cmidrule(lr){2-13}
 & \multirow{4}{*}{afc5} & \multirow{4}{*}{1000} & \multirow{4}{*}{100} & \multirow{4}{*}{100} & 3 & 89 & 89 & 100.0\% & 0.9 & 89 & 100.0\% & 109.1 \\
 &  &  &  &  & 4 & 80 & 80 & 100.0\% & 30.3 & 73 & 91.2\% & 321.3 \\
 &  &  &  &  & 5 & 70 & 61 & 87.1\% & 188.4 & 31 & 44.3\% & 638.7 \\
 &  &  &  &  & 6 & 60 & 29 & 48.3\% & 374.8 & 6 & 10.0\% & 741.2 \\
\cmidrule(lr){2-13}
 & \multirow{4}{*}{bfc1} & \multirow{4}{*}{1000} & \multirow{4}{*}{100} & \multirow{4}{*}{99} & 3 & 91 & 91 & 100.0\% & 4.7 & 91 & 100.0\% & 232.2 \\
 &  &  &  &  & 4 & 82 & 81 & 98.8\% & 82.3 & 67 & 81.7\% & 554.5 \\
 &  &  &  &  & 5 & 68 & 47 & 69.1\% & 192.4 & 9 & 13.2\% & 661.0 \\
 &  &  &  &  & 6 & 59 & 13 & 22.0\% & 401.1 & 2 & 3.4\% & 642.5 \\
\cmidrule(lr){2-13}
 & \multirow{4}{*}{bcv1} & \multirow{4}{*}{3604} & \multirow{4}{*}{100} & \multirow{4}{*}{98} & 3 & 87 & 87 & 100.0\% & 7.5 & 87 & 100.0\% & 60.0 \\
 &  &  &  &  & 4 & 81 & 81 & 100.0\% & 13.6 & 80 & 98.8\% & 143.5 \\
 &  &  &  &  & 5 & 73 & 69 & 94.5\% & 133.8 & 52 & 71.2\% & 514.0 \\
 &  &  &  &  & 6 & 64 & 49 & 76.6\% & 359.1 & 14 & 21.9\% & 644.1 \\
\cmidrule(lr){2-13}
 & \multirow{4}{*}{bcv2} & \multirow{4}{*}{3604} & \multirow{4}{*}{100} & \multirow{4}{*}{94} & 3 & 84 & 84 & 100.0\% & 8.7 & 84 & 100.0\% & 67.3 \\
 &  &  &  &  & 4 & 74 & 74 & 100.0\% & 19.6 & 74 & 100.0\% & 90.2 \\
 &  &  &  &  & 5 & 67 & 67 & 100.0\% & 49.4 & 64 & 95.5\% & 269.1 \\
 &  &  &  &  & 6 & 62 & 55 & 88.7\% & 131.0 & 36 & 58.1\% & 488.8 \\
\cmidrule(lr){2-13}
 & \multirow{4}{*}{bcv3} & \multirow{4}{*}{4504} & \multirow{4}{*}{100} & \multirow{4}{*}{99} & 3 & 90 & 90 & 100.0\% & 12.6 & 90 & 100.0\% & 108.0 \\
 &  &  &  &  & 4 & 85 & 85 & 100.0\% & 25.2 & 84 & 98.8\% & 238.2 \\
 &  &  &  &  & 5 & 76 & 72 & 94.7\% & 185.3 & 46 & 60.5\% & 676.6 \\
 &  &  &  &  & 6 & 68 & 41 & 60.3\% & 401.9 & 9 & 13.2\% & 1070.2 \\
\cmidrule(lr){2-13}
 & \multirow{1}{*}{bcv4} & \multirow{1}{*}{34688} & \multirow{1}{*}{10} & \multirow{1}{*}{9} & 3 & 9 & 8 & 88.9\% & 793.3 & 7 & 77.8\% & 1090.5 \\
\midrule
\multirow{8}{*}{F-MNIST} & \multirow{4}{*}{bcv6} & \multirow{4}{*}{3604} & \multirow{4}{*}{100} & \multirow{4}{*}{83} & 3 & 66 & 66 & 100.0\% & 10.0 & 66 & 100.0\% & 61.2 \\
 &  &  &  &  & 4 & 58 & 58 & 100.0\% & 57.8 & 57 & 98.3\% & 197.0 \\
 &  &  &  &  & 5 & 49 & 34 & 69.4\% & 169.3 & 28 & 57.1\% & 621.1 \\
 &  &  &  &  & 6 & 45 & 18 & 40.0\% & 329.1 & 2 & 4.4\% & 328.5 \\
\cmidrule(lr){2-13}
 & \multirow{4}{*}{bcv7} & \multirow{4}{*}{4504} & \multirow{4}{*}{100} & \multirow{4}{*}{86} & 3 & 68 & 68 & 100.0\% & 16.6 & 68 & 100.0\% & 105.6 \\
 &  &  &  &  & 4 & 60 & 54 & 90.0\% & 134.7 & 46 & 76.7\% & 474.5 \\
 &  &  &  &  & 5 & 54 & 32 & 59.3\% & 304.8 & 6 & 11.1\% & 846.4 \\
 &  &  &  &  & 6 & 47 & 4 & 8.5\% & 602.3 & 0 & 0.0\% & -- \\
\midrule
\multirow{1}{*}{CIFAR-10} & \multirow{1}{*}{bcv5} & \multirow{1}{*}{4852} & \multirow{1}{*}{10} & \multirow{1}{*}{9} & 3 & 5 & 4 & 80.0\% & 302.0 & 3 & 60.0\% & -- \\
\bottomrule
\end{tabular}
}

%% file: svg-inkscape/All_e5_svg-tex.pdf_tex
\begingroup%
  \makeatletter%
  \providecommand\color[2][]{%
    \errmessage{(Inkscape) Color is used for the text in Inkscape, but the package 'color.sty' is not loaded}%
    \renewcommand\color[2][]{}%
  }%
  \providecommand\transparent[1]{%
    \errmessage{(Inkscape) Transparency is used (non-zero) for the text in Inkscape, but the package 'transparent.sty' is not loaded}%
    \renewcommand\transparent[1]{}%
  }%
  \providecommand\rotatebox[2]{#2}%
  \newcommand*\fsize{\dimexpr\f@size pt\relax}%
  \newcommand*\lineheight[1]{\fontsize{\fsize}{#1\fsize}\selectfont}%
  \ifx\svgwidth\undefined%
    \setlength{\unitlength}{720bp}%
    \ifx\svgscale\undefined%
      \relax%
    \else%
      \setlength{\unitlength}{\unitlength * \real{\svgscale}}%
    \fi%
  \else%
    \setlength{\unitlength}{\svgwidth}%
  \fi%
  \global\let\svgwidth\undefined%
  \global\let\svgscale\undefined%
  \makeatother%
  \begin{picture}(1,0.6)%
    \lineheight{1}%
    \setlength\tabcolsep{0pt}%
    \put(0,0){\includegraphics[width=\unitlength,page=1]{svg-inkscape/All_e5_svg-tex.pdf}}%
  \end{picture}%
\endgroup%

%% file: svg-inkscape/Fully-Connected_e5_svg-tex.pdf_tex
\begingroup%
  \makeatletter%
  \providecommand\color[2][]{%
    \errmessage{(Inkscape) Color is used for the text in Inkscape, but the package 'color.sty' is not loaded}%
    \renewcommand\color[2][]{}%
  }%
  \providecommand\transparent[1]{%
    \errmessage{(Inkscape) Transparency is used (non-zero) for the text in Inkscape, but the package 'transparent.sty' is not loaded}%
    \renewcommand\transparent[1]{}%
  }%
  \providecommand\rotatebox[2]{#2}%
  \newcommand*\fsize{\dimexpr\f@size pt\relax}%
  \newcommand*\lineheight[1]{\fontsize{\fsize}{#1\fsize}\selectfont}%
  \ifx\svgwidth\undefined%
    \setlength{\unitlength}{720bp}%
    \ifx\svgscale\undefined%
      \relax%
    \else%
      \setlength{\unitlength}{\unitlength * \real{\svgscale}}%
    \fi%
  \else%
    \setlength{\unitlength}{\svgwidth}%
  \fi%
  \global\let\svgwidth\undefined%
  \global\let\svgscale\undefined%
  \makeatother%
  \begin{picture}(1,0.6)%
    \lineheight{1}%
    \setlength\tabcolsep{0pt}%
    \put(0,0){\includegraphics[width=\unitlength,page=1]{svg-inkscape/Fully-Connected_e5_svg-tex.pdf}}%
  \end{picture}%
\endgroup%

%% file: svg-inkscape/Convolutional_e5_svg-tex.pdf_tex
\begingroup%
  \makeatletter%
  \providecommand\color[2][]{%
    \errmessage{(Inkscape) Color is used for the text in Inkscape, but the package 'color.sty' is not loaded}%
    \renewcommand\color[2][]{}%
  }%
  \providecommand\transparent[1]{%
    \errmessage{(Inkscape) Transparency is used (non-zero) for the text in Inkscape, but the package 'transparent.sty' is not loaded}%
    \renewcommand\transparent[1]{}%
  }%
  \providecommand\rotatebox[2]{#2}%
  \newcommand*\fsize{\dimexpr\f@size pt\relax}%
  \newcommand*\lineheight[1]{\fontsize{\fsize}{#1\fsize}\selectfont}%
  \ifx\svgwidth\undefined%
    \setlength{\unitlength}{720bp}%
    \ifx\svgscale\undefined%
      \relax%
    \else%
      \setlength{\unitlength}{\unitlength * \real{\svgscale}}%
    \fi%
  \else%
    \setlength{\unitlength}{\svgwidth}%
  \fi%
  \global\let\svgwidth\undefined%
  \global\let\svgscale\undefined%
  \makeatother%
  \begin{picture}(1,0.6)%
    \lineheight{1}%
    \setlength\tabcolsep{0pt}%
    \put(0,0){\includegraphics[width=\unitlength,page=1]{svg-inkscape/Convolutional_e5_svg-tex.pdf}}%
  \end{picture}%
\endgroup%

%% file: svg-inkscape/Ablation_bafc1_e5_svg-tex.pdf_tex
\begingroup%
  \makeatletter%
  \providecommand\color[2][]{%
    \errmessage{(Inkscape) Color is used for the text in Inkscape, but the package 'color.sty' is not loaded}%
    \renewcommand\color[2][]{}%
  }%
  \providecommand\transparent[1]{%
    \errmessage{(Inkscape) Transparency is used (non-zero) for the text in Inkscape, but the package 'transparent.sty' is not loaded}%
    \renewcommand\transparent[1]{}%
  }%
  \providecommand\rotatebox[2]{#2}%
  \newcommand*\fsize{\dimexpr\f@size pt\relax}%
  \newcommand*\lineheight[1]{\fontsize{\fsize}{#1\fsize}\selectfont}%
  \ifx\svgwidth\undefined%
    \setlength{\unitlength}{720bp}%
    \ifx\svgscale\undefined%
      \relax%
    \else%
      \setlength{\unitlength}{\unitlength * \real{\svgscale}}%
    \fi%
  \else%
    \setlength{\unitlength}{\svgwidth}%
  \fi%
  \global\let\svgwidth\undefined%
  \global\let\svgscale\undefined%
  \makeatother%
  \begin{picture}(1,0.6)%
    \lineheight{1}%
    \setlength\tabcolsep{0pt}%
    \put(0,0){\includegraphics[width=\unitlength,page=1]{svg-inkscape/Ablation_bafc1_e5_svg-tex.pdf}}%
  \end{picture}%
\endgroup%

%% file: sections/related.tex
\section{Related Work} \label{sec:related}

Neural network verification~\cite{katz2017reluplex,li2023sok,kaulen20256th} provides rigorous certificates for deep neural networks under some specifications, enabling their reliable deployment in safety-critical scenarios~\cite{huang2025commit,katz2019marabou}.
Verification methods can verify a wide range of input specifications, such as robustness against locally $\ell_p$ bounded perturbations~\cite{wang2021beta,lirpa,deeppoly}, semantic distortions~\cite{li2021tss,deeppoly,balunovic2019certifying}, data poisoning~\cite{levine2021deep,weber2023rab}, and label flipping~\cite{rosenfeld2020certified}.
Typical verification methods use off-the-shelf solvers~\cite{tjeng2019mip,tsay2021partition}, linear relaxations~\cite{crown,lirpa,gehr2018ai2,weng2018towards}, dual variable optimization~\cite{wong2018provable}, branch and bound~\cite{wang2021beta,katz2019marabou}, Lipschitz-constrained layers~\cite{tsuzuku2018lipschitz,huang2021training,xu2022lot}, and randomized smoothing~\cite{cohen2019certified,salman2019provably,yang2020randomized}.

Among $\ell_p$ norms, the $\ell_0$ norm as perturbation bound type is particularly challenging for local robustness verification due to non-convexity and non-differentiability.
Existing verification methods tend to modify the inference protocol through systematic or randomized ablation~\cite{levine2020robustness,levine2020randomized}, which provides strong certificates but induces significant inference overhead.
On the other hand, direct verification of standard neural networks
is harder: bound propagation with closed-form \lo concretization~\cite{lirpa} is sound but, on its own, too loose for larger budgets.
With $\alpha$-CROWN~\cite{alphacrown} enabled, auto\_LiRPA~\cite{lirpa} tightens these bounds by optimizing the slopes of the ReLU lower relaxations with gradient ascent, but it still bounds the entire $\ell_0$-ball as a single problem, without branching.
In contrast, \calzone~\cite{calzone}, \coverd~\cite{coverd}, and \cvdtt~\cite{shapira2026tight} run branch-and-bound over covering designs of perturbable pixel sets~(see \Cref{sec:covering-bab}).
For a related $\ell_0$-structured specification, Bunel et al.~\cite{bunel2024verified} branch by support extension to verify neural compressed-sensing decoders, close in spirit to our concrete branching (\Cref{app:ext-discussion}).
Our \tool is a more general branch-and-bound-based verification method for $\ell_0$-bounded perturbations that integrates two branching strategies, with node-constrained closed-form concretization, and a strategy planner that chooses a starting level per instance.
Since \tool keeps its relaxations fixed, relaxation optimization is orthogonal to our node abstraction, which only changes the concretization step, and could in principle tighten the bound of every node at the cost of per-node optimization.

More detailed discussion of related work is deferred to \Cref{app:ext-discussion}.

%% file: sections/concl.tex
\section{Conclusion} \label{sec:concl}

We introduced \tool, a combinatorial branch-and-bound framework for verifying local robustness of neural networks against \lo perturbations.
The key insight is a node abstraction that unifies upper-shadow (concrete) and covering-design (super) branching, making the \lo-ball decomposable across levels.
On a wide range of benchmarks, \tool verifies substantially more instances than state-of-the-art verifiers, showing that exploiting the combinatorial structure of sparse perturbations is an effective direction for $\ell_0$ verification.

%% file: sections/ack.tex
\begin{ack}
We would like to thank anonymous reviewers for their feedback.
This research was conducted with the support of Natural Sciences and Engineering Research Council of Canada (NSERC) Discovery Grants, Digital Research Alliance of Canada, and Simon Fraser University (SFU) Startup Grants.

We declare no competing interests.
\end{ack}

%% file: appendices/extended-related.tex
\newpage

\section{Extended Discussion on Related Work}
\label{app:ext-discussion}

This appendix extends our discussion of related work in \Cref{sec:related}.

\paragraph{\boldmath\lo-bounded adversary and sparse attacks.}
In the $\ell_0$-bounded adversarial model, the attacker imposes perturbations to a fixed and small amount of input coordinates (i.e., pixels in the image domain) to manipulate the model predictions while remaining stealthy.
The $\ell_0$ norm, as a ``pseudo-norm'', measures the number of changed coordinates rather than the magnitude of change.
Hence, in the literature, adversary under $\ell_0$-bounded constraint is also known as few-pixel or sparse attacks~\cite{vargas2019understanding,pomponi2022pixle,sigma0}.
These attacks usually deploy an effective local search algorithm or continuous approximations combined with gradient descent to find adversarial examples under $\ell_0$-norm constraints.
Their effectiveness demonstrates the internal fragility of neural network robustness.

A closely related type of adversary is the patch attack, where the perturbable coordinates not only have a small $\ell_0$ but also need to be spatially concentrated.
While imposing more constraints, patch attacks are still powerful while stealthy~\cite{sharma2022adversarial,tao2023hard,fu2022patchfool}.
Further, the effectiveness of patch attacks in the real physical world, e.g., against autonomous driving systems~\cite{eykholt2018robust}, has been revealed, which necessitates effective defenses against $\ell_0$-bounded attacks and patch attacks.

\paragraph{Comparison with other \boldmath\lo-bounded defenses and verification methods.}
To counteract the $\ell_0$-bounded adversary, empirical defenses are proposed, which strengthen the model robustness against existing attacks~\cite{Wu2020Defending,yu2021defending,kang2024diffender}.
However, empirical defenses could not provide a rigorous guarantee in the presence of future stronger attacks.
In contrast, verification methods establish a rigorous guarantee to enable its deployment in safety-critical scenarios.

Among verification methods, one line of research is to alter the inference protocol.
Intuitively, when the prediction is derived from majority voting over model predictions on randomly ablated inputs or features, the prediction would be robust against small input perturbations.
Using this principle, randomized ablation/smoothing then aggregation~\cite{levine2021deep,zhang2020clipped,levine2020randomized,xiang2022patchcleanser} enables certified robustness against $\ell_0$ adversary and patch adversary.
However, they alter the inference procedure and incur a large inference-time overhead.

It is indeed challenging to provide robust certificates for ordinary neural networks, which is the focus of our paper.
The challenge primarily lies in the non-convexity and non-differentiability of the $\ell_0$ norm.
Besides relying on MILP~(mixed-integer programming) solver, recent tools on $\ell_0$ verification are \calzone~\cite{calzone} and \coverd~\cite{coverd} and \cvdtt\cite{shapira2026tight}.
\Cref{sec:covering-bab} describes their covering-design-based branch-and-bound precisely.
We view this framework ``branch-and-bound'' on the covering set.
Our tool \tool is built on this design and is also a branch-and-bound based framework.
However, our \tool branches in a new domain, where each node $o$ consists of a ``focus set'' $o.\gS$ and the soft perturbation budget for the focus set $o.l$.
As a result, our \tool integrates both covering-design-based branching~(super branching, which decomposes the focus set) and
upper-shadow-based
branching~(concrete branching, which extends the focus set).
The extended branch-and-bound enables a new dimension of branching design choices and thus significantly improves the verification efficiency while remaining tight, as our analysis and experiments have shown.

\paragraph{\boldmath\lo-bounded in broader domains.}
\tool and baseline approaches are primarily evaluated in the image domain following the literature~\cite{crown,lirpa,calzone,coverd}.
The $\ell_0$ adversary is also studied in other modalities such as tabular data~\cite{chen2021learning} and language data~\cite{wang2021certified}.
In particular, in the language domain, the $\ell_0$ adversary is known as word substitution attacks or suffix jailbreak attacks, where corresponding certified methods are primarily based on majority voting, randomized smoothing, and ablation~\cite{wang2021certified,ye2020safer,robey2023smoothllm}.
Though these methods incur large inference overhead, the other way, direct verification, is still challenging due to the non-linearity and large scales of language models~\cite{Shi2020Robustness}.
We leave as future work to further scale the framework to support verification against $\ell_0$-bounded adversary for language models.

\paragraph{Relation to support-extension branching.}\label{app:support-extension}
Bunel et al.~\cite{bunel2024verified} verify that a neural compressed-sensing decoder recovers the support of every signal with a given number of non-zero entries. It branches by extending a fixed support one coordinate at a time in canonical index order, with every skipped coordinate fixed to zero (the ``off''-set). Once a support is fully determined, Bunel et al.\cite{bunel2024verified} continue by splitting the input box instead of calling MILP. Their off-set is technically compatible with our node abstraction by excluding the pixels in off-set in the concretization (\Eqref{eq:mingx} and \Eqref{eq:maxgx}) steps, which can slightly tighten bounds.
However, it can change the branching domain in two ways. First, every node less than level $\epsilon$ would carry its own partition of the pixels, which makes nodes more expensive to represent and bound. Second, off-sets make children disjoint and thus rule out cross-parent pruning (\Cref{sec:sap}). Specifically, without off-sets, a level-$(l+1)$ concrete node is dropped as soon as any of its $l+1$ parents is proven safe, whereas with off-sets only its canonical parent's safety implies its own.

%% file: appendices/limitations.tex
\section{Limitations} \label{sec:limitations}

While \tool achieves state-of-the-art efficiency on the evaluated benchmarks, exact \lo verification remains inherently combinatorial and computationally hard. Therefore, within practical time and memory budgets, \tool is evaluated on networks with up to thousands of neurons rather than the largest neural networks. This limitation does not affect the soundness of the verification results.
To further improve the scalability, a promising path forward is to combine the branch-and-bound framework with stronger neural relaxations or tighter bound propagation techniques. We leave the extension to other domains such as language models as future work.

Another limitation is that \tool is designed primarily as a verifier, so it is not optimized for quickly finding adversarial examples. In particular, \tool manages nodes in a breadth-first manner to enable batched bounding~(\Cref{sec:sap}). This design choice substantially improves verification throughput, but it can delay the exploration of deep unbranchable nodes where MILP may quickly expose an adversarial example. In practice, this limitation can be mitigated by running complementary tools to find adversarial examples before verification, as done in our evaluation.

In addition, covering-design-based super branching may introduce duplicated bounding effort, because different child super nodes can cover overlapping perturbation cases. Reducing such redundancy while preserving the efficacy of super-node branching is an important direction for future work.

Finally, the planner used in our experiments is sensitive to timing noise, so repeated runs may verify slightly different sets of borderline instances (\Cref{app:parallelization-ablation,app:reproducibility-note}).

%% file: appendices/impacts.tex
\section{Societal Impacts} \label{sec:impacts}

As positive societal impacts, local robustness verification of image classifiers can help people better understand and improve the reliability and stability of neural networks. If a network is verified to be robust for a given image under a perturbation budget $\epsilon$, then the classifier's prediction is guaranteed to remain unchanged for any perturbation involving up to $\epsilon$ pixels. Thus, robustness verification has great potential to help protect against adversarial attacks in critical settings such as medical imaging, autonomous driving, and security screening.
As for negative societal impacts, we are not aware of any direct harms arising specifically from local robustness verification.

%% file: appendices/Formula.tex
\section{Formulas}\label{sec:Formulas}
Here, we derive a closed-form solution to compute the minimum of any affine form $g(\vx)=\omega^\T\vx+b$. For each pixel, its least possible contribution can be derived as
\begin{equation}\label{eq:pixel-min}
    \lD_v
    =\min_{\vx_v \in [\lvx_v,\uvx_v]}\Delta_v
    =\sum^C_{c=1}\omega_{v,c}(\vx^\prime_{v,c}-\nomvx_{v,c}),\quad
    \text{where }\vx^\prime_{v,c}=
        \begin{cases}
            \uvx_{v,c}, &  \omega_{v,c} \le 0,\\[2pt]
            \lvx_{v,c}, &  \omega_{v,c} > 0,
        \end{cases}
\end{equation}
and the minimum of $g(\vx)$ over a node-constrained space $\eoball$ can be derived as follows:
\begin{equation}\label{eq:mingx}
    \min_{\vx \in \eoball} g(\vx)
= g(\nomvx) + \min_{\vx\in\eoball}\sum_{v=1}^V\Delta_v
= g(\nomvx) + \sum_{v \in
\underline{\gA}\left(o.l,o.\gS\right)\cup\underline{\gA}\left(\epsilon-o.l,\gV\setminus\underline{\gA}\left(o.l,o.\gS\right)\right)}\lD_v,
\end{equation}
where $\underline{\gA}(a,A):=\arg{\operatorname{top}\text{-}a_{v \in A}-\lD_v}$ is a set of $a$ (where $a\in\set{0, \ldots, \epsilon}$) most negatively contributed pixels from set $A\subseteq\gV$.

In practice, the evaluation of \Eqref{eq:maxgx} and \Eqref{eq:mingx} is CUDA-accelerated.

%% file: appendices/proofs.tex
\section{Proofs}\label{sec:proofs}

\begin{lemma}[Node-constrained concretization] \label{lem:node-constrained-concretization}
Let $\nomvx\in\gD$ and $\eps<V$. For any affine form $g(\vx)=\omega^\top\vx+b$ and any node $o$, the RHS of \Eqref{eq:maxgx} evaluates to $\max_{\vx \in \eoball} g(\vx)$, for any choice of the top-a sets when ties occur. Similarly, the RHS of \Eqref{eq:mingx} evaluates to the $\min_{\vx \in \eoball} g(\vx)$.
\end{lemma}
\begin{proof}
Write $T(\vx) := \{v : \vx_v \neq \nomvx_v\}$ and
\[
  \gT \;:=\; \bigl\{\, T \subseteq \gV \;:\; |T| \le \eps,\;
  |T \setminus o.\gS| \le \eps - o.l \,\bigr\},
\]
so that $\eoball = \{\vx \in \gD : T(\vx) \in \gT\}$ by
\Eqref{eq:eball} and \Eqref{eq:eoball}. Note that $\gT$ is downward
closed.

\emph{Step 0 (decoupling).}
We claim
\begin{equation}\label{eq:decoupled}
  \max_{\vx \in \eoball} g(\vx)
  \;=\; g(\nomvx) \;+\; \max_{T \in \gT} \sum_{v \in T} \uD_v .
\end{equation}
For ``$\le$'': let $\vx \in \eoball$. Then $\Delta_v = 0$ for
$v \notin T(\vx)$ and $\Delta_v \le \uD_v$ for $v \in T(\vx)$ by
\Eqref{eq:pixel-max}, so
$g(\vx) \le g(\nomvx) + \sum_{v \in T(\vx)} \uD_v$, and
$T(\vx) \in \gT$. For ``$\ge$'': fix $A \in \gT$ and define $\vx$ by
$\vx_v := \vx'_v$ for $v \in A$ and $\vx_v := \nomvx_v$ otherwise,
with $\vx'_v$ as in \Eqref{eq:pixel-max}. Then $\vx \in \gD$ and
$T(\vx) \subseteq A$, so $T(\vx) \in \gT$ by downward closure; in
particular $|T(\vx)| \le \eps$ gives $\vx \in \eball$ and
$|T(\vx) \setminus o.\gS| \le \eps - o.l$ gives $\vx \in \eoball$.
Moreover $\Delta_v = \uD_v$ for every $v \in A$ and $\Delta_v = 0$
elsewhere, so $g(\vx) = g(\nomvx) + \sum_{v \in A} \uD_v$.

\emph{Step 1 (an optimal set uses the full budget).}
Since $\gT$ is finite and nonempty, the maximum on the right of
\Eqref{eq:decoupled} is attained; we call any $T \in \gT$ attaining
it a \emph{maximizer}. Since $\nomvx \in \gD$, the choice
$\vx_v = \nomvx_v$ is feasible in \Eqref{eq:pixel-max}, hence
\begin{equation}\label{eq:nonneg}
  \uD_v \;\ge\; 0 \qquad \text{for every } v \in \gV .
\end{equation}
Let $T \in \gT$ with $|T| < \eps$; we exhibit $u \notin T$ with
$T \cup \{u\} \in \gT$. If $o.\gS \not\subseteq T$, pick
$u \in o.\gS \setminus T$; then
$(T \cup \{u\}) \setminus o.\gS = T \setminus o.\gS$ and the node
constraint is unaffected. If $o.\gS \subseteq T$, then
$|T \setminus o.\gS| = |T| - |o.\gS| \le |T| - o.l < \eps - o.l$
using $|o.\gS| \ge o.l$, so any $u \in \gV \setminus T$ (which exists
because $|T| < \eps \le V$) satisfies
$|(T \cup \{u\}) \setminus o.\gS| \le |T \setminus o.\gS| + 1
 \le \eps - o.l$. In both cases $|T \cup \{u\}| = |T| + 1 \le \eps$.
By \Eqref{eq:nonneg} the objective does not decrease, so iterating
yields a maximizer of size exactly $\eps$. Any $T \subseteq \gV$ with
$|T| = \eps$ satisfies $|T \setminus o.\gS| = \eps - |T \cap o.\gS|$,
whence
\begin{equation}\label{eq:inside-form}
  \text{for every } T \subseteq \gV \text{ with } |T| = \eps: \quad
  |T \setminus o.\gS| \le \eps - o.l
  \iff |T \cap o.\gS| \ge o.l .
\end{equation}

\emph{Step 2 (exchange into $P$).}
Put $P := \overline{\gA}(o.l, o.\gS)$, so $|P| = o.l$,
$P \subseteq o.\gS$, and $\uD_p \ge \uD_q$ for all $p \in P$ and
$q \in o.\gS \setminus P$. This is the only property of
$\overline{\gA}$ the argument uses, so ties may be broken
arbitrarily. Let $T$ be a maximizer with $|T| = \eps$ and suppose
$P \not\subseteq T$; fix $p \in P \setminus T$. Since $p \notin T$
and $P \subseteq o.\gS$ we have $|T \cap P| \le |P| - 1 = o.l - 1$,
and $|T \cap o.\gS| \ge o.l$ by \Eqref{eq:inside-form}, so
$|(T \cap o.\gS) \setminus P| \ge 1$; fix $q$ in this set. Then
$\uD_p \ge \uD_q$, so $T' := (T \setminus \{q\}) \cup \{p\}$
satisfies
$\sum_{v \in T'} \uD_v
 = \sum_{v \in T} \uD_v - \uD_q + \uD_p
 \ge \sum_{v \in T} \uD_v$.
Moreover $|T'| = \eps$ and
$|T' \cap o.\gS| = |T \cap o.\gS| \ge o.l$, so $T' \in \gT$ by
\Eqref{eq:inside-form} and $T'$ is again a maximizer. As
$|T' \cap P| = |T \cap P| + 1$ and $|T \cap P|$ is bounded by
$|P| = o.l$, iterating terminates at a maximizer containing $P$.

\emph{Step 3 (exchange into $Q$).}
Put $Q := \overline{\gA}(\eps - o.l,\, \gV \setminus P)$, which is
well defined since $\eps - o.l \le V - o.l = |\gV \setminus P|$.
First observe that every $T \supseteq P$ with $|T| = \eps$ lies in
$\gT$: indeed $|T \cap o.\gS| \ge |P| = o.l$ because
$P \subseteq o.\gS$, so \Eqref{eq:inside-form} applies. Hence
$T \setminus P$ ranges over all $(\eps - o.l)$-subsets of
$\gV \setminus P$, unconstrained.

Let $T$ be a maximizer with $P \subseteq T$, $|T| = \eps$, and
suppose $T \setminus P \neq Q$. Since $Q \subseteq \gV \setminus P$
we have $Q \setminus T = Q \setminus (T \setminus P)$, and as
$|T \setminus P| = |Q| = \eps - o.l$ it follows that
$|(T \setminus P) \setminus Q| = |Q \setminus T| \ge 1$. Fix
$v \in (T \setminus P) \setminus Q$ and $w \in Q \setminus T$; then
$\uD_w \ge \uD_v$ by the definition of $Q$, so
$T'' := (T \setminus \{v\}) \cup \{w\}$ is again a maximizer
containing $P$, with $|T'' \cap Q| = |T \cap Q| + 1$. As
$|T \cap Q|$ is bounded by $|Q| = \eps - o.l$, iterating terminates
at $T = P \cup Q$. Since $Q \subseteq \gV \setminus P$, this union
is disjoint and $|P \cup Q| = o.l + (\eps - o.l) = \eps$, as
required. Substituting into \Eqref{eq:decoupled} gives
\Eqref{eq:maxgx}.

\emph{Minimum.}
Apply the above to $-g$. Its per-pixel contribution is $-\Delta_v$,
so its best contribution is
$\max_{\vx_v} (-\Delta_v) = -\lD_v$ with $\lD_v$ as in
\Eqref{eq:pixel-min}, and selecting the largest values of $-\lD_v$
is exactly the rule defining $\underline{\gA}$ in
\Eqref{eq:mingx}. Since
$\min_{\vx \in \eoball} g(\vx) = -\max_{\vx \in \eoball} (-g)(\vx)$,
\Eqref{eq:mingx} follows.
\end{proof}

\begin{lemma}[Validity of node-constrained bounds]\label{lem:bound-validity}
Under the assumptions of \Cref{lem:node-constrained-concretization}, let $o$ be any node. For each $j\neq\nomy$, let $\lvm$ be the affine lower bound of the margin $m_{\nomy,j}$ computed by LiRPA-style bound propagation~\cite{lirpa,crown,deeppoly} in which every concretization step, for all intermediate affine layers and for the margin itself, is performed over $\eoball$ via \Eqref{eq:maxgx} and \Eqref{eq:mingx}. (Thus $\lvm$ depends on $o$ through the intermediate concrete bounds; as in \Eqref{eq:bound-o}, we keep this dependence implicit.) Then, for every $j\neq\nomy$, we have
\begin{equation}\label{eq:lvm-valid}
    \forall\vx\in\eoball.\quad\lvm(\vx)\le m_{\nomy,j}(\vx),
\end{equation}
and consequently
\begin{equation}\label{eq:bound-valid}
    \Bound{o}\le\min_{j\neq\nomy}~\min_{\vx\in\eoball} m_{\nomy,j}(\vx),
\end{equation}
i.e., $\Bound{o}$ is a valid concrete lower bound of the margins over $\eoball$. Moreover, the right-hand side of \Eqref{eq:mingx} with $g=\lvm$ evaluates to exactly $\min_{\vx\in\eoball}\lvm(\vx)$, so the closed-form value computed in practice coincides with $\Bound{o}$ as defined in \Eqref{eq:bound-o}.
\end{lemma}

\begin{proof}
Fix $o$ and write $\gX:=\eoball$. Let $h^{(1)},\dots,h^{(K)}$ be the affine layers of $f$ in topological order. For each $i$, the bound propagation back-propagates linear relaxations to the input to obtain affine functions $\underline{h}^{(i)}$ and $\overline{h}^{(i)}$ of $\vx$, and concretizes them entrywise into concrete bounds $\vl^{(i)}$ and $\vu^{(i)}$, which are then used to linearly relax the nonlinear activation that takes $h^{(i)}$ as input. The affine lower bound $\lvm$ is obtained by the same back-propagation applied directly to the margin $m_{\nomy,j}$.

\emph{Step 1 (soundness of linear relaxation).}
Let $q$ be either an affine layer $h^{(i)}$ or a margin $m_{\nomy,j}$, and let $\underline{q}$ and $\overline{q}$ be the affine bounds obtained by back-propagating from $q$ (for $q=m_{\nomy,j}$, only $\underline{q}=\lvm$ is needed). The soundness of LiRPA-style bound propagation~\cite{lirpa,crown,deeppoly} rests on the following pointwise property: for every input $\vx$ at which each activation preceding $q$ receives an input within the concrete bounds used for its relaxation, i.e., $\vl^{(i')}\le h^{(i')}(\vx)\le\vu^{(i')}$ for every layer $h^{(i')}$ feeding such an activation, we have
\begin{equation}\label{eq:lirpa-sound}
    \underline{q}(\vx)\le q(\vx)\le\overline{q}(\vx).
\end{equation}
Indeed, each linear relaxation sandwiches its activation on the interval given by the concrete bounds of its input, and back-propagating the relaxations through affine layers preserves the inequalities. Note that \Eqref{eq:lirpa-sound} does not refer to any input set: the input set enters the pipeline only through the concretization step that produces the concrete bounds, which prior work instantiates for boxes~\cite{deeppoly}, $\ell_p$-balls~\cite{crown}, or the full $\ell_0$-ball $\eball$~\cite{lirpa}. Our bounding procedure follows exactly the same pipeline, including all intermediate lower and upper bounds, and differs only in that every concretization is performed over $\gX$ via \Eqref{eq:maxgx} and \Eqref{eq:mingx}. It therefore suffices to show that the resulting concrete bounds are valid on $\gX$.

\emph{Step 2 (intermediate bounds).}
We show by strong induction on $i$ that $\vl^{(i)}\le h^{(i)}(\vx)\le\vu^{(i)}$ for all $\vx\in\gX$. Assume this holds for all $i'<i$ (vacuously for $i=1$, where no activation precedes $h^{(1)}$ and hence $\underline{h}^{(1)}=\overline{h}^{(1)}=h^{(1)}$). Every activation preceding $h^{(i)}$ takes as input some $h^{(i')}$ with $i'<i$, so by the induction hypothesis the premise of \Eqref{eq:lirpa-sound} holds at every $\vx\in\gX$, which gives $\underline{h}^{(i)}(\vx)\le h^{(i)}(\vx)\le\overline{h}^{(i)}(\vx)$ for all $\vx\in\gX$. Each entry of $\underline{h}^{(i)}$ and $\overline{h}^{(i)}$ is an affine form of $\vx$, and $\vl^{(i)}$ (resp.\ $\vu^{(i)}$) is obtained by evaluating the right-hand side of \Eqref{eq:mingx} (resp.\ \Eqref{eq:maxgx}) on each entry of $\underline{h}^{(i)}$ (resp.\ $\overline{h}^{(i)}$). By \Cref{lem:node-constrained-concretization}, these values are exactly the entrywise $\min_{\vx'\in\gX}\underline{h}^{(i)}(\vx')$ and $\max_{\vx'\in\gX}\overline{h}^{(i)}(\vx')$, respectively. Hence, for every $\vx\in\gX$,
\begin{equation*}
    \vl^{(i)}=\min_{\vx'\in\gX}\underline{h}^{(i)}(\vx')
    \le\underline{h}^{(i)}(\vx)\le h^{(i)}(\vx)\le\overline{h}^{(i)}(\vx)
    \le\max_{\vx'\in\gX}\overline{h}^{(i)}(\vx')=\vu^{(i)}.
\end{equation*}

\emph{Step 3 (margin bound).}
By Step 2, all concrete bounds are valid on $\gX$, so the premise of \Eqref{eq:lirpa-sound} holds for $q=m_{\nomy,j}$ at every $\vx\in\gX$, which gives \Eqref{eq:lvm-valid} for every $j\neq\nomy$. Taking the minimum of both sides of \Eqref{eq:lvm-valid} over $\vx\in\gX$ and then over $j\neq\nomy$ yields \Eqref{eq:bound-valid} by \Eqref{eq:bound-o}. Finally, since $\lvm$ is an affine form of $\vx$, \Cref{lem:node-constrained-concretization} shows that the right-hand side of \Eqref{eq:mingx} with $g=\lvm$ evaluates to exactly $\min_{\vx\in\gX}\lvm(\vx)$, so $\Bound{o}$ is computed in closed-form, without enumerating the perturbation patterns admitted by $o$.

Note that \Eqref{eq:lvm-valid} is claimed only on $\eoball$: the intermediate bounds, and hence the relaxations and $\lvm$, are specific to $o$ and need not remain valid on $\eball\setminus\eoball$. This suffices, because both \Eqref{eq:o-local-robust} and \Eqref{eq:bound-o} quantify only over $\eoball$.
\end{proof}

\begin{lemma}[Bound soundness]\label{lem:bound-soundness}
For any node $o$, if $\Bound{o}>0$, then $\Safe{o}$ holds.
\end{lemma}
\begin{proof}
By \Cref{lem:bound-validity}, $$\min_{j\neq\nomy}\min_{\vx\in\eoball} m_{\nomy,j}(\vx)\ge\Bound{o}.$$ Hence, if $\Bound{o}>0$, the LHS is strictly positive, which is exactly $\Safe{o}$ by \Eqref{eq:o-local-robust}.
\end{proof}

\begin{lemma}[Initial node sufficiency]\label{lem:Initial Node Safety}
Recall the definition of the initial node for each level $l$:
\begin{equation}\label{eq:initnode}
\Theta_l=
\begin{cases}
\left(\emptyset,0\right), & \text{if } l=0,\\[4pt]
\left(\gV,l\right), & \text{if } l>0 .
\end{cases}
\end{equation}
If $\Safe{\Theta_l}$ holds for some $l\in \set{0,\dots,\epsilon}$, then $f$ is \lo locally robust at $\nomvx$ with budget $\eps$.
\end{lemma}
\begin{proof}
    We can observe that the only difference between \Eqref{eq:local-robust} and \Eqref{eq:o-local-robust} is the input space: $\eball$ for \Eqref{eq:local-robust}, while $\eoball$ for \Eqref{eq:o-local-robust}. Comparing \Eqref{eq:eoball} with \Eqref{eq:eball}, $\eoball$ enforces a constraint on the set of perturbed pixels in addition to $\eball$. Thus, we have $\eoball\subseteq\eball$, and the equality holds when $o.\gS=\gV$ or $o.l=0$. More precisely, if a node $o$ satisfies $o.\gS=\gV$ or $o.l=0$, and $\Safe{o}$ holds, then \Eqref{eq:local-robust} holds and the local robustness of $f$ over $\eball$ is verified. By the definition of initial nodes~\Eqref{eq:initnode}, for any initial node $\Theta_l$, it satisfies $\Theta_l.\gS=\gV$ or $\Theta_l.l=0$. Therefore, when $\Safe{\Theta_l}$ holds, then \Eqref{eq:local-robust} holds and the local robustness of $f$ over $\eball$ is verified.
\end{proof}

\begin{lemma}[Child containment]\label{lem:child-containment}
Given a branchable node $o$, we have
\[
    \forall o'\in\Branch{o}.\quad \eopball \subseteq \eoball .
\]
I.e., every child ball is contained in its parent ball.
\end{lemma}

\begin{proof}
Let us discuss $o$ in two cases.
\begin{enumerate}
\item If $o = (o.\gS, o.l)$ is a super node, then its child node is derived from $o' \in \sBranch{k}{o}$. By \Eqref{eq:superbranch}, we have $o' = (o'.\gS, o.l)$ where $o'.\gS \subset o.\gS$ and $o.l \leq k = |o'.\gS| < |o.\gS|$.
Let $S$ be a non-empty set s.t. $S = o.\gS \setminus o'.\gS$.
By \Eqref{eq:eoball}, we have
\[
\begin{array}{rcl}

\eopball
&=& \set{\vx\in\eball:~|\set{v:~\vx_v\neq\nomvx_v}\setminus o'.\gS|\le\epsilon-o.l} \\
&=& \set{\vx\in\eball:~|\set{v:~\vx_v\neq\nomvx_v}\setminus (o.\gS \setminus S)|\le\epsilon-o.l} \\
&\subseteq& \set{\vx\in\eball:~|\set{v:~\vx_v\neq\nomvx_v}\setminus o.\gS|\le\epsilon-o.l} \\
&=& \eoball
\end{array}
\]

\item If $o = (o.\gS, o.l)$ is a concrete node, then its child node is derived from $o' \in \cBranch{o}$. By  \Eqref{eq:concretebranch}, we have $o' = (o'.\gS, o.l+1)$ where $o'.\gS = o.\gS \cup \{v'\}$ and $v' \not \in o.\gS$. Further, by \Eqref{eq:eoball}, we have
\[
\begin{array}{rcl}
\eopball
&=& \set{\vx\in\eball:~|\set{v:~\vx_v\neq\nomvx_v}\setminus o'.\gS|\le\epsilon-o'.l} \\
&=& \set{\vx\in\eball:~|\set{v:~\vx_v\neq\nomvx_v}\setminus (o.\gS \cup \{v'\})|\le\epsilon-(o.l+1)} \\

&\subseteq& \set{\vx\in\eball:~|\set{v:~\vx_v\neq\nomvx_v}\setminus o.\gS|\le\epsilon-o.l} \\
&=& \eoball \\

\end{array}
\]
   where the inclusion holds because, for any set $T$ of pixels, $T\setminus o.\gS\subseteq\big(T\setminus(o.\gS\cup\set{v'})\big)\cup\set{v'}$, and hence $|T\setminus o.\gS|\le|T\setminus(o.\gS\cup\set{v'})|+1$.
\end{enumerate}

By the above two cases, we prove $\eopball \subseteq \eoball$ for $\branch$.
\end{proof}

\begin{lemma}[Branching exhaustiveness]\label{lem:branching-exhaustiveness}
For any branchable node $o$, the union of child balls generated by $\Branch{o}$ contains the parent ball:
\begin{equation}\label{eq:branching-exhaustiveness}
    \bigcup_{o'\in\Branch{o}}\eopball \supseteq \eoball .
\end{equation}
It applies to both branching policies. If $o$ is a super node, then
\begin{equation}\label{eq:super-branching-exhaustiveness}
\bigcup_{o'\in\sBranch{k}{o}}\eopball \supseteq \eoball
\end{equation}
holds; if $o$ is a concrete node, then
\begin{equation}\label{eq:concrete-branching-exhaustiveness}
\bigcup_{o'\in\cBranch{o}}\eopball \supseteq \eoball
\end{equation}
holds.
\end{lemma}

\begin{proof}
Let us discuss $o$ in two cases.
\begin{enumerate}

\item

For any $\vx \in \eoball$ where $o = (o.\gS, o.l)$ is a super node, we have $|o.\gS| > o.l$. Writing $T = \set{v:~\vx_v\neq\nomvx_v}$, by \Eqref{eq:eball} and \Eqref{eq:eoball}, we have $|T|\le\eps$ and $|T\setminus o.\gS| \leq \epsilon-o.l$.

Let $T_i \subseteq T \cap  o.\gS$ be any subset with $|T_i| = \min(o.l, |T \cap o.\gS|)$.
Also, since $T_i \subseteq o.\gS$ and $|T_i| = \min(o.l, |T \cap o.\gS|) \leq o.l < |o.\gS|$, there exists $T' \subseteq  o.\gS$ s.t. $|T'| = o.l$ and $T_i \subseteq T' \subseteq o.\gS$.

By the definition of covering design, there exists $o'.\gS\in\gC(o.\gS,k,o.l)$ s.t. $T' \subseteq o'.\gS \subset o.\gS$. Therefore, we have $T_i \subseteq T' \subseteq o'.\gS \subset o.\gS$. Moreover, since $T_i \subseteq T$ and $T_i \subseteq o'.\gS$, we know $T_i \subseteq T \cap o'.\gS$ and hence $|T_i| = \min(o.l, |T \cap o.\gS|) \leq |T \cap o'.\gS|$.

Then we discuss $|T \cap o.\gS|$ in two cases.

\begin{enumerate}
\item $|T \cap o.\gS| \geq o.l$.
Then $|T_i| = \min(o.l, |T \cap o.\gS|) = o.l \leq |T \cap o'.\gS|$ and $|T \setminus o'.\gS| = |T| - |T \cap o'.\gS| \leq |T| - o.l \leq \epsilon - o.l$.
\item $|T \cap o.\gS| < o.l$.
Then $|T_i| = \min(o.l, |T \cap o.\gS|) = |T \cap o.\gS| \leq |T \cap o'.\gS|$.
Since $T_i = T \cap o.\gS \subseteq o'.\gS$, then $T \setminus o'.\gS \subseteq T \setminus o.\gS$.
Namely, we can prove $|T \setminus o'.\gS| \leq |T \setminus o.\gS| \leq \epsilon - o.l$.
\end{enumerate}

Both cases verify that $|T \setminus o'.\gS| \leq \epsilon - o.l$.
By \Eqref{eq:eoball}, we can prove if $\vx \in \eoball$, then there exists $o' \in \sBranch{k}{o}$ s.t. $\vx \in \eopball$. Hence \Eqref{eq:super-branching-exhaustiveness} holds.

\item If $o = (o.\gS, o.l)$ is a concrete node, then its child node is derived from $o' \in \cBranch{o}$.

Let $\vx' \in \eoball$. By \Eqref{eq:eoball}, $\vx'$ satisfies $|\set{v:~\vx'_v \neq \nomvx_v}\setminus o.\gS|\le\epsilon-o.l$.
Further, by \Eqref{eq:concretebranch}, we know there exists a child node $o' \in \cBranch{o}$ s.t. $o'.\gS = o.\gS \cup \set{v'}$ where $v' \in \gV\setminus o.\gS$.

On the one hand, if $\set{v:~\vx'_v \neq \nomvx_v} \subseteq o.\gS$, then $|\set{v:~\vx'_v \neq \nomvx_v}\setminus o.\gS| = 0$. Furthermore, $|\set{v:~\vx'_v \neq \nomvx_v}\setminus (o.\gS \cup \{v'\})| = 0$ where $v' \in \gV\setminus o.\gS$. Also, as $o$ is a branchable concrete node, $o.l < \epsilon$, $\epsilon-o.l >0$. Then $\epsilon-(o.l +1) \ge 0$. For $\vx'$, $|\set{v:~\vx'_v \neq \nomvx_v}\setminus (o.\gS \cup \{v'\})|\le\epsilon-(o.l+1)$ still holds. Namely, by \Eqref{eq:concretebranch}, $\vx' \in \eopball$ holds.

On the other hand, if $\set{v:~\vx'_v \neq \nomvx_v} \not \subseteq o.\gS$, then $|\set{v:~\vx'_v \neq \nomvx_v}\setminus o.\gS| > 0$. That's to say, there exists $v' \in \set{v:~\vx'_v \neq \nomvx_v}$ but $v' \not \in o.\gS$. Then $v' \in \gV \setminus o.\gS$ and $|\set{v:~\vx'_v \neq \nomvx_v}\setminus (o.\gS \cup \set{v'})| = |\set{v:~\vx'_v \neq \nomvx_v}\setminus o.\gS| - 1$. Furthermore, since $|\set{v:~\vx'_v \neq \nomvx_v}\setminus o.\gS| \leq \epsilon - o.l$, then $|\set{v:~\vx'_v \neq \nomvx_v}\setminus (o.\gS \cup \set{v'})| \leq \epsilon - o.l - 1 = \epsilon - (o.l+1)$.
By \Eqref{eq:concretebranch}, we can find a child node $o' = (o.\gS \cup \set{v'}, o.l+1) \in \cBranch{o}$ s.t. $\vx' \in \eopball$.

By the above two cases, we prove if $\vx' \in \eoball$, then $\vx' \in  \bigcup_{o'\in\Branch{o}}\eopball$. In another words, $\bigcup_{o'\in\cBranch{o}}\eopball \supseteq \eoball$ holds when $o$ is a concrete node.

\end{enumerate}
By the above two cases, we prove \Eqref{eq:super-branching-exhaustiveness} and \Eqref{eq:concrete-branching-exhaustiveness}, and thus \Eqref{eq:branching-exhaustiveness} is proven.

\end{proof}

\begin{corollary}[Parent safety from children]\label{cor:sound-branching}
If all child nodes generated by $\Branch{o}$ are safe, then the parent node
$o$ is safe.
\end{corollary}
\begin{proof}
    By Lemma~\ref{lem:branching-exhaustiveness}, every input in $\eoball$ belongs to at least one child ball. Since all child nodes are safe, all inputs in all child balls are safe. Hence all inputs in $\eoball$ are safe.
\end{proof}

\begin{corollary}[Branching exactness]\label{cor:branching-exactness}
For any branchable node $o$, the union of all child balls generated by $\Branch{o}$ equals exactly the parent ball, i.e.,
\begin{equation}\label{eq:branching-exactness}
    \bigcup_{o'\in\Branch{o}}\eopball = \eoball .
\end{equation}
\end{corollary}
\begin{proof}
By Lemma~\ref{lem:child-containment}, for any child node $o' \in \Branch{o}$, we have $\eopball \subseteq \eoball$. Hence

\[
\bigcup_{o'\in\Branch{o}}\eopball \subseteq \eoball.
\]
The reverse is proven by Lemma~\ref{lem:branching-exhaustiveness}. Combining the two inclusions gives \Eqref{eq:branching-exactness}.
\end{proof}

\relsoundness*

\begin{proof}

By Lemma~\ref{lem:Initial Node Safety}, we know network $f$ is locally robust w.r.t. $\nomvx$ and $\epsilon$ if the initial node $\Theta_l$ is safe, i.e., $\Safe{\Theta_l}$. However, directly proving $\Theta_l$ is safe is challenging; instead, we prove  safety indirectly using decomposition. By Corollary~\ref{cor:sound-branching}, we know that $\Theta_l$ is provably safe if all its child nodes generated by $\branch$ are safe. On the one hand, by~\Cref{lem:bound-soundness}, for any child node $o'$ meeting $\Bound{o'} > 0$, it is proven safe as $\Bound{o}$ estimates the lower bound of the margin function by \Eqref{eq:bound-o} and \Eqref{eq:o-local-robust}. On the other hand, the rest child nodes are proven safe if an MILP solver return \texttt{infeasible} for all these nodes and cannot identify any adversarial example. Furthermore, since we assume the MILP solver is sound, it can correctly validate $\texttt{MILP}(f, \eopball)$ for any child node $o'$.

If $\Tool{f, \nomvx, \epsilon}$ returns \texttt{Verified}, then $\gO$ is dequeued to be empty as the MILP solver returns \texttt{infeasible} for all child nodes and $\Bound{o}$ directly proves a node is safe. Then $\Safe{\Theta_l}$ is proven true and, therefore, $f$ is locally robust w.r.t. $\nomvx$ and $\epsilon$ by Lemma~\ref{lem:Initial Node Safety}.

\end{proof}

\begin{lemma}[Finite Branching] \label{lem:finite-branching}
If $o$ is branchable, then $\Branch{o}$ generates non-empty, finite child nodes.
\begin{enumerate}
\item
If $o=(o.\gS,o.l)$ is a super node, $\sBranch{k}{o}$ contains $|\gC(o.\gS,k,o.l)|$ nodes, each child node $o'$ is of the form $(o'.\gS,o.l)$ with $o'.\gS \subset o.\gS$ and $|o'.\gS|=k$, where $o.l\le k < |o.\gS|$.
\item
If $o$ is a concrete node, $\cBranch{o}$ contains exactly $V - o.l$ nodes, each of the form $(o.\gS \cup \set{v}, o.l+1)$ with $|o.\gS \cup \set{v}| = o.l + 1$.
\end{enumerate}
\end{lemma}

\begin{proof}
Let us discuss $o$ into two cases.
\begin{enumerate}
\item
If $o$ is a super node, then, according to \calzone~\cite{calzone} and the definition of covering design~\cite{GordonKuperbergPatashnik1995,Schonheim1964,Todorov1985}, $\gC(o.\gS,k,o.l)$ in \Eqref{eq:superbranch} is a finite non-empty collection of $k$-element subsets of $o.\gS$ where $k < |o.\gS|$ for super node $o$.

\item
If $o$ is a concrete node, then, by \Eqref{eq:concretebranch}, any child node $o' \in \cBranch{o}$ has one more element (i.e., $o'.\gS = o.\gS \cup \set{v}$ where $v \in\gV\setminus o.\gS$) and increments its level by one (i.e., $o'.l = o.l + 1$)  for concrete node $o$. In total, $\cBranch{o}$ generates $V - |o.\gS|$ child nodes where $o.l = |o.\gS|$. That is to say, $\cBranch{o}$ contains exactly $V - |o.\gS| = V - o.l$ nodes, each of the form $(o.\gS \cup \set{v}, o.l+1)$ with $|o.\gS \cup \set{v}| = o.l + 1$.
\end{enumerate}

From the above two cases, we prove the lemma.
\end{proof}

\begin{definition}[Progress measure]\label{def:progress-measure}
We use $\mu$ to measure the progress of a node $o$ during the process of branching propagation. Specifically, define
\[
  \mu(o) := (\epsilon - o.l,\ |o.\gS|) \in \{0,\dots,\epsilon\} \times \{0,\dots,V\},
\]
ordered lexicographically: $(a,b) <_{\mathrm{lex}} (a',b')$ iff $a < a'$, or $a = a'$ and $b < b'$.
\end{definition}

\begin{lemma}[Branching progress]\label{lem:branching-progress}
For every branchable node $o$ and every child node $o' \in \textsc{Branch}(o)$, we have
$\mu(o') <_{\mathrm{lex}} \mu(o)$.
\end{lemma}
\begin{proof}
We discuss $o$ into two cases.
\begin{enumerate}
\item
If $o$ is a super node, then by \Eqref{eq:superbranch}, $o'.l = o.l$ and $|o'.\gS| = k < |o.\gS|$, so the first components of $\mu(o')$ and $\mu(o)$ agree and the second strictly decreases.
\item
If $o$ is a concrete node, then by \Eqref{eq:concretebranch}, $o'.l = o.l + 1$, so the first component strictly decreases (this suffices for $<_{\mathrm{lex}}$ even though the second component grows by one).
\end{enumerate}
Both decrease their child nodes' progress measure.
\end{proof}

\begin{lemma}[Concrete propagation] \label{lem:concrete-propagation}
For any branchable concrete node $o$, every child node in $\cBranch{o}$ is a concrete node.
\end{lemma}
\begin{proof}
    By \Eqref{eq:concretebranch}, every child node $o'$ generated by $\cBranch{o}$ satisfies $o'.\gS=o.\gS \cup \set{v}$ and $ o'.l=o.l+1$. Since the parent node $o$ is concrete, i.e., $|o.\gS|=o.l$, we have $|o'.\gS|=|o.\gS|+1=o.l+1=o'.l$, which proves $o'$ is a concrete node.
\end{proof}

\begin{lemma}[Finite Propagation] \label{lem:finite-propagation}
Given $l_s\in\set{0,\dots,\eps}$, any branching propagation process starting from $\Theta_{l_s}$ has at most $(\eps-l_s)+(V-l_s)$ branching depth, which is finite.
\end{lemma}
\begin{proof}
We discuss in two cases:
\begin{enumerate}
\item If $l_s\in\set{1,\dots,\eps}$, then \Cref{algo:verify} starts with $\Theta_{l_s}$, whose progress measure $\mu(\Theta_{l_s})$ equals $(\eps-l_s, V)$. By \Cref{lem:branching-progress}, it strictly decreases along the super branching propagation process, so it has at most $(V-l_s)$ super branching steps away from level-$l_s$ concrete nodes, whose progress measure is $(\eps-l_s,l_s)$. This strictly decreases along concrete branching propagation process according to \Cref{lem:branching-progress}, and since the rest of the branching propagation process only includes concrete branching and the search tree consists only concrete nodes (\Cref{lem:concrete-propagation}), level-$l_s$ concrete nodes have at most $(\eps-l_s)$ concrete branching steps away from finally reaching the unbranchable concrete nodes (the leaves).
\item If $l_s=0$, then \Cref{algo:verify} starts with $\Theta_0$, which is a branchable concrete node, whose progress measure $\mu(\Theta_0)$ equals $(\eps, 0)$. By \Cref{lem:concrete-propagation}, the whole branching propagation process only includes concrete branching and the search tree consists only concrete nodes. By \Cref{lem:branching-progress} the progress measure strictly decreases along the concrete branching propagation process, so it has at most $\eps$ concrete-branching steps away from finally reaching the unbranchable concrete nodes (the leaves).
\end{enumerate}

From the above two cases, we prove the lemma.
\end{proof}

\relcompleteness*

\begin{proof}

By \Cref{lem:finite-propagation}, we know Algorithm~\ref{algo:verify} finally terminates and every unbranchable node at level $\epsilon$ is a concrete node. If network $f$ is robust, the MILP solver returns infeasible for each unbranchable node. That is to say, all unbranchable nodes are safe.
Moreover, by Corollary~\ref{cor:sound-branching}, the initial node $\Theta_l$ is proven safe as all its child nodes are safe. Namely, $\Tool{f, \nomvx, \epsilon}$ returns \texttt{Verified}.

\end{proof}

%% file: appendices/impl.tex
\section{Technical Details and Optimization} \label{sec:impl}

\subsection{Super Branching Policy for Initial Nodes}\label{sec:cvd}
For initial node at level $l>0$, $\gR^{(l)}[|\Theta_l.\gS|]$ (i.e., $\gR^{(l)}[V]$) is special. We either assign $\gR^{(l)}[V]$ to $l$ (especially for small $l$, e.g., $l\le2$), i.e., generate child nodes with their focus sets coming from $\binom{\gV}{l}$, or pick the best \cvd~\cite{coverd} (especially for larger $l$, e.g., $l\ge3$), which outperforms pure covering design~\cite{calzone} (\Cref{sec:covering-bab}).

\subsection{Detailed Verification Planning Technique}\label{sec:plan-detail}
Given $\hmP,\hmT$, the number of unknown-parents-only concrete nodes for each level $\tvn\in\sR^{(\epsilon+1)}$ can be estimated as follows:
\begin{equation}\label{eq:nl}
    \tvn_l=
        \begin{cases}
            1, & \text{if } l=0,\\[4pt]
            \left\lceil \binom{V}{l} \cdot \left(1-\hmP_{l-1,l-1}\right)^2 \right\rceil, & \text{if } l>0.
        \end{cases}
\end{equation}
Therefore, the per-level estimated duration on bounding those nodes can be computed by $\tvt_l=\hmT_{l,l}\cdot\tvn_l$. However, \Eqref{eq:nl} is not accurate, but empirically works.

We provide a more detailed version of \Cref{algo:plan-detailed} for choosing the best verification plan.
\begin{algorithm}[H]
\caption{Choose Best Plan}
\label{algo:plan-detailed}
\begin{algorithmic}[1]
\Procedure{\textsc{ChooseBestPlan}}{$f,\nomvx,\epsilon$}
\Statex \textbf{Input:} A classifier $f$, a nominal input image $\nomvx\in \gD$, and a maximum number of perturbed pixels $\epsilon$.
\Statex \textbf{Output:} The level $l_s$ to start verification, and a super branching strategy $\gR$.
\State $t_{min}\gets\infty$
\State $(l_s,\gR) \gets (0,\emptyset)$
\State $(\hmP,\hmT)\gets\texttt{SampleNodesOfAllShape}(f,\nomvx,\epsilon)$
\ForAll{$l\in\set{0,\dots,\eps}$}
    \State $t_{tail} \gets \sum_{i=l+1}^{\epsilon}{\tvt_i}$
    \State $t_{\texttt{con}} \gets \hmT_{l,l} \cdot \binom{V}{l} + t_{tail}$
    \If{$t_{\texttt{con}} < t_{min}$}
        \State $t_{min} \gets t_{\texttt{con}}$
        \State $(l_s,\gR[V]) \gets (l,l)$
    \EndIf
    \If{$l\in\set{3,4,5,6}$}\Comment{\text{all }\cvd\text{supported }$t$}
        \State $(\gR^{(l)},\gG^{(l)},t_{\texttt{sup}}^{(l)}) \gets $ \texttt{GetRefinementStrategyAndBestCVD}$(\hmP_{:,l}, \hmT_{:,l},l)$
        \If{$t_{\texttt{sup}}^{(l)} + t_{tail} < t_{min}$}
            \State $t_{min} \gets t_{\texttt{sup}}^{(l)} + t_{tail}$
            \State $(l_s,\gR) \gets (l,\gR^{(l)})$
            \State $\gR[V] \gets \gG^{(l)}$
        \EndIf
    \EndIf
\EndFor
\State \Return{$(l_s, \gR)$}
\EndProcedure
\end{algorithmic}
\end{algorithm}

\paragraph{Fitting the safe rates and bounding times.}
At each level $l$, the planner samples super nodes across all focus-set sizes $k \ge l$ (Line~4 of \Cref{algo:plan-detailed}).
Let $n_{k,l}$ denote the number of sampled nodes of size $k$, and $s_{k,l}$ the number proven safe.
Directly using the empirical safe rate $s_{k,l}/n_{k,l}$ introduces two problems.
First, whenever all sampled nodes of size $k$ are verified safe, the empirical rate saturates at $1$.
The planner then assumes size-$k$ super nodes never require further branching and selects the super branching strategy among such sizes (Line~12) purely by measured bounding time.
Because these runtimes are nearly identical, minor timing noise ends up dictating the branching plan, causing run-to-run variance on the exact same instance.
Second, due to sampling variance, the raw estimates are not guaranteed to be monotonic in $k$, even though larger focus sets admit strictly more perturbations and should be harder to verify.

To address both issues, we smooth the safe-rate estimates at each level $l$ by fitting a logistic curve in $\log k$~\cite{cox1958regression}:
\[
  \hmP_{k,l} = \sigma(a_l + b_l \log k), \qquad b_l \le 0,
\]
where $\sigma$ is the standard logistic function.
We estimate $a_l$ and $b_l$ via binomial maximum likelihood over all sampled sizes, with $n_{k,l}$ trials and success probability $\hmP_{k,l}$~\cite{mccullagh1989glm}.
The constraint $b_l \le 0$ ensures the rate is non-increasing in $k$, and $\hmP_{k,l}$ remains strictly in $(0,1)$ unless all samples at level $l$ are uniformly safe or unsafe.
Empirically, the curve matches the samples within binomial noise.
For concrete nodes (\Eqref{eq:nl}), we keep the direct empirical estimates, as each level has only a single concrete focus-set size and thus no curve to fit.
Similarly, we model the per-node bounding time linearly as $\hmT_{k,l} = c_l + d_l k$ ($d_l \ge 0$) using the robust Theil--Sen estimator~\cite{theil1950rank,sen1968estimates}.
To avoid skewing the regression, we exclude the first sampled size (which incurs GPU warm-up overhead) as well as any size that exhausts fewer nodes than the full sampling budget (where per-node latency is dominated by batch overhead).
Both models are fitted using SciPy~\cite{virtanen2020scipy}.

\subsection{Detailed Concrete Branching and Pruning Technique in Practice}\label{sec:sap}
In practice, we perform concrete-node branching and pruning in batches.
Rather than being expanded one by one, all unknown concrete nodes at the same level are collected and branched together, then pruned together.
\paragraph{Canonical concrete branching.} We refer to the practical concrete branching as the \emph{canonical concrete branching}. We represent each concrete node by its focus set in canonical increasing order, i.e., a level-$l$ node is stored as a tuple
$(v_1,\ldots,v_l)$ with $v_1<\cdots<v_l$.
This allows us to generate children by canonical augmentation~\cite{knuth2005generating,ruskey2003combinatorial}: instead of appending every pixel in $\gV\setminus\gS$, we append only pixels larger than the last element of the tuple.
For a concrete node's focus set $\gS=(v_1,\ldots,v_l)$, the batched concrete split therefore generates
\[
    \set{(v_1,\ldots,v_l,v):~v_l < v \le V},
\]
with the convention $v_0:=0$ when $l=0$.
This produces each $(l+1)$-subset exactly once, through its unique canonical predecessor given by its first $l$ elements, and avoids duplicate children that would otherwise arise from independently taking the full upper shadow of each unknown node.

Concretely, let $U_l\in\mathbb{N}^{n_l\times l}$ store the $n_l$ unknown concrete nodes at level $l$ in increasing order.
For each row $U_l[i]$, we compute the number of valid extensions as
$V-U_l[i,l]$, where $U_l[i,l]$ denotes the last element of the tuple.
The implementation then repeats each parent row according to this number and appends the corresponding range of larger indices.
This gives a vectorized implementation of canonical concrete branching on GPU, avoiding explicit Python loops over nodes or children.

\paragraph{Pruning concrete child nodes.} These generated concrete child nodes are then pruned given the collection of unknown concrete nodes $\UNs$ at level $l$, as shown in~\Cref{algo:prune}. Specifically, for each child node $o'$ at level $l+1$, we check how many subsets are in $\UNs$.
If fewer than $l+1$ of them are in $\UNs$, some parent of $o'$ is safe, so $o'$ is pruned.
In practice, this process is CUDA-accelerated.
\input{algorithms/Prune}

\subsection{Probe and Probe Thickness}\label{sec:probe}
At the beginning of the planning, we will first do probing before we do the sampling. Starting from $\Theta_0$, we check the worst bound level by level, while keeping track of a fixed number (\texttt{probe\_thickness}) of worst case unknown concrete nodes (i.e., least tight bound that $\Bound{o}<0$) which will be branched on to generate next level concrete nodes.

The goal is to check on which levels we will have unknown concrete nodes if we choose that level to start verification. If a level $l\in\set{3,4,5,6}$ (all \cvd~\cite{coverd} supported $t$) contains unknown concrete nodes, then \tool will place an upper bound (we set it to be 15) on the mean focus set size of $\gR^{(l)}[V]$ to avoid excessive steps for recursively super branching the parent super nodes of those unknown concrete nodes. When level gets deeper, some unknown super nodes cannot be detected via sampling due to the sheer volume of the combinatorial space if they are few, which results in inaccurate $\hmP$ that can fool the verification planner to select a best geometry (for constructing \cvd) where the mean focus set size is very large, and to compute a super branching strategy with dense refinement steps. This can result in an explosion of the redundant child super nodes, making the estimation of the overall verification time varies from the real case.

\subsection{A More Detailed Version for \Cref{algo:verify}}
Here we show a more detailed version which reflects \tool in practice. Note that \Cref{algo:verify-detailed,algo:VerifyUnknownNodes} instantiate \Cref{algo:verify} with a level-synchronous worklist, batched bounding, canonical augmentation, and cross-parent pruning.

\begin{figure}[!htbp]
\begin{algorithm}[H]
\caption{Verify the \lo local robustness for neural network classifier $f$}
\label{algo:verify-detailed}
\begin{algorithmic}[1]
\Procedure{\tool}{$f,\nomvx,\epsilon$}
\Statex \textbf{Input:} A classifier $f$, a nominal input image $\nomvx$, and a perturbation budget $\epsilon$
\Statex \textbf{Output:} \texttt{Verified} if the verification is successful, an adversarial example otherwise
\State $(l_s,\gR^{(l_s)}) \gets$ \Call{ChooseBestPlan}{$f,\nomvx,\epsilon$}
\State $\gO\gets\set{\Theta_{l_s}}$ \Comment{Enqueue an initial node at level $l_s$}
\State $\UNs \gets \emptyset$
\While{$\gO \ne \emptyset$}
    \State $o \gets\gO.\texttt{dequeue}()$
    \If{$\Bound{o}\le0$}
        \If{$|o.\gS| > o.l$}
            \State $\gO\gets\gO\cup\sBranch{\gR^{(l_s)}[|o.\gS|]}{o}$
        \Else \Comment{Collect unknown concrete nodes at level $l_s$}
            \State $~\UNs\gets\UNs\cup\set{o}$
        \EndIf
    \EndIf
\EndWhile
\State \Return \Call{VerifyUnknownNodes}{$f,\nomvx,\epsilon,\UNs$}
\EndProcedure
\end{algorithmic}
\end{algorithm}
\end{figure}

\input{algorithms/VerifyUnknownNodes}

%% file: algorithms/Prune.tex
\begin{figure}[!htbp]
\begin{algorithm}[H]
\caption{Prune those concrete child nodes with any safe parents}\label{algo:prune}
\begin{algorithmic}[1]
\Procedure{\textsc{Prune}}{$\UNs, \CNs$}
\Statex \textbf{Input:} A list of all unknown concrete nodes $\UNs$ from the current level and a list of child concrete nodes $\CNs$ at the next level (generated by $\UNs$).
\Statex \textbf{Output:} All concrete nodes $\CNs{}'$ at the next level without any safe parents, i.e., all their parents are from $\UNs$.
\State $\CNs{}'\gets\emptyset$
\ForAll{$o\in\CNs$}
    \If{$|\set{o'\in\UNs:o'.\gS\subset o.\gS}|=o.l$} \Comment{$o.l=o'.l+1$}
        \State $\CNs{}'\gets\CNs{}'\cup \set{o}$
    \EndIf
\EndFor
\State \Return{$\CNs{}'$}
\EndProcedure
\end{algorithmic}
\end{algorithm}
\end{figure}

%% file: algorithms/VerifyUnknownNodes.tex
\begin{figure}[!htbp]
\begin{algorithm}[H]
\caption{Verify the next level concrete nodes without any safe parents}
\label{algo:VerifyUnknownNodes}
\begin{algorithmic}[1]
\Procedure{\textsc{VerifyUnknownNodes}}{$f,\nomvx,\epsilon,\UNs$}
\Statex \textbf{Input:} A classifier $f$, a nominal input $\nomvx\in \gD$, a perturbation budget $\epsilon$, and all unknown concrete nodes $\UNs$ from the current level.
\Statex \textbf{Output:} \texttt{Verified} if the verification is successful, an adversarial example otherwise
\If{$\UNs=\emptyset$} \Return{\texttt{Verified}} \EndIf
\If{$\UNs_1.l=\epsilon$}
    \ForAll{$o\in\UNs$}
        \State $(\texttt{feasible},\vx^{\texttt{adv}})\gets \texttt{MILP}\left(f, \eoball\right)$
        \If{$\texttt{feasible}$} \Return $\vx^{\texttt{adv}}$ \EndIf
    \EndFor
    \State \Return{\texttt{Verified}}
\EndIf
\State $\CNs\gets\underset{o\in\UNs}{\bigcup}\cBranch{o}$ \Comment{Canonical concrete branching applies}
\State $\CNs\gets\Call{Prune}{\UNs,\CNs}$\Comment{$\CNs_1.l=\UNs_1.l+1$}
\State $\UNsp \gets \set{o^\prime\in\CNs:~\Bound{o^\prime}\le0}$ \Comment{ $\UNsp_{1}.l=\UNs_1.l+1$}
\State \Return \Call{VerifyUnknownNodes}{$f,\nomvx,\epsilon,\UNsp$}
\EndProcedure
\end{algorithmic}
\end{algorithm}
\end{figure}

%% file: appendices/hyperparameters.tex
\section{Hyper-parameters} \label{sec:hyperparameters}

The hyper-parameters of \tool are as follows: the number of GPU workers is set to 4, and the number of MILP workers to 5 (except for bcv5, where we use 50 for consistency with \coverd~\cite{coverd}).
The maximum number of sampled super nodes for each $k$-sized focus set at each level is set to 200 per GPU worker (in total 800), except for bcv4 it's 100 per GPU worker (in total 400). The upper bound of $k$ is set to 200 (same as \coverd).
Super nodes are stored in a batch pool before batched bounding. The batch size for bounding super nodes are set to 800 for fully connected networks and 400 for CNNs (5 for bcv4). The size of the super node pool for each focus-set size are set to 2 times (4 times for bcv4) the super batch size (the batch size to bound super nodes), which is 1,600 for fully connected networks and 800 for CNNs (20 for bcv4).
For concrete nodes, the maximum number of samples per level is set to 5,000 per GPU worker (in total 20,000) for fully connected networks and 500 per GPU worker (in total 2,000) for CNNs (in total 400 for bcv4). The concrete batch size is set to 4,000 for fully connected networks and 200 for CNNs (5 for bcv4).
The probe thickness (at most how many worse concrete nodes to keep track of at each level during probing; see \Cref{sec:probe}) is set to 50 for fully connected and 5 for CNNs.
Finally, the timeout is set to 1800 seconds (3600 for bcv4).

The base seed defaults to $42$, which is the same as in \coverd. And we keep the same \emph{per instance seed} settings, too, where for every instance (at index $i$), the running seed will be set to $42+i$.

Note that if the estimated duration is larger than twice the timeout, we regard the instance as timed out and skip the verification to save time. This heuristic can only ever cost a verified instance, so the reported numbers are conservative.

%% file: appendices/extended-eval.tex
\section{Supplementary Evaluation} \label{sec:extended-eval}

\subsection{Details of Benchmarks}\label{sec:benchmark-details}

\paragraph{The \boldmath\lo adversarial training.} Adversarially trained networks are frequently used in robustness verification. Some adversarially trained networks in prior work \cite{coverd,calzone} are trained under $\ell_\infty$ settings (against PGD attack \cite{pgd}). We include networks adversarially retrained under \lo settings in our experiment, which are generally easier to be verified safe within the time limit by \lo robustness verifier. Note that the certified adversarial defense training (as referenced above) is only applicable to fully connected networks.

\paragraph{Why the two exception networks.} We report bcv4 and bcv5 only at $\epsilon=3$. These two networks are challenging for different reasons. The bcv4 network is substantially larger and deeper than the other benchmarks, as reflected by its number of neurons in \Cref{tab:main}. Consequently, its concrete margin lower bounds are looser than those of shallower networks, and \tool already relies heavily on MILP solving even at $\epsilon=3$. The bcv5 network is not significantly larger than the other benchmarks, but it is evaluated on multi-channel CIFAR-10 inputs. Since we consider pixel-wise $\ell_0$ perturbations, $\epsilon=3$ allows modifying three pixels, corresponding to $3\text{ pixels} \times 3\text{ channels}=9$ input features. This also leads to loose bounds and substantial reliance on MILP solving even at $\epsilon=3$. Therefore, we restrict both bcv4 and bcv5 to $\epsilon=3$ in the evaluation (same as prior work~\cite{calzone,coverd,shapira2026tight}). We use a timeout of $3600$ seconds for bcv4 due to its size, and $1800$ seconds for all other networks.

\begin{table}[t]
    \centering
    \caption{Main results of \tool on pixel-wise \lo robustness verification. For each benchmark and perturbation budget $\epsilon$, we report the number of correctly classified images, candidate instances, verified instances, timed out instances, mean runtime on verified instances, starting level statistics and number of instances where concrete branching applied.}
    \label{tab:extended-main}
    \input{figures/fir-result6}
    \vspace{2pt}
    \caption*{
    \footnotesize
    \textbf{Note.}
    We use a timeout of $3600$ seconds for bcv4 and $1800$ seconds for all other networks.
    \textbf{Img} is the number of evaluated images, and \textbf{CC} is the number of correctly classified images.
    \textbf{Cand.} denotes the number of verification candidate instances at the target budget $\epsilon$, after excluding images for which an adversarial example within the budget is already known from \sigmao, \tool at smaller budgets, or any prior verifier.
    \textbf{R} is the number of verified images.
    $l_s<\epsilon$ counts verified images whose starting level is smaller than $\epsilon$, while $l_s=\epsilon$ counts verified images whose starting level equals $\epsilon$. $\branch^{\texttt{con}}$ counts, among the $l_s<\epsilon$ cases, how many use the concrete branching policy.
    }
    \vspace{-13pt}
\end{table}

\paragraph{Architectural preference in starting level.} For fully connected networks, there is a non-negligible fraction of instances that use concrete branching according to~\Cref{tab:extended-main}. However, for CNNs, almost all starting levels are $\epsilon$ and thus concrete branching is barely used. This is due to the sparsity of the matrix representation of the convolution operator, which results in sparse-weighted affine bounds when the layer-wise relaxation is backpropagated. Concretizing a sparse-weighted affine bound over an input space constrained by nodes whose level is less than $\epsilon$ can result in looser concrete bounds than those of non-sparse affine bounds.
This is because for a sparse-weighted affine bound, the contribution $\Delta$ is concentrated only on a few pixels (the size of the flattened convolutional kernel), i.e., only a few $\Delta_v$ are non-zero. Thus for a node $o$ where $o.l<\epsilon$, there is at least one perturbation budget that can select the most-contributing pixels across $\gV$, and it will select the non-zero ones, which is a large proportion compared to the extreme case -- perturbing all the pixels. Therefore, the verification planner will always pick $\epsilon$ as the starting level since the safe rates of the $\eps$-level nodes are significantly higher than other levels' due to their tighter bounds.

\subsection{Qualitative Examples}\label{app:qualitative}
\begin{figure}[t]
  \centering
  \includegraphics[width=\textwidth]{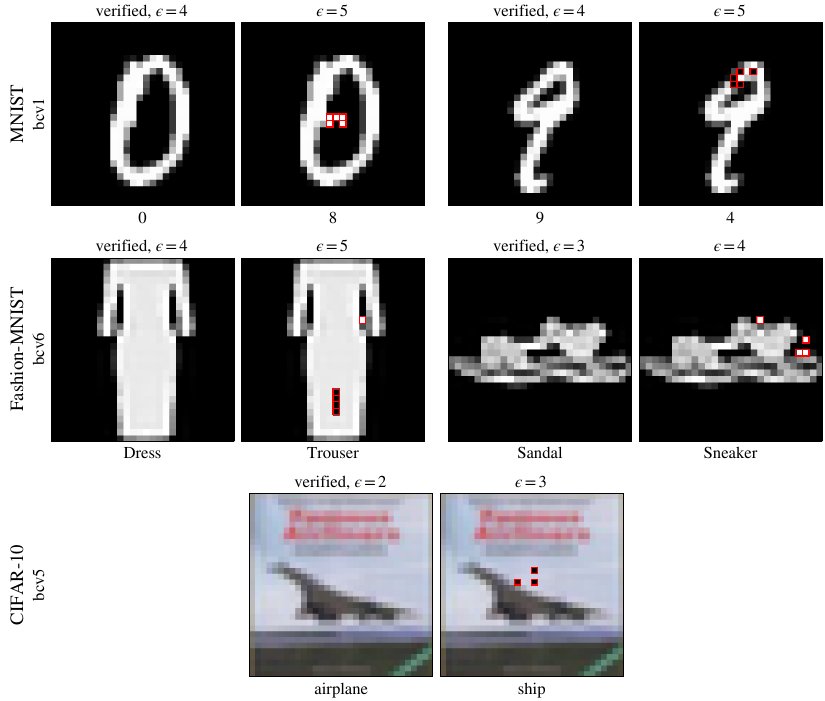}
  \caption{Qualitative examples. Each pair shows a correctly classified image that \tool verifies at budget $\eps$
    (left), and an adversarial example at budget $\eps+1$ with the perturbed pixels marked in red (right).}
  \label{fig:qualitative}
\end{figure}
\Cref{fig:qualitative} shows examples where \tool verifies a
correctly classified image at budget $\eps$, while an adversarial example exists
at $\eps+1$. Adversarial examples are found by the MILP fallback (pixel-wise) and
checked by a forward pass.

\subsection{Detailed Comparison with Baselines}\label{sec:baseline-detailed}
\begin{figure}
    \centering
    \def\svgwidth{0.32\textwidth}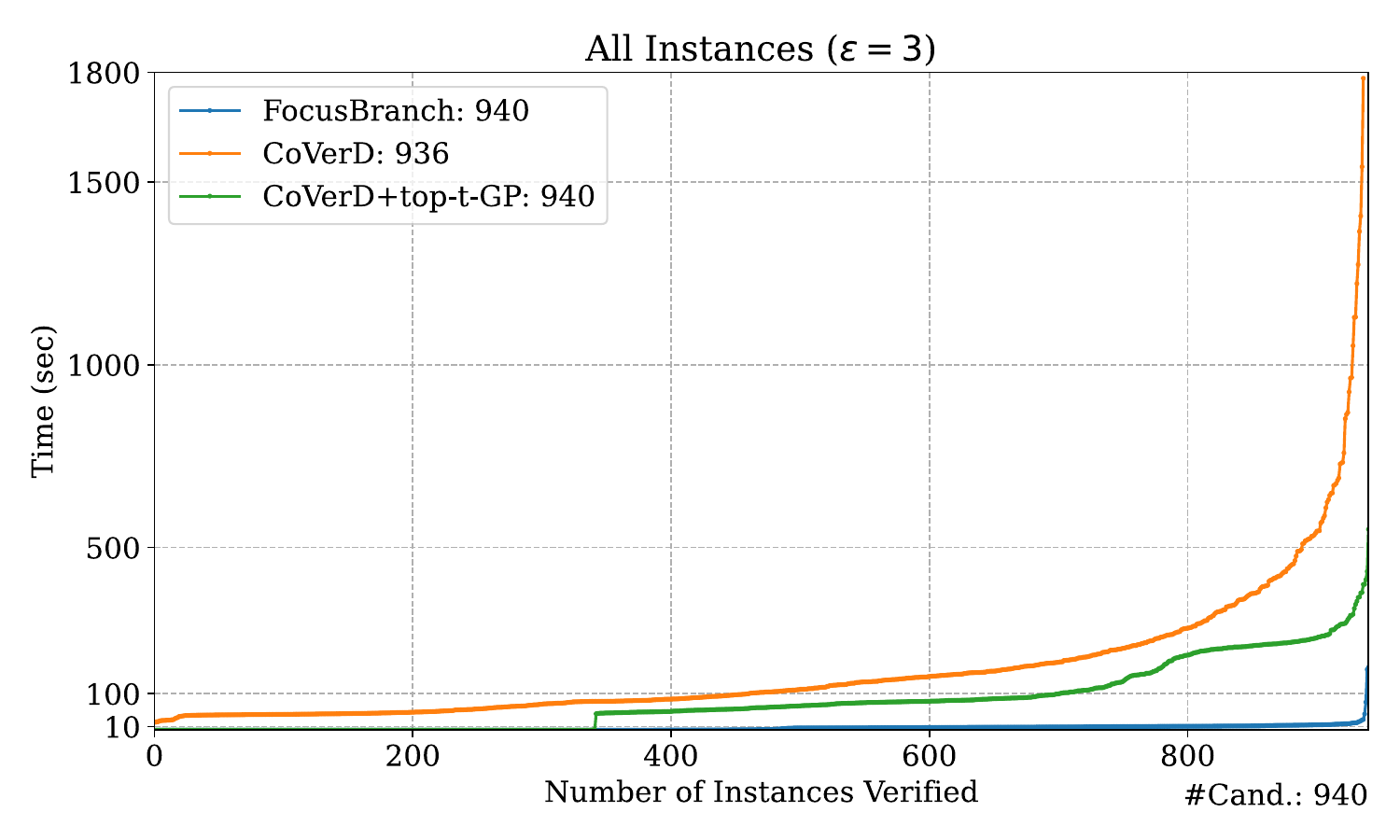
    \def\svgwidth{0.32\textwidth}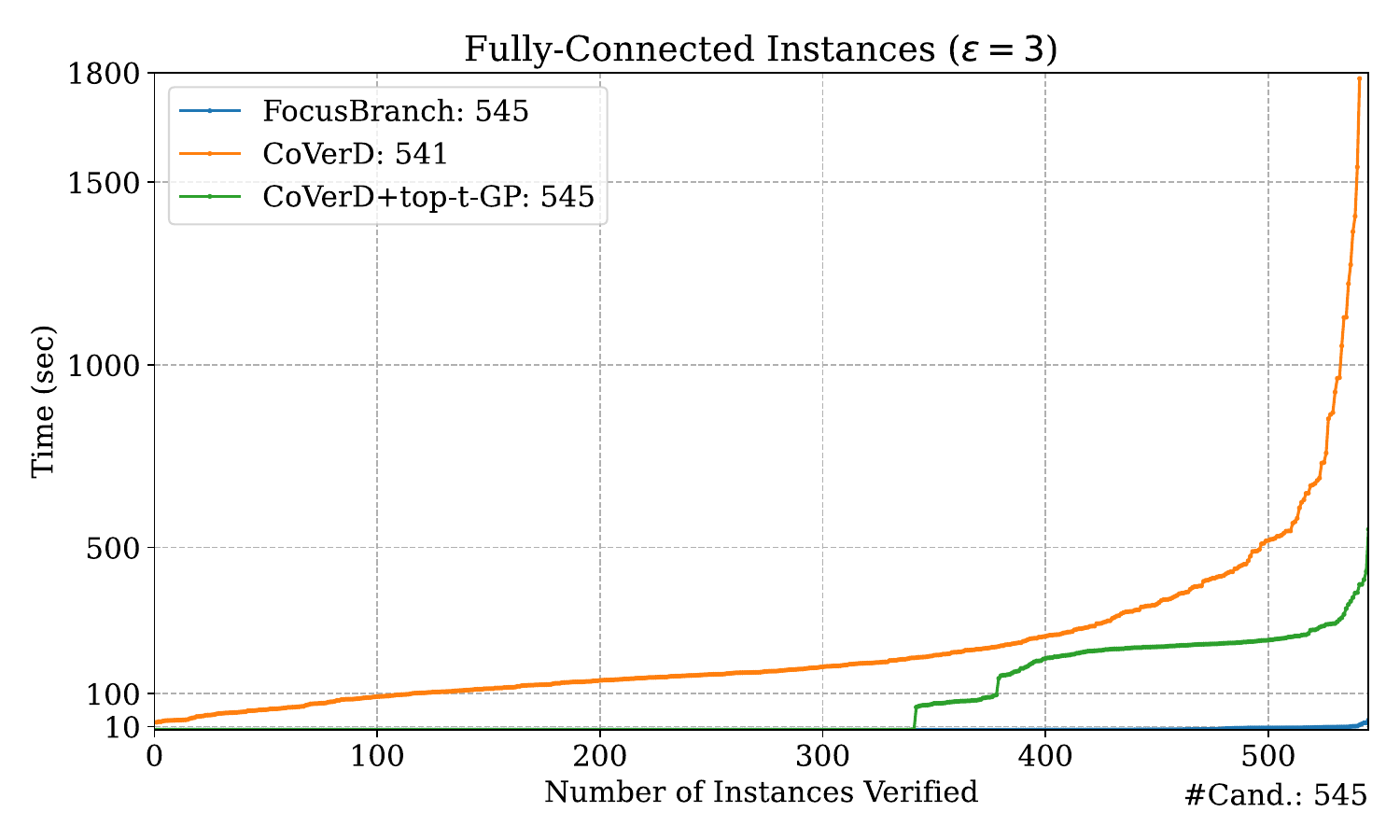
    \def\svgwidth{0.32\textwidth}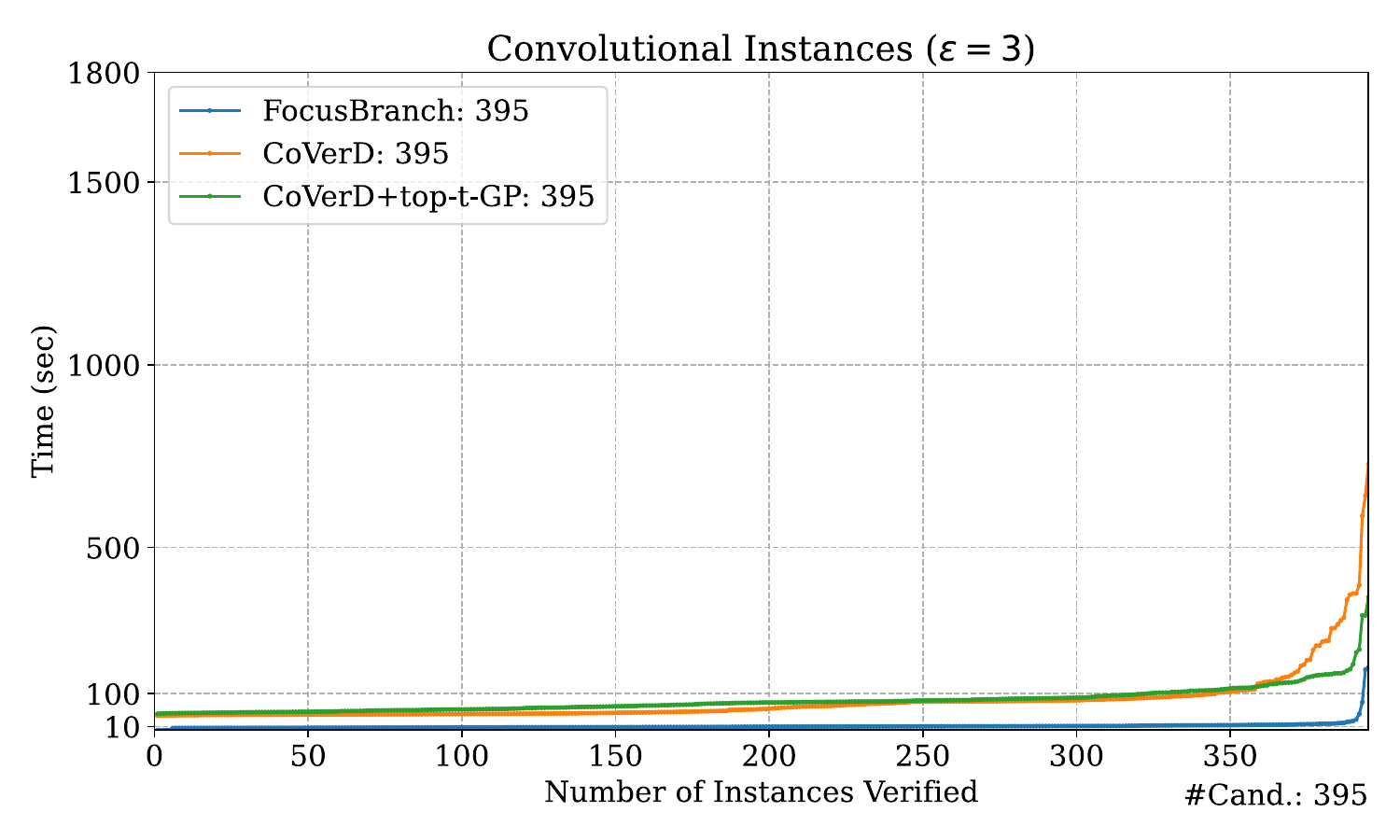
    \caption{Comparison between \tool, \coverd and \cvdtt at $\epsilon=3$.}
    \label{fig:comp-e3}
\end{figure}

\begin{figure}
    \centering
    \def\svgwidth{0.32\textwidth}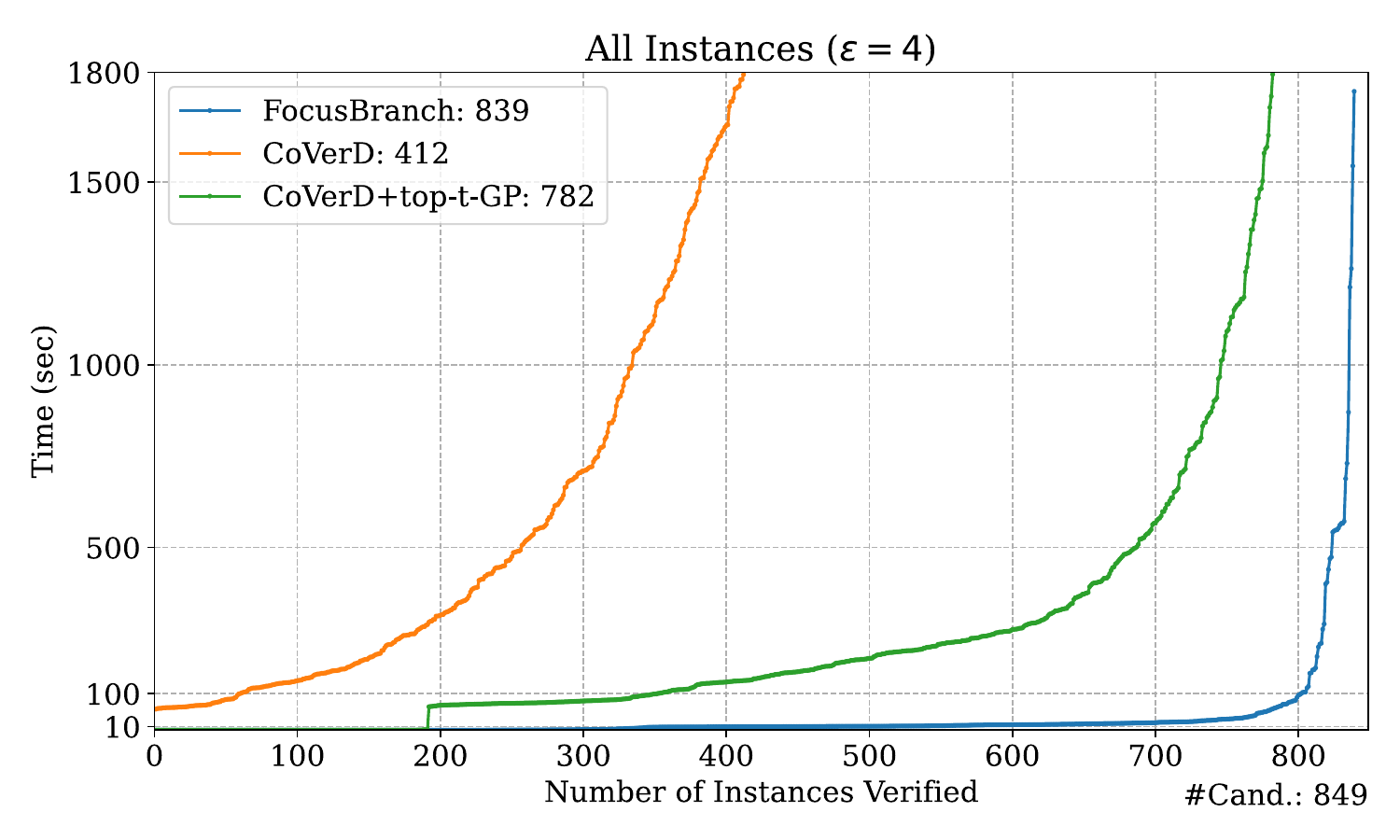
    \def\svgwidth{0.32\textwidth}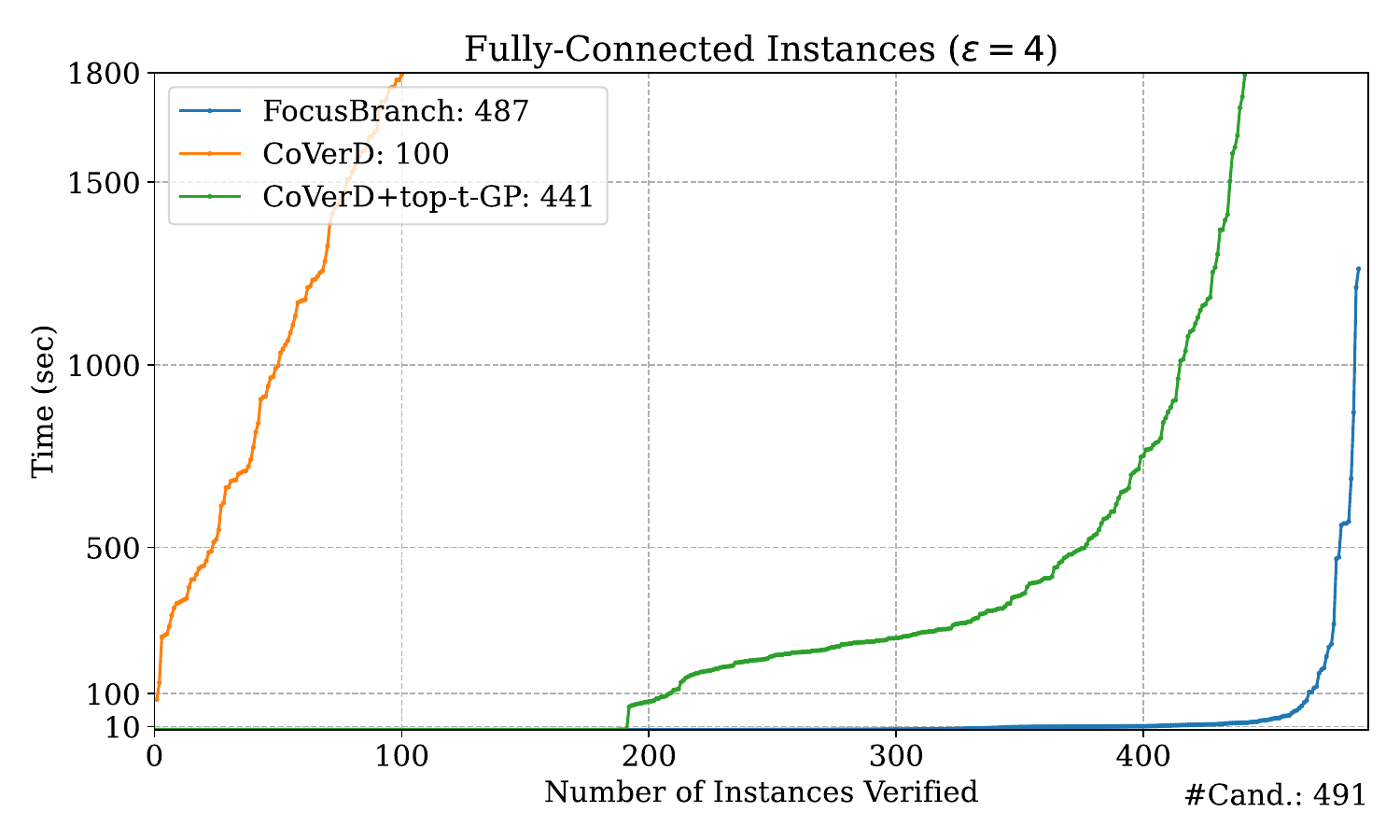
    \def\svgwidth{0.32\textwidth}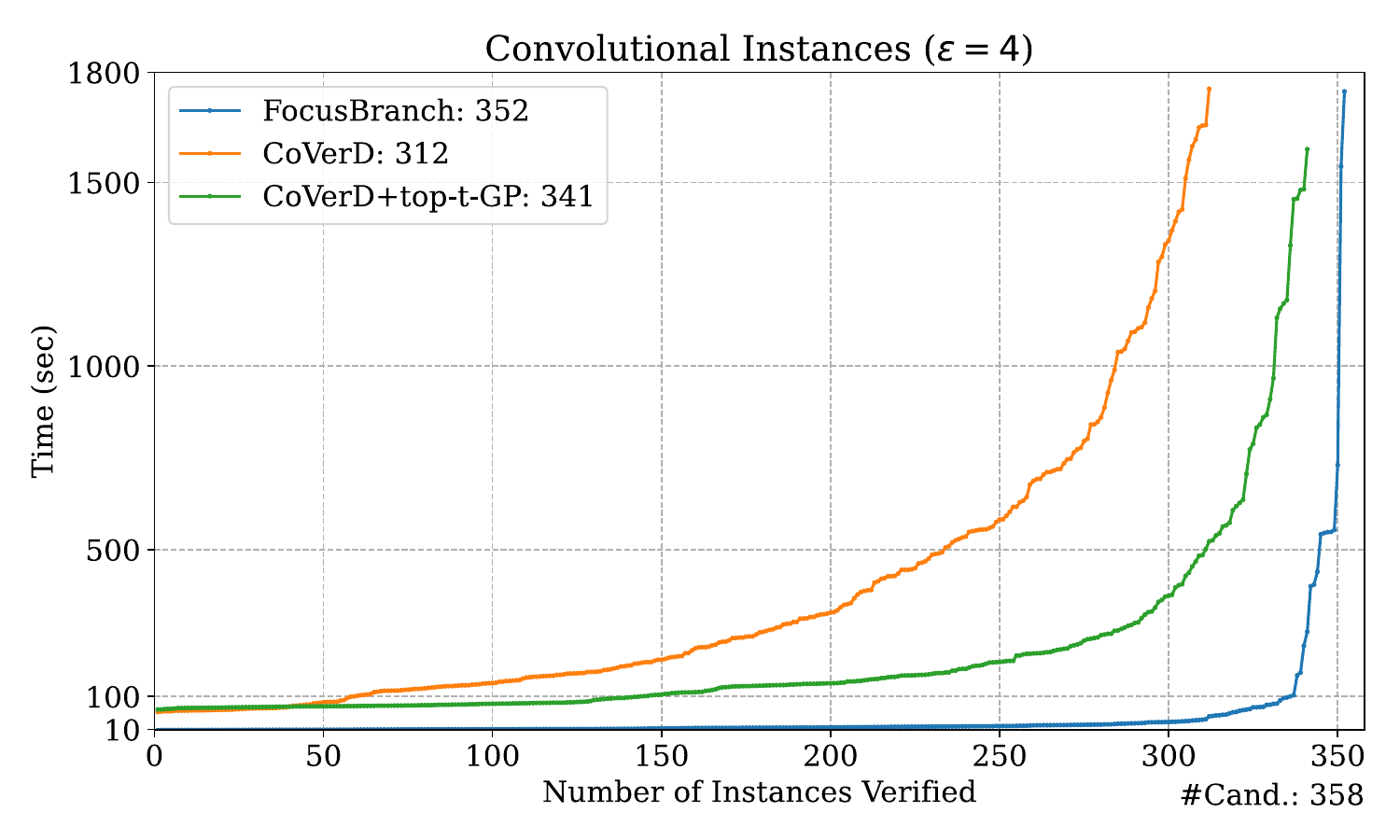
    \caption{Comparison between \tool, \coverd and \cvdtt at $\epsilon=4$.}
    \label{fig:comp-e4}
\end{figure}

\begin{figure}
    \centering
    \def\svgwidth{0.32\textwidth}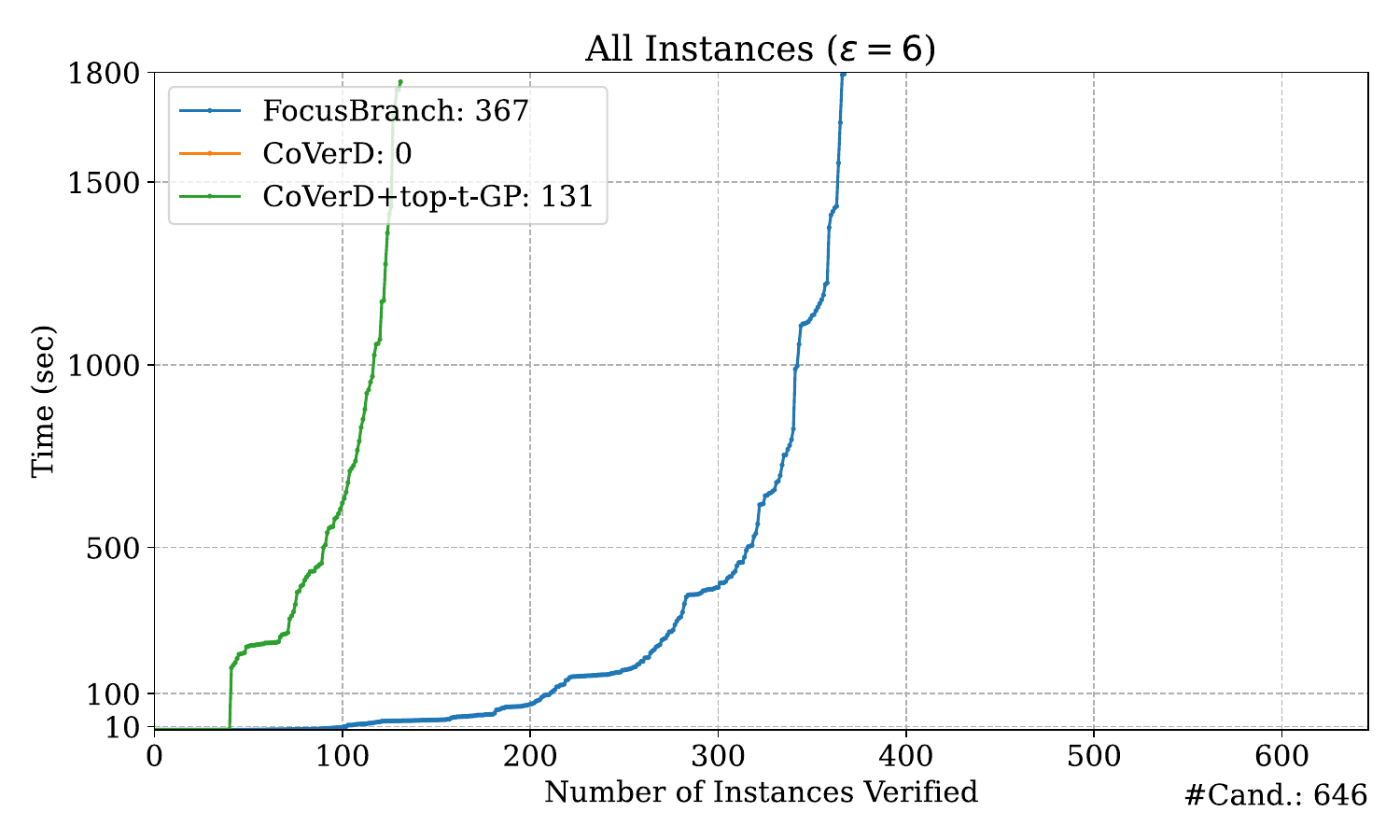
    \def\svgwidth{0.32\textwidth}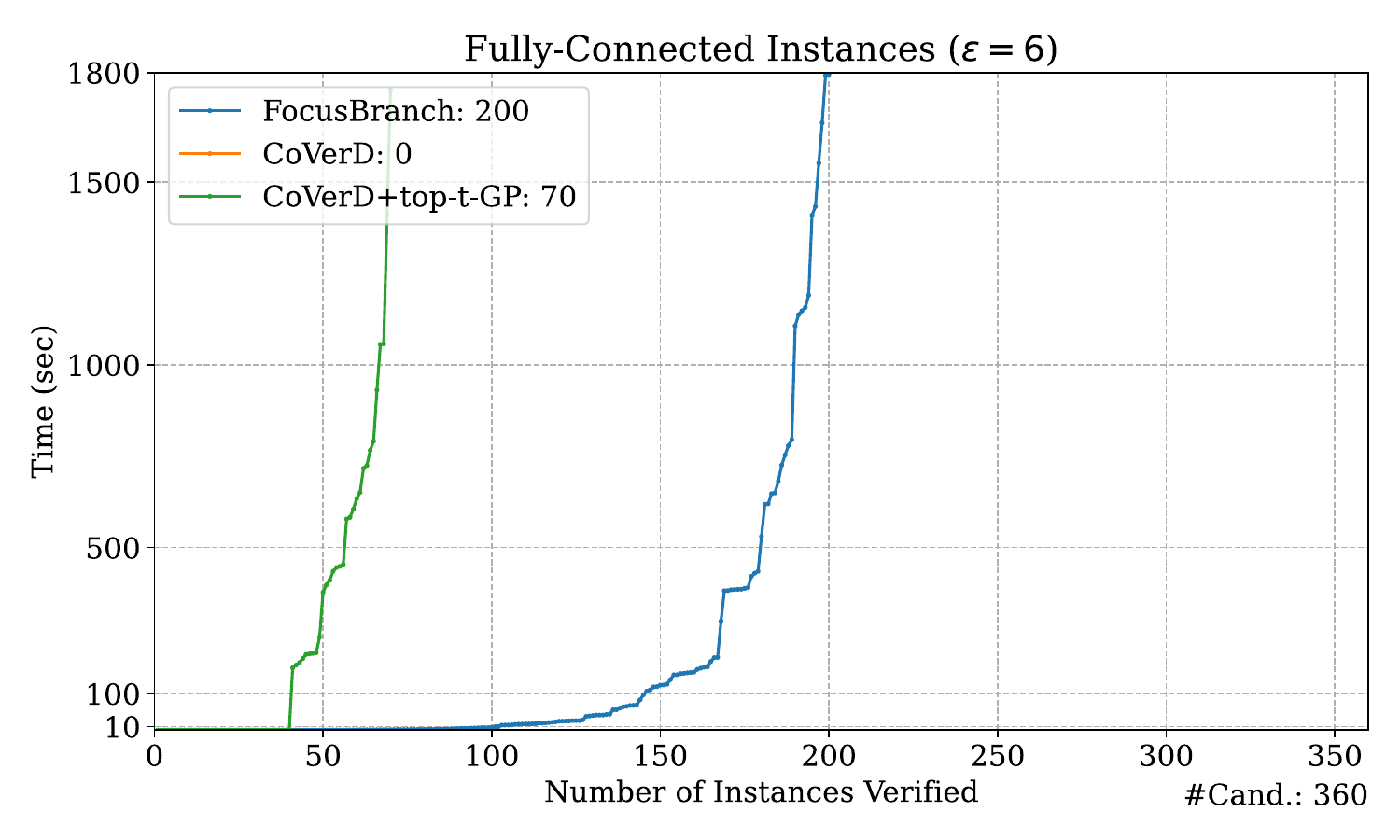
    \def\svgwidth{0.32\textwidth}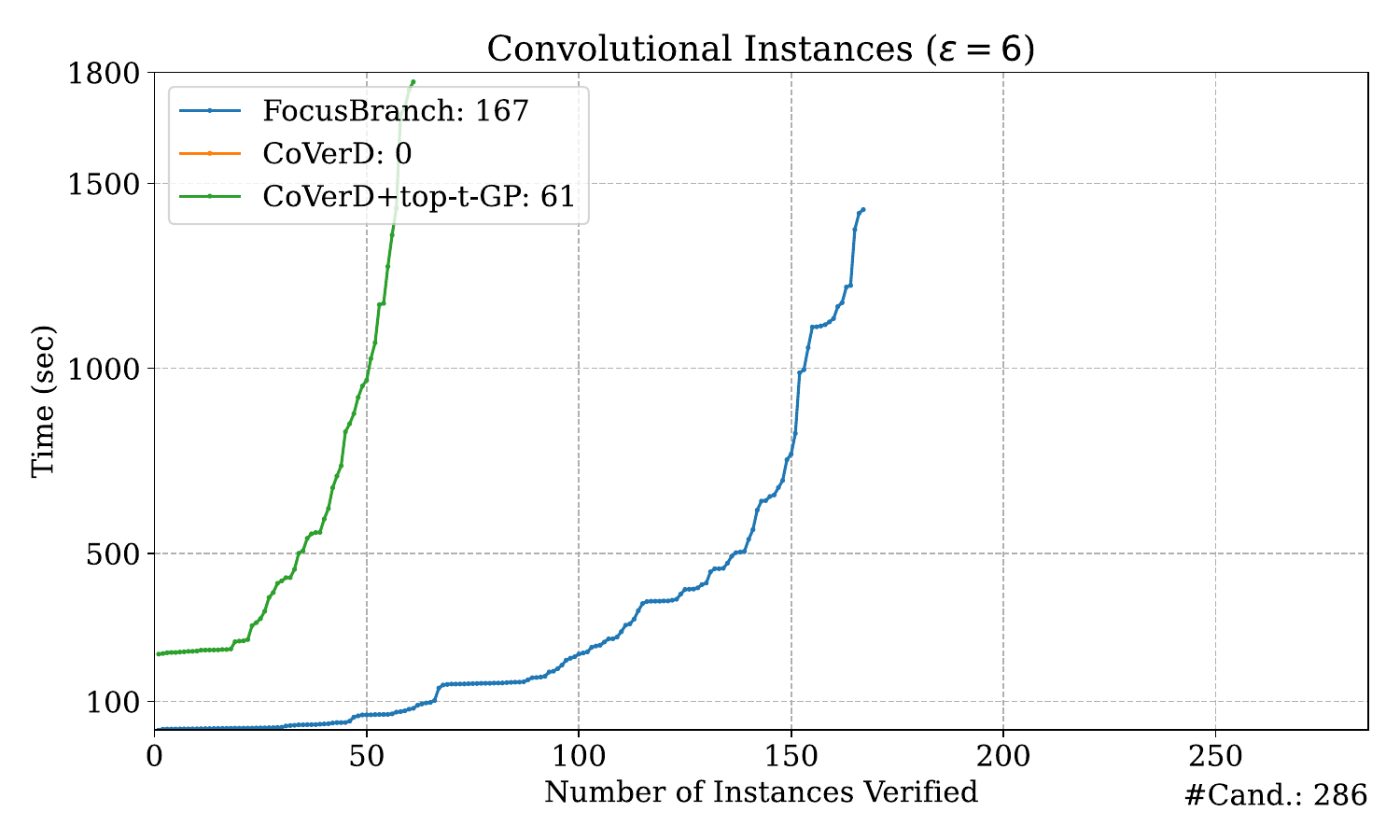
    \caption{Comparison between \tool, \coverd and \cvdtt at $\epsilon=6$.}
    \label{fig:comp-e6}
\end{figure}

Detailed results of comparison with baselines are shown as cactus plots at $\epsilon=3$ (\Cref{fig:comp-e3}), 4 (\Cref{fig:comp-e4}) and 6 (\Cref{fig:comp-e6}).

\Cref{tab:bcv45} shows the sample-wise comparison with \cvdtt on bcv4 and bcv5, which are excluded from the aggregated cactus plots. Note that we failed to run \cvdtt on bcv5 using the same number of 50 MILP instances as in~\cite{shapira2026tight}. Specifically, they ran experiments on a dual AMD EPYC 7713 server with 2TB RAM and eight A100 GPUs. Thus we borrow the statistics from their results, aggregate across seeds, translate verification time from minute-base to second-base, and manually timed out those verified but above 1800s. However, the bcv5 statistics borrowed are not directly comparable with \tool's bcv5 statistics, since the two ran on different hardware.

\input{figures/bcv45}

\subsection{Comparison with MILP} \label{sec:extended-eval-milp}
We also compared with a pure MILP solution by formulating the pixel-wise \lo constraint together with the method in~\cite{tjeng2019mip}, using Gurobi~\cite{gurobi} as the solver. Note that MILP failed to verify any convolutional instances at $\eps\in\set{3,4,5,6}$, and thus in \Cref{tab:comp-milp}, the MILP lines only reflect their results on fully connected networks.

\Cref{tab:comp-milp} shows that \tool verifies more fully connected instances than MILP while also being faster on average; moreover, unlike MILP, it remains effective on CNNs, substantially increasing the total number of verified instances across architectures.

\begin{table}[t]
\centering
\caption{Comparison between \tool and an MILP solution that aggregates $\epsilon\in\set{3,4,5,6}$. \tool\textsubscript{FC} denotes our results aggregating across all fully connected networks while \tool\textsubscript{all} denotes our results aggregating across architectures.}
\label{tab:comp-milp}
\begin{tabular}{ccc}
\toprule
Method & \#Verified & Average Time (s) \\
\midrule
MILP & 1245 & 58.7 \\
\tool\textsubscript{FC} & 1599 & 44.6 \\
\tool\textsubscript{all} & 2799 & 67.7 \\
\bottomrule
\end{tabular}
\end{table}

\subsection{Comparison with auto\_LiRPA}\label{sec:comparison-autolirpa}

We do not include auto\_LiRPA~\cite{lirpa} in this comparison. In our experiments, auto\_LiRPA reports \texttt{Verified} on some instances for which adversarial examples within the target budget are known, as found by MILP, \tool, \coverd, \cvdtt and \sigmao. Therefore, its reported verification results are not directly reliable for this benchmark setting. This unreliability stems from the implementation, not from the underlying method, which is sound.
With $\alpha$-CROWN disabled, the method is exactly the level-$0$ special case of \tool (\Cref{sec:cert-o}), whose bound is valid by \Cref{lem:bound-validity}, and enabling $\alpha$-CROWN only selects among valid ReLU relaxations.

In particular, auto\_LiRPA's $\ell_0$ concretization assumes every input entry lies in $[0,1]$, regardless of the input bounds it is given,\footnote{See \texttt{PerturbationL0Norm} in \url{https://github.com/Verified-Intelligence/auto_LiRPA/blob/5a098e8f9fb5786a428a024981d833d303921f2d/auto_LiRPA/perturbations.py\#L88-L128}.} whereas our networks can take standardized inputs.

\subsection{Additional Ablation Study}
\begin{figure}[!t]
    \centering
    \def\svgwidth{1\textwidth}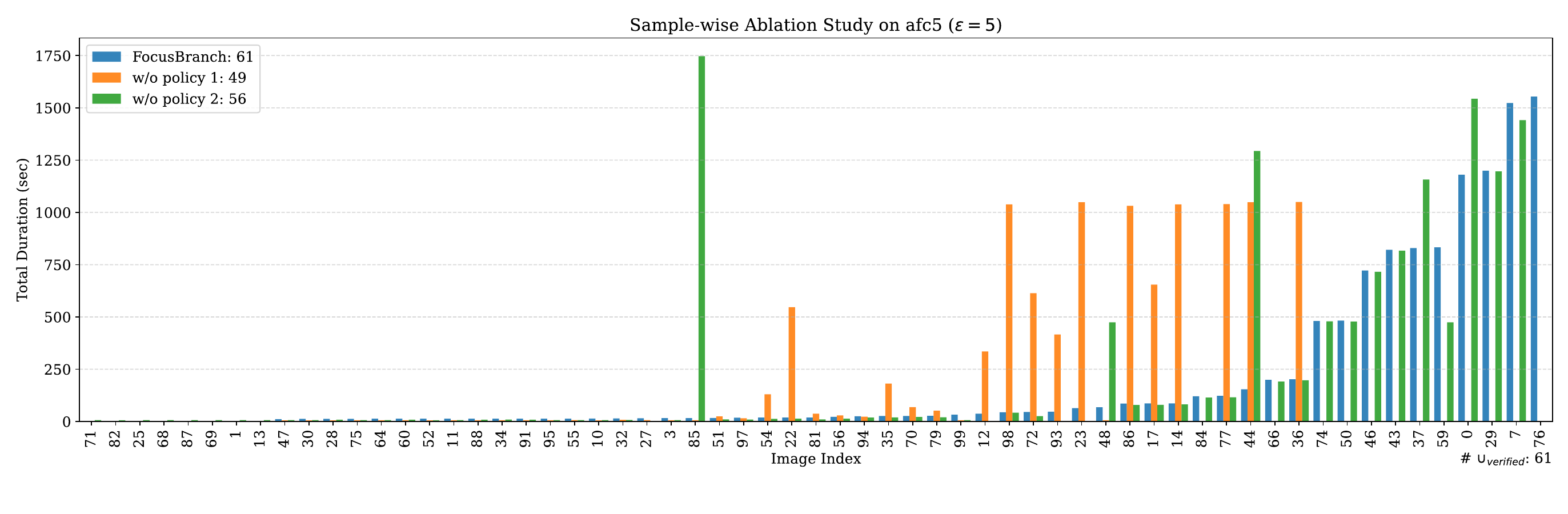
    \vspace{-7pt}
    \caption{Sample-wise ablation study on afc5 at $\epsilon=5$.}
    \label{fig:abl-afc5-samplewise}
    \vspace{-7pt}
\end{figure}

Instead of showing the aggregated view, we study the sample-wise discrepancy as shown in \Cref{fig:abl-afc5-samplewise}.

\begin{figure}
    \centering
    \def\svgwidth{0.45\textwidth}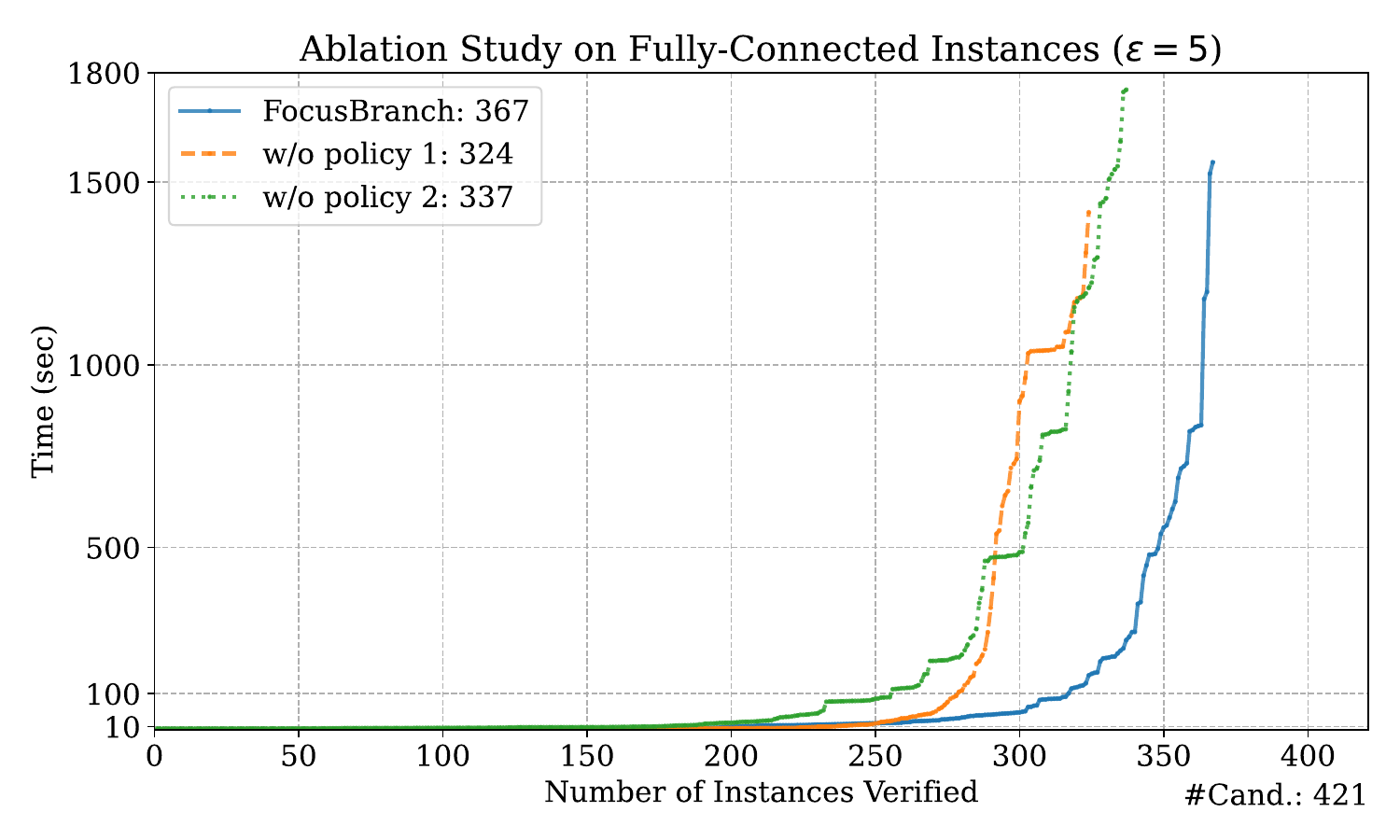
    \def\svgwidth{0.45\textwidth}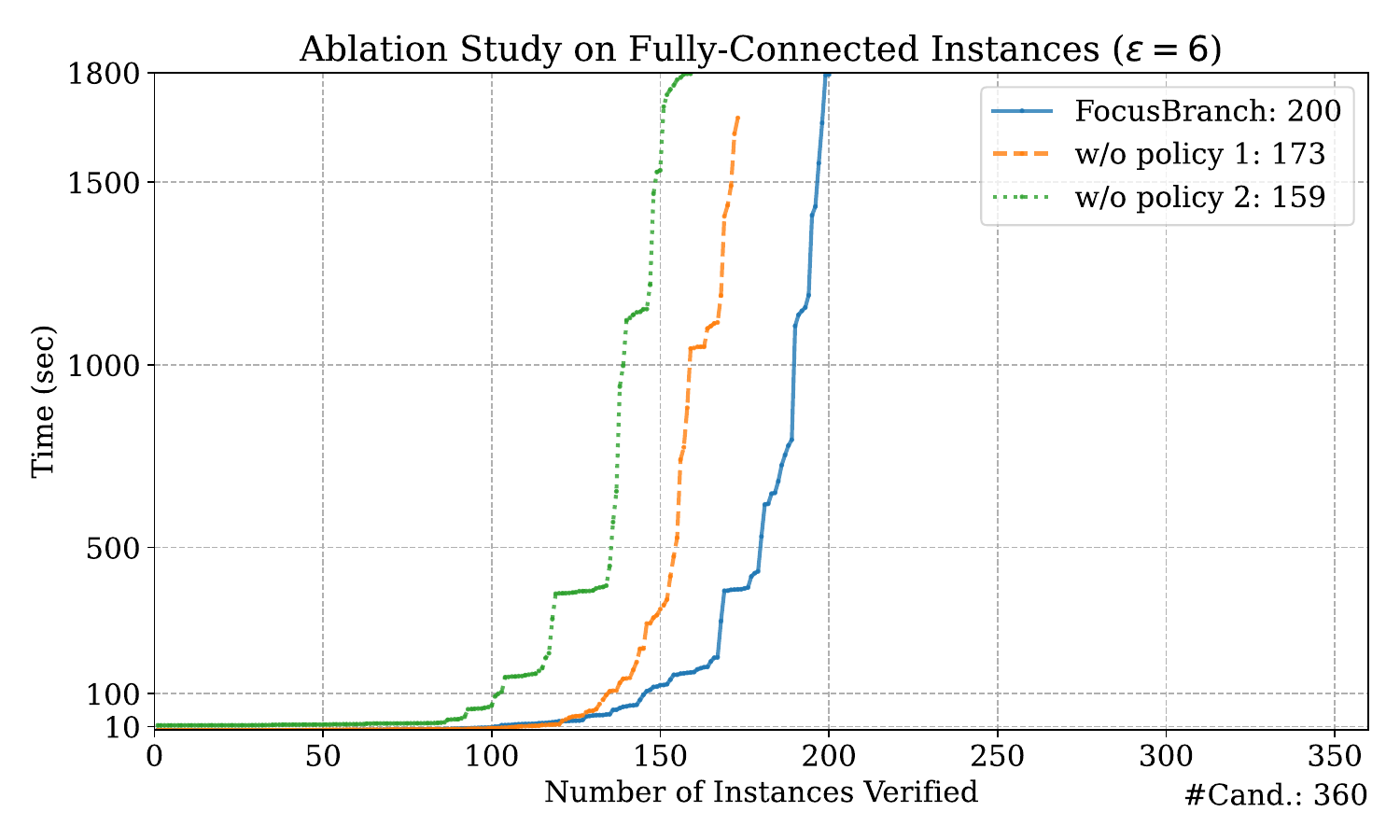
    \caption{Ablation study on fully connected instances at $\epsilon\in\set{5,6}$.}
    \label{fig:abl-fc-e56}
\end{figure}

We also performed ablation study for all fully connected instances at $\eps\in\set{5,6}$ as shown in \Cref{fig:abl-fc-e56}.

\subsection{Effect of Parallelization}
\label{app:parallelization-ablation}

Both \cvdtt and \tool parallelize (on several GPU workers), and batching is a routine GPU practice to increase the level of parallelization, on which we do not claim a contribution, as is disclosed in \Cref{sec:comparison} precisely for transparency.

To isolate the effect of parallelization from that of the algorithm, we rerun the experiment at $\eps\in\set{5,6}$ for \tool and \cvdtt (except for bcv4 and bcv5 we only run $\eps=3$), with all parallelization disabled (batch-size, workers are set to 1), under the same timeout and hardware with only one GPU. \tool also uses a super-node pool of size 1. As our sample sizes are set per GPU worker (\Cref{sec:hyperparameters}), \tool draws $4\times$ fewer super-node samples, and we also reduce its concrete-node samples per level to 400 for fully connected networks and 100 for CNNs (from 20000 and 2000), making its planning less reliable. \cvdtt would not be affected because it shares the workload with all GPU workers.

\input{figures/b1results}

As shown in \Cref{tab:main_results_b1}, turning off parallelization greatly slows down the speed for both methods, as implied by the decreased number of verified instances, which is expected. Without parallelization, both tools verify fewer instances, \tool more so as it also loses batching, yet \tool still verifies 432 instances against 352 and is $2.9\times$ faster on those both verify.
This advantage comes from the fully connected networks (292 vs.\ 197 verified, $5.5\times$), where almost all verified instances
start below level $\eps$, so one safe node with small level covers many $\epsilon$-level nodes, whereas \cvdtt always starts at the last ($\epsilon$-)level and must bound significantly more nodes. On afc1--afc4, \tool without parallelization even verifies more instances than \cvdtt with parallelization (268 vs.\ 201, \Cref{tab:main}).
On CNNs, almost all verified instances start at level $\eps$, where \cvdtt is a special case of \tool (\Cref{sec:comparison}), and the two tools are close (140 vs.\ 155 verified, $1.1\times$); the speedup of \tool on CNNs with parallelization ($11.5\times$ on the same rows) thus comes mainly from batching.

The 32 instances that only \cvdtt verifies are borderline, even though \tool verifies each of them within 2 minutes when parallelization is enabled. The outcome can change from run to run (holds for both tools): on afc5 at $\eps=5$, a repeated run with the same settings verified 18 instead of 15 instances for \tool, and 12 instead of 18 for \cvdtt, reversing the comparison. We attribute this variance to timing noise in the planner. Repeated runs of \tool almost always choose the same starting level and the same CVD for the initial node, and bound nodes at about the same speed, so the difference most likely lies in the rest of the super branching strategy. Both tools plan it from sampled bounding times, and when several focus-set sizes are almost equally fast, timing noise decides between them. The fits in \Cref{sec:plan-detail}, which these runs predate, make the plan of \tool deterministic (\Cref{app:reproducibility-note}).

\subsection{How Often Is MILP Invoked?}
\label{app:milp-usage}

Bounding in \tool is performed by incomplete bound propagation
(LiRPA/CROWN-style) with the closed-form concretization over node-constrained
$\ell_0$-balls of~\Cref{sec:cert-o} (\Eqref{eq:maxgx} and \Eqref{eq:mingx}).
MILP is used only as a fallback, on nodes that are simultaneously unbranchable
(level-$\epsilon$ concrete) and unknown. Table~\ref{tab:milp-usage} reports how
often this fallback is triggered across the full benchmark.

\begin{table}[h]
\centering
\caption{Share of verified instances that invoke the MILP solver at least once.}
\label{tab:milp-usage}
\begin{tabular}{cccc}
\toprule
$\epsilon$ & Verified & Instances invoking MILP & Share \\
\midrule
3 & 952 & 57 & 6.0\% \\
4 & 839 & 85 & 10.1\% \\
5 & 641 & 54 & 8.4\% \\
6 & 367 & 10 & 2.7\% \\
\midrule
All & \textbf{2799} & \textbf{206} & \textbf{7.4\%} \\
\bottomrule
\end{tabular}
\end{table}

MILP is invoked in $152/1599$ ($9.5\%$) of fully connected verified instances and
$54/1200$ ($4.5\%$) of convolutional ones; the remaining $92.6\%$ of verified
instances never call Gurobi, and $12$ of the $46$ network\,$\times\,\epsilon$
cells invoke it zero times. The speedups we report are therefore not the result
of accelerating an MILP solver.

\subsection{Choice of MILP Solver}\label{sec:milp-solver}

Like prior \lo verifiers~\cite{calzone,coverd,shapira2026tight}, \tool solves its fallback problems as MILPs~\cite{tjeng2019mip} with Gurobi~\cite{gurobi}. This is the standard choice for complete verification, and using the same solver as the baselines keeps the comparison fair. Since each fallback problem is posed over the input box of an unbranchable node, a complete verifier such as $\alpha,\!\beta$-CROWN~\cite{wang2021beta} could be used instead. To assess this alternative, we replaced Gurobi with $\alpha,\!\beta$-CROWN at $\eps=3$ on afc4, the smallest network in our benchmark, keeping all other settings unchanged. Using $\alpha,\!\beta$-CROWN was on average $10\times$ slower than using Gurobi. Our soundness and completeness results require only a sound (and complete) solver, so the fallback solver is interchangeable. Given this slowdown and the consistency with prior work, we keep Gurobi.

\subsection{Reproducibility Note}\label{app:reproducibility-note}
All experimental results in the main paper were generated using the raw empirical estimates, prior to introducing the regression fits in \Cref{sec:plan-detail}.
Since the planner in the original version is sensitive to timing noise, re-running those experiments may yield slightly different runtimes and occasionally flip the outcome on borderline instances near the timeout.
With the smoothed fits enabled, the planner deterministically selects the exact same plan across repeated runs in our tests.
The original version used to generate the reported numbers is tagged \texttt{v0.9.0} in our code repository. Later versions include the fits.

%% file: figures/fir-result6.tex
\resizebox{0.8\linewidth}{!}{
\begin{tabular}{cccccccccccccc}
\toprule
Dataset & Network & Neurons & Img & CC & $\epsilon$ & Cand. & R & T/O & R/Cand.(\%) & Time (s) & $l_s<\epsilon$ & $\branch^{\texttt{con}}$ & $l_s=\epsilon$ \\
\midrule
\multirow{37}{*}{MNIST} & \multirow{4}{*}{afc1} & \multirow{4}{*}{512} & \multirow{4}{*}{100} & \multirow{4}{*}{100} & 3 & 90 & 90 & 0 & 100.0\% & 0.3 & 90 & 21 & 0 \\
 &  &  &  &  & 4 & 72 & 70 & 2 & 97.2\% & 10.1 & 69 & 28 & 1 \\
 &  &  &  &  & 5 & 57 & 52 & 5 & 91.2\% & 17.8 & 52 & 30 & 0 \\
 &  &  &  &  & 6 & 46 & 29 & 17 & 63.0\% & 53.7 & 28 & 22 & 1 \\
\cmidrule(lr){2-14}
 & \multirow{4}{*}{afc2} & \multirow{4}{*}{1024} & \multirow{4}{*}{100} & \multirow{4}{*}{100} & 3 & 94 & 94 & 0 & 100.0\% & 0.3 & 94 & 14 & 0 \\
 &  &  &  &  & 4 & 88 & 87 & 1 & 98.9\% & 6.1 & 87 & 36 & 0 \\
 &  &  &  &  & 5 & 79 & 70 & 9 & 88.6\% & 48.0 & 70 & 45 & 0 \\
 &  &  &  &  & 6 & 68 & 38 & 30 & 55.9\% & 90.9 & 38 & 26 & 0 \\
\cmidrule(lr){2-14}
 & \multirow{4}{*}{afc3} & \multirow{4}{*}{1536} & \multirow{4}{*}{100} & \multirow{4}{*}{100} & 3 & 92 & 92 & 0 & 100.0\% & 0.5 & 92 & 11 & 0 \\
 &  &  &  &  & 4 & 84 & 84 & 0 & 100.0\% & 12.9 & 84 & 30 & 0 \\
 &  &  &  &  & 5 & 74 & 67 & 7 & 90.5\% & 34.9 & 67 & 43 & 0 \\
 &  &  &  &  & 6 & 66 & 42 & 24 & 63.6\% & 193.1 & 42 & 29 & 0 \\
\cmidrule(lr){2-14}
 & \multirow{4}{*}{afc4} & \multirow{4}{*}{384} & \multirow{4}{*}{100} & \multirow{4}{*}{99} & 3 & 89 & 89 & 0 & 100.0\% & <0.1 & 89 & 3 & 0 \\
 &  &  &  &  & 4 & 85 & 85 & 0 & 100.0\% & 0.5 & 85 & 20 & 0 \\
 &  &  &  &  & 5 & 73 & 70 & 3 & 95.9\% & 5.4 & 70 & 34 & 0 \\
 &  &  &  &  & 6 & 61 & 49 & 12 & 80.3\% & 53.5 & 49 & 29 & 0 \\
\cmidrule(lr){2-14}
 & \multirow{4}{*}{afc5} & \multirow{4}{*}{1000} & \multirow{4}{*}{100} & \multirow{4}{*}{100} & 3 & 89 & 89 & 0 & 100.0\% & 0.9 & 88 & 27 & 1 \\
 &  &  &  &  & 4 & 80 & 80 & 0 & 100.0\% & 30.3 & 62 & 35 & 18 \\
 &  &  &  &  & 5 & 70 & 61 & 9 & 87.1\% & 188.4 & 23 & 14 & 38 \\
 &  &  &  &  & 6 & 60 & 29 & 31 & 48.3\% & 374.8 & 6 & 3 & 23 \\
\cmidrule(lr){2-14}
 & \multirow{4}{*}{bfc1} & \multirow{4}{*}{1000} & \multirow{4}{*}{100} & \multirow{4}{*}{99} & 3 & 91 & 91 & 0 & 100.0\% & 4.7 & 52 & 31 & 39 \\
 &  &  &  &  & 4 & 82 & 81 & 1 & 98.8\% & 82.3 & 25 & 18 & 56 \\
 &  &  &  &  & 5 & 68 & 47 & 21 & 69.1\% & 192.4 & 8 & 7 & 39 \\
 &  &  &  &  & 6 & 59 & 13 & 46 & 22.0\% & 401.1 & 2 & 0 & 11 \\
\cmidrule(lr){2-14}
 & \multirow{4}{*}{bcv1} & \multirow{4}{*}{3604} & \multirow{4}{*}{100} & \multirow{4}{*}{98} & 3 & 87 & 87 & 0 & 100.0\% & 7.5 & 1 & 0 & 86 \\
 &  &  &  &  & 4 & 81 & 81 & 0 & 100.0\% & 13.6 & 1 & 0 & 80 \\
 &  &  &  &  & 5 & 73 & 69 & 4 & 94.5\% & 133.8 & 0 & 0 & 69 \\
 &  &  &  &  & 6 & 64 & 49 & 15 & 76.6\% & 359.1 & 0 & 0 & 49 \\
\cmidrule(lr){2-14}
 & \multirow{4}{*}{bcv2} & \multirow{4}{*}{3604} & \multirow{4}{*}{100} & \multirow{4}{*}{94} & 3 & 84 & 84 & 0 & 100.0\% & 8.7 & 4 & 2 & 80 \\
 &  &  &  &  & 4 & 74 & 74 & 0 & 100.0\% & 19.6 & 6 & 0 & 68 \\
 &  &  &  &  & 5 & 67 & 67 & 0 & 100.0\% & 49.4 & 1 & 0 & 66 \\
 &  &  &  &  & 6 & 62 & 55 & 7 & 88.7\% & 131.0 & 1 & 0 & 54 \\
\cmidrule(lr){2-14}
 & \multirow{4}{*}{bcv3} & \multirow{4}{*}{4504} & \multirow{4}{*}{100} & \multirow{4}{*}{99} & 3 & 90 & 90 & 0 & 100.0\% & 12.6 & 1 & 1 & 89 \\
 &  &  &  &  & 4 & 85 & 85 & 0 & 100.0\% & 25.2 & 1 & 0 & 84 \\
 &  &  &  &  & 5 & 76 & 72 & 4 & 94.7\% & 185.3 & 0 & 0 & 72 \\
 &  &  &  &  & 6 & 68 & 41 & 27 & 60.3\% & 401.9 & 0 & 0 & 41 \\
\cmidrule(lr){2-14}
 & \multirow{1}{*}{bcv4} & \multirow{1}{*}{34688} & \multirow{1}{*}{10} & \multirow{1}{*}{9} & 3 & 9 & 8 & 1 & 88.9\% & 793.3 & 0 & 0 & 8 \\
\midrule
\multirow{8}{*}{F-MNIST} & \multirow{4}{*}{bcv6} & \multirow{4}{*}{3604} & \multirow{4}{*}{100} & \multirow{4}{*}{83} & 3 & 66 & 66 & 0 & 100.0\% & 10.0 & 0 & 0 & 66 \\
 &  &  &  &  & 4 & 58 & 58 & 0 & 100.0\% & 57.8 & 0 & 0 & 58 \\
 &  &  &  &  & 5 & 49 & 34 & 15 & 69.4\% & 169.3 & 0 & 0 & 34 \\
 &  &  &  &  & 6 & 45 & 18 & 27 & 40.0\% & 329.1 & 0 & 0 & 18 \\
\cmidrule(lr){2-14}
 & \multirow{4}{*}{bcv7} & \multirow{4}{*}{4504} & \multirow{4}{*}{100} & \multirow{4}{*}{86} & 3 & 68 & 68 & 0 & 100.0\% & 16.6 & 0 & 0 & 68 \\
 &  &  &  &  & 4 & 60 & 54 & 6 & 90.0\% & 134.7 & 0 & 0 & 54 \\
 &  &  &  &  & 5 & 54 & 32 & 22 & 59.3\% & 304.8 & 0 & 0 & 32 \\
 &  &  &  &  & 6 & 47 & 4 & 43 & 8.5\% & 602.3 & 0 & 0 & 4 \\
\midrule
\multirow{1}{*}{CIFAR-10} & \multirow{1}{*}{bcv5} & \multirow{1}{*}{4852} & \multirow{1}{*}{10} & \multirow{1}{*}{9} & 3 & 5 & 4 & 1 & 80.0\% & 302.0 & 0 & 0 & 4 \\
\bottomrule
\end{tabular}

}

%% file: svg-inkscape/All_e3_svg-tex.pdf_tex
\begingroup%
  \makeatletter%
  \providecommand\color[2][]{%
    \errmessage{(Inkscape) Color is used for the text in Inkscape, but the package 'color.sty' is not loaded}%
    \renewcommand\color[2][]{}%
  }%
  \providecommand\transparent[1]{%
    \errmessage{(Inkscape) Transparency is used (non-zero) for the text in Inkscape, but the package 'transparent.sty' is not loaded}%
    \renewcommand\transparent[1]{}%
  }%
  \providecommand\rotatebox[2]{#2}%
  \newcommand*\fsize{\dimexpr\f@size pt\relax}%
  \newcommand*\lineheight[1]{\fontsize{\fsize}{#1\fsize}\selectfont}%
  \ifx\svgwidth\undefined%
    \setlength{\unitlength}{720bp}%
    \ifx\svgscale\undefined%
      \relax%
    \else%
      \setlength{\unitlength}{\unitlength * \real{\svgscale}}%
    \fi%
  \else%
    \setlength{\unitlength}{\svgwidth}%
  \fi%
  \global\let\svgwidth\undefined%
  \global\let\svgscale\undefined%
  \makeatother%
  \begin{picture}(1,0.6)%
    \lineheight{1}%
    \setlength\tabcolsep{0pt}%
    \put(0,0){\includegraphics[width=\unitlength,page=1]{svg-inkscape/All_e3_svg-tex.pdf}}%
  \end{picture}%
\endgroup%

%% file: svg-inkscape/Fully-Connected_e3_svg-tex.pdf_tex
\begingroup%
  \makeatletter%
  \providecommand\color[2][]{%
    \errmessage{(Inkscape) Color is used for the text in Inkscape, but the package 'color.sty' is not loaded}%
    \renewcommand\color[2][]{}%
  }%
  \providecommand\transparent[1]{%
    \errmessage{(Inkscape) Transparency is used (non-zero) for the text in Inkscape, but the package 'transparent.sty' is not loaded}%
    \renewcommand\transparent[1]{}%
  }%
  \providecommand\rotatebox[2]{#2}%
  \newcommand*\fsize{\dimexpr\f@size pt\relax}%
  \newcommand*\lineheight[1]{\fontsize{\fsize}{#1\fsize}\selectfont}%
  \ifx\svgwidth\undefined%
    \setlength{\unitlength}{720bp}%
    \ifx\svgscale\undefined%
      \relax%
    \else%
      \setlength{\unitlength}{\unitlength * \real{\svgscale}}%
    \fi%
  \else%
    \setlength{\unitlength}{\svgwidth}%
  \fi%
  \global\let\svgwidth\undefined%
  \global\let\svgscale\undefined%
  \makeatother%
  \begin{picture}(1,0.6)%
    \lineheight{1}%
    \setlength\tabcolsep{0pt}%
    \put(0,0){\includegraphics[width=\unitlength,page=1]{svg-inkscape/Fully-Connected_e3_svg-tex.pdf}}%
  \end{picture}%
\endgroup%

%% file: svg-inkscape/Convolutional_e3_svg-tex.pdf_tex
\begingroup%
  \makeatletter%
  \providecommand\color[2][]{%
    \errmessage{(Inkscape) Color is used for the text in Inkscape, but the package 'color.sty' is not loaded}%
    \renewcommand\color[2][]{}%
  }%
  \providecommand\transparent[1]{%
    \errmessage{(Inkscape) Transparency is used (non-zero) for the text in Inkscape, but the package 'transparent.sty' is not loaded}%
    \renewcommand\transparent[1]{}%
  }%
  \providecommand\rotatebox[2]{#2}%
  \newcommand*\fsize{\dimexpr\f@size pt\relax}%
  \newcommand*\lineheight[1]{\fontsize{\fsize}{#1\fsize}\selectfont}%
  \ifx\svgwidth\undefined%
    \setlength{\unitlength}{720bp}%
    \ifx\svgscale\undefined%
      \relax%
    \else%
      \setlength{\unitlength}{\unitlength * \real{\svgscale}}%
    \fi%
  \else%
    \setlength{\unitlength}{\svgwidth}%
  \fi%
  \global\let\svgwidth\undefined%
  \global\let\svgscale\undefined%
  \makeatother%
  \begin{picture}(1,0.6)%
    \lineheight{1}%
    \setlength\tabcolsep{0pt}%
    \put(0,0){\includegraphics[width=\unitlength,page=1]{svg-inkscape/Convolutional_e3_svg-tex.pdf}}%
  \end{picture}%
\endgroup%

%% file: svg-inkscape/All_e4_svg-tex.pdf_tex
\begingroup%
  \makeatletter%
  \providecommand\color[2][]{%
    \errmessage{(Inkscape) Color is used for the text in Inkscape, but the package 'color.sty' is not loaded}%
    \renewcommand\color[2][]{}%
  }%
  \providecommand\transparent[1]{%
    \errmessage{(Inkscape) Transparency is used (non-zero) for the text in Inkscape, but the package 'transparent.sty' is not loaded}%
    \renewcommand\transparent[1]{}%
  }%
  \providecommand\rotatebox[2]{#2}%
  \newcommand*\fsize{\dimexpr\f@size pt\relax}%
  \newcommand*\lineheight[1]{\fontsize{\fsize}{#1\fsize}\selectfont}%
  \ifx\svgwidth\undefined%
    \setlength{\unitlength}{720bp}%
    \ifx\svgscale\undefined%
      \relax%
    \else%
      \setlength{\unitlength}{\unitlength * \real{\svgscale}}%
    \fi%
  \else%
    \setlength{\unitlength}{\svgwidth}%
  \fi%
  \global\let\svgwidth\undefined%
  \global\let\svgscale\undefined%
  \makeatother%
  \begin{picture}(1,0.6)%
    \lineheight{1}%
    \setlength\tabcolsep{0pt}%
    \put(0,0){\includegraphics[width=\unitlength,page=1]{svg-inkscape/All_e4_svg-tex.pdf}}%
  \end{picture}%
\endgroup%

%% file: svg-inkscape/Fully-Connected_e4_svg-tex.pdf_tex
\begingroup%
  \makeatletter%
  \providecommand\color[2][]{%
    \errmessage{(Inkscape) Color is used for the text in Inkscape, but the package 'color.sty' is not loaded}%
    \renewcommand\color[2][]{}%
  }%
  \providecommand\transparent[1]{%
    \errmessage{(Inkscape) Transparency is used (non-zero) for the text in Inkscape, but the package 'transparent.sty' is not loaded}%
    \renewcommand\transparent[1]{}%
  }%
  \providecommand\rotatebox[2]{#2}%
  \newcommand*\fsize{\dimexpr\f@size pt\relax}%
  \newcommand*\lineheight[1]{\fontsize{\fsize}{#1\fsize}\selectfont}%
  \ifx\svgwidth\undefined%
    \setlength{\unitlength}{720bp}%
    \ifx\svgscale\undefined%
      \relax%
    \else%
      \setlength{\unitlength}{\unitlength * \real{\svgscale}}%
    \fi%
  \else%
    \setlength{\unitlength}{\svgwidth}%
  \fi%
  \global\let\svgwidth\undefined%
  \global\let\svgscale\undefined%
  \makeatother%
  \begin{picture}(1,0.6)%
    \lineheight{1}%
    \setlength\tabcolsep{0pt}%
    \put(0,0){\includegraphics[width=\unitlength,page=1]{svg-inkscape/Fully-Connected_e4_svg-tex.pdf}}%
  \end{picture}%
\endgroup%

%% file: svg-inkscape/Convolutional_e4_svg-tex.pdf_tex
\begingroup%
  \makeatletter%
  \providecommand\color[2][]{%
    \errmessage{(Inkscape) Color is used for the text in Inkscape, but the package 'color.sty' is not loaded}%
    \renewcommand\color[2][]{}%
  }%
  \providecommand\transparent[1]{%
    \errmessage{(Inkscape) Transparency is used (non-zero) for the text in Inkscape, but the package 'transparent.sty' is not loaded}%
    \renewcommand\transparent[1]{}%
  }%
  \providecommand\rotatebox[2]{#2}%
  \newcommand*\fsize{\dimexpr\f@size pt\relax}%
  \newcommand*\lineheight[1]{\fontsize{\fsize}{#1\fsize}\selectfont}%
  \ifx\svgwidth\undefined%
    \setlength{\unitlength}{720bp}%
    \ifx\svgscale\undefined%
      \relax%
    \else%
      \setlength{\unitlength}{\unitlength * \real{\svgscale}}%
    \fi%
  \else%
    \setlength{\unitlength}{\svgwidth}%
  \fi%
  \global\let\svgwidth\undefined%
  \global\let\svgscale\undefined%
  \makeatother%
  \begin{picture}(1,0.6)%
    \lineheight{1}%
    \setlength\tabcolsep{0pt}%
    \put(0,0){\includegraphics[width=\unitlength,page=1]{svg-inkscape/Convolutional_e4_svg-tex.pdf}}%
  \end{picture}%
\endgroup%

%% file: svg-inkscape/All_e6_svg-tex.pdf_tex
\begingroup%
  \makeatletter%
  \providecommand\color[2][]{%
    \errmessage{(Inkscape) Color is used for the text in Inkscape, but the package 'color.sty' is not loaded}%
    \renewcommand\color[2][]{}%
  }%
  \providecommand\transparent[1]{%
    \errmessage{(Inkscape) Transparency is used (non-zero) for the text in Inkscape, but the package 'transparent.sty' is not loaded}%
    \renewcommand\transparent[1]{}%
  }%
  \providecommand\rotatebox[2]{#2}%
  \newcommand*\fsize{\dimexpr\f@size pt\relax}%
  \newcommand*\lineheight[1]{\fontsize{\fsize}{#1\fsize}\selectfont}%
  \ifx\svgwidth\undefined%
    \setlength{\unitlength}{720bp}%
    \ifx\svgscale\undefined%
      \relax%
    \else%
      \setlength{\unitlength}{\unitlength * \real{\svgscale}}%
    \fi%
  \else%
    \setlength{\unitlength}{\svgwidth}%
  \fi%
  \global\let\svgwidth\undefined%
  \global\let\svgscale\undefined%
  \makeatother%
  \begin{picture}(1,0.6)%
    \lineheight{1}%
    \setlength\tabcolsep{0pt}%
    \put(0,0){\includegraphics[width=\unitlength,page=1]{svg-inkscape/All_e6_svg-tex.pdf}}%
  \end{picture}%
\endgroup%

%% file: svg-inkscape/Fully-Connected_e6_svg-tex.pdf_tex
\begingroup%
  \makeatletter%
  \providecommand\color[2][]{%
    \errmessage{(Inkscape) Color is used for the text in Inkscape, but the package 'color.sty' is not loaded}%
    \renewcommand\color[2][]{}%
  }%
  \providecommand\transparent[1]{%
    \errmessage{(Inkscape) Transparency is used (non-zero) for the text in Inkscape, but the package 'transparent.sty' is not loaded}%
    \renewcommand\transparent[1]{}%
  }%
  \providecommand\rotatebox[2]{#2}%
  \newcommand*\fsize{\dimexpr\f@size pt\relax}%
  \newcommand*\lineheight[1]{\fontsize{\fsize}{#1\fsize}\selectfont}%
  \ifx\svgwidth\undefined%
    \setlength{\unitlength}{720bp}%
    \ifx\svgscale\undefined%
      \relax%
    \else%
      \setlength{\unitlength}{\unitlength * \real{\svgscale}}%
    \fi%
  \else%
    \setlength{\unitlength}{\svgwidth}%
  \fi%
  \global\let\svgwidth\undefined%
  \global\let\svgscale\undefined%
  \makeatother%
  \begin{picture}(1,0.6)%
    \lineheight{1}%
    \setlength\tabcolsep{0pt}%
    \put(0,0){\includegraphics[width=\unitlength,page=1]{svg-inkscape/Fully-Connected_e6_svg-tex.pdf}}%
  \end{picture}%
\endgroup%

%% file: svg-inkscape/Convolutional_e6_svg-tex.pdf_tex
\begingroup%
  \makeatletter%
  \providecommand\color[2][]{%
    \errmessage{(Inkscape) Color is used for the text in Inkscape, but the package 'color.sty' is not loaded}%
    \renewcommand\color[2][]{}%
  }%
  \providecommand\transparent[1]{%
    \errmessage{(Inkscape) Transparency is used (non-zero) for the text in Inkscape, but the package 'transparent.sty' is not loaded}%
    \renewcommand\transparent[1]{}%
  }%
  \providecommand\rotatebox[2]{#2}%
  \newcommand*\fsize{\dimexpr\f@size pt\relax}%
  \newcommand*\lineheight[1]{\fontsize{\fsize}{#1\fsize}\selectfont}%
  \ifx\svgwidth\undefined%
    \setlength{\unitlength}{720bp}%
    \ifx\svgscale\undefined%
      \relax%
    \else%
      \setlength{\unitlength}{\unitlength * \real{\svgscale}}%
    \fi%
  \else%
    \setlength{\unitlength}{\svgwidth}%
  \fi%
  \global\let\svgwidth\undefined%
  \global\let\svgscale\undefined%
  \makeatother%
  \begin{picture}(1,0.6)%
    \lineheight{1}%
    \setlength\tabcolsep{0pt}%
    \put(0,0){\includegraphics[width=\unitlength,page=1]{svg-inkscape/Convolutional_e6_svg-tex.pdf}}%
  \end{picture}%
\endgroup%

%% file: figures/bcv45.tex
\begin{table}[t]
\centering
\caption{Sample-wise comparison between \tool and \cvdtt on bcv4 and bcv5. Those incorrectly classified images and images containing adversarial examples within the perturbation budget are marked ``-'' in the table. The timeout for bcv4 is 3600s and for bcv5 is 1800s. $^{*}$As we failed to run \cvdtt on bcv5 with 50 MILP instances on our machine, this row is taken from the results table in~\cite{shapira2026tight} directly, and thus it is not directly comparable.}
\label{tab:bcv45}
\small
\setlength{\tabcolsep}{4pt}
\vspace{5pt}
\resizebox{.9\textwidth}{!}{
\begin{tabular}{ccccccccccccc}
\toprule
Dataset                   & Network               & Method                & 0 & 1 & 2 & 3 & 4 & 5 & 6 & 7 & 8 & 9 \\
\midrule
\multirow{2}{*}{MNIST}    & \multirow{2}{*}{bcv4} & \cvdtt & 1286.4  & 1008.5  & \texttt{T/O}  & 1002.5  & 1397.5  & \texttt{T/O}  & 1248.2  & 905.5  & -  & 784.9  \\ \cline{3-13}
                          &                       & \tool  & 554.2  & 448.3  & \texttt{T/O}  & 367.6  & 611.0  & 3145.6  & 557.1  & 363.3  & -  & 299.0  \\
\midrule
\multirow{2}{*}{CIFAR-10} & \multirow{2}{*}{bcv5} & \cvdtt{}$^{*}$ & 1608.0  & 1540.0  & 1698.0  & -  & -  & \texttt{T/O}  & -  & \texttt{T/O}  & -  & -  \\ \cline{3-13}
                          &                       & \tool  & 132.3  & 133.2  & 141.6  & -  & -  & 801.1  & -  & \texttt{T/O}  & -  & -  \\
\bottomrule
\end{tabular}
}
\vspace{5pt}
\end{table}

%% file: svg-inkscape/Ablation_bafc1_samplewise_e5_svg-tex.pdf_tex
\begingroup%
  \makeatletter%
  \providecommand\color[2][]{%
    \errmessage{(Inkscape) Color is used for the text in Inkscape, but the package 'color.sty' is not loaded}%
    \renewcommand\color[2][]{}%
  }%
  \providecommand\transparent[1]{%
    \errmessage{(Inkscape) Transparency is used (non-zero) for the text in Inkscape, but the package 'transparent.sty' is not loaded}%
    \renewcommand\transparent[1]{}%
  }%
  \providecommand\rotatebox[2]{#2}%
  \newcommand*\fsize{\dimexpr\f@size pt\relax}%
  \newcommand*\lineheight[1]{\fontsize{\fsize}{#1\fsize}\selectfont}%
  \ifx\svgwidth\undefined%
    \setlength{\unitlength}{1317.6bp}%
    \ifx\svgscale\undefined%
      \relax%
    \else%
      \setlength{\unitlength}{\unitlength * \real{\svgscale}}%
    \fi%
  \else%
    \setlength{\unitlength}{\svgwidth}%
  \fi%
  \global\let\svgwidth\undefined%
  \global\let\svgscale\undefined%
  \makeatother%
  \begin{picture}(1,0.32786885)%
    \lineheight{1}%
    \setlength\tabcolsep{0pt}%
    \put(0,0){\includegraphics[width=\unitlength,page=1]{svg-inkscape/Ablation_bafc1_samplewise_e5_svg-tex.pdf}}%
  \end{picture}%
\endgroup%

%% file: svg-inkscape/Ablation_e5_svg-tex.pdf_tex
\begingroup%
  \makeatletter%
  \providecommand\color[2][]{%
    \errmessage{(Inkscape) Color is used for the text in Inkscape, but the package 'color.sty' is not loaded}%
    \renewcommand\color[2][]{}%
  }%
  \providecommand\transparent[1]{%
    \errmessage{(Inkscape) Transparency is used (non-zero) for the text in Inkscape, but the package 'transparent.sty' is not loaded}%
    \renewcommand\transparent[1]{}%
  }%
  \providecommand\rotatebox[2]{#2}%
  \newcommand*\fsize{\dimexpr\f@size pt\relax}%
  \newcommand*\lineheight[1]{\fontsize{\fsize}{#1\fsize}\selectfont}%
  \ifx\svgwidth\undefined%
    \setlength{\unitlength}{720bp}%
    \ifx\svgscale\undefined%
      \relax%
    \else%
      \setlength{\unitlength}{\unitlength * \real{\svgscale}}%
    \fi%
  \else%
    \setlength{\unitlength}{\svgwidth}%
  \fi%
  \global\let\svgwidth\undefined%
  \global\let\svgscale\undefined%
  \makeatother%
  \begin{picture}(1,0.6)%
    \lineheight{1}%
    \setlength\tabcolsep{0pt}%
    \put(0,0){\includegraphics[width=\unitlength,page=1]{svg-inkscape/Ablation_e5_svg-tex.pdf}}%
  \end{picture}%
\endgroup%

%% file: svg-inkscape/Ablation_e6_svg-tex.pdf_tex
\begingroup%
  \makeatletter%
  \providecommand\color[2][]{%
    \errmessage{(Inkscape) Color is used for the text in Inkscape, but the package 'color.sty' is not loaded}%
    \renewcommand\color[2][]{}%
  }%
  \providecommand\transparent[1]{%
    \errmessage{(Inkscape) Transparency is used (non-zero) for the text in Inkscape, but the package 'transparent.sty' is not loaded}%
    \renewcommand\transparent[1]{}%
  }%
  \providecommand\rotatebox[2]{#2}%
  \newcommand*\fsize{\dimexpr\f@size pt\relax}%
  \newcommand*\lineheight[1]{\fontsize{\fsize}{#1\fsize}\selectfont}%
  \ifx\svgwidth\undefined%
    \setlength{\unitlength}{720bp}%
    \ifx\svgscale\undefined%
      \relax%
    \else%
      \setlength{\unitlength}{\unitlength * \real{\svgscale}}%
    \fi%
  \else%
    \setlength{\unitlength}{\svgwidth}%
  \fi%
  \global\let\svgwidth\undefined%
  \global\let\svgscale\undefined%
  \makeatother%
  \begin{picture}(1,0.6)%
    \lineheight{1}%
    \setlength\tabcolsep{0pt}%
    \put(0,0){\includegraphics[width=\unitlength,page=1]{svg-inkscape/Ablation_e6_svg-tex.pdf}}%
  \end{picture}%
\endgroup%

%% file: figures/b1results.tex
\begin{table}[t]
\centering
\caption{Main results without parallelization on a single GPU at $\epsilon \in \{5, 6\}$, except for bcv4 and bcv5 at $\epsilon = 3$. Neurons is the number of neurons in the network, Img is the number of evaluated images, and CC is the number of correctly classified images. Cand.\ denotes the number of verification candidate instances at the target budget $\epsilon$, after excluding images for which an adversarial example within the budget is already known (from any run, including the 4-GPU main results, so Cand.\ is the same as there). R is the number of verified images, R/Cand is the verified rate, and Time reports the mean verification time (in seconds) over each tool's verified instances. Speedup is how many times faster \tool is than \cvdtt (geometric mean over the instances verified by both tools).}
\label{tab:main_results_b1}
\resizebox{0.8\linewidth}{!}{
\begin{tabular}{cccccccccccccc}
\toprule
\multirow{2}{*}{\textbf{Dataset}} &
\multirow{2}{*}{\textbf{Network}} &
\multirow{2}{*}{\textbf{Neurons}} &
\multirow{2}{*}{\textbf{Img}} &
\multirow{2}{*}{\textbf{CC}} &
\multirow{2}{*}{{\bf $\epsilon$}} &
\multirow{2}{*}{\textbf{Cand.}} &
\multicolumn{3}{c}{\textbf{\tool}} &
\multicolumn{3}{c}{\textbf{\cvdtt}} &
\multirow{2}{*}{\textbf{Speedup}$\uparrow$} \\
\cmidrule(lr){8-10}\cmidrule(lr){11-13}
 &  &  &  &  &  &  & $R\uparrow$ & R/Cand.$\uparrow$ & Time (s)$\downarrow$ & $R\uparrow$ & R/Cand.$\uparrow$ & Time (s)$\downarrow$ &  \\
\midrule
\multirow{19}{*}{MNIST} & \multirow{2}{*}{afc1} & \multirow{2}{*}{512} & \multirow{2}{*}{100} & \multirow{2}{*}{100} & 5 & 57 & 38 & 66.7\% & 275.0 & 26 & 45.6\% & 116.4 & 4.7$\times$ \\
 &  &  &  &  & 6 & 46 & 20 & 43.5\% & 136.8 & 10 & 21.7\% & 378.0 & 5.1$\times$ \\
\cmidrule(lr){2-14}
 & \multirow{2}{*}{afc2} & \multirow{2}{*}{1024} & \multirow{2}{*}{100} & \multirow{2}{*}{100} & 5 & 79 & 37 & 46.8\% & 124.2 & 28 & 35.4\% & 264.9 & 5.7$\times$ \\
 &  &  &  &  & 6 & 68 & 20 & 29.4\% & 211.8 & 13 & 19.1\% & 649.0 & 16.7$\times$ \\
\cmidrule(lr){2-14}
 & \multirow{2}{*}{afc3} & \multirow{2}{*}{1536} & \multirow{2}{*}{100} & \multirow{2}{*}{100} & 5 & 74 & 39 & 52.7\% & 103.5 & 32 & 43.2\% & 475.8 & 5.4$\times$ \\
 &  &  &  &  & 6 & 66 & 19 & 28.8\% & 124.8 & 11 & 16.7\% & 131.6 & 1.3$\times$ \\
\cmidrule(lr){2-14}
 & \multirow{2}{*}{afc4} & \multirow{2}{*}{384} & \multirow{2}{*}{100} & \multirow{2}{*}{99} & 5 & 73 & 61 & 83.6\% & 119.3 & 35 & 47.9\% & 56.3 & 6.3$\times$ \\
 &  &  &  &  & 6 & 61 & 34 & 55.7\% & 299.7 & 19 & 31.1\% & 124.7 & 6.7$\times$ \\
\cmidrule(lr){2-14}
 & \multirow{2}{*}{afc5} & \multirow{2}{*}{1000} & \multirow{2}{*}{100} & \multirow{2}{*}{100} & 5 & 70 & 15 & 21.4\% & 458.0 & 18 & 25.7\% & 1033.5 & 5.3$\times$ \\
 &  &  &  &  & 6 & 60 & 3 & 5.0\% & 313.5 & 1 & 1.7\% & $<$0.1 & 3.2$\times$ \\
\cmidrule(lr){2-14}
 & \multirow{2}{*}{bfc1} & \multirow{2}{*}{1000} & \multirow{2}{*}{100} & \multirow{2}{*}{99} & 5 & 68 & 4 & 5.9\% & 602.0 & 4 & 5.9\% & 1206.0 & 2.5$\times$ \\
 &  &  &  &  & 6 & 59 & 2 & 3.4\% & 1093.5 & 0 & 0.0\% & -- & -- \\
\cmidrule(lr){2-14}
 & \multirow{2}{*}{bcv1} & \multirow{2}{*}{3604} & \multirow{2}{*}{100} & \multirow{2}{*}{98} & 5 & 73 & 29 & 39.7\% & 854.3 & 32 & 43.8\% & 766.1 & 1.0$\times$ \\
 &  &  &  &  & 6 & 64 & 6 & 9.4\% & 1251.9 & 6 & 9.4\% & 1016.2 & 0.8$\times$ \\
\cmidrule(lr){2-14}
 & \multirow{2}{*}{bcv2} & \multirow{2}{*}{3604} & \multirow{2}{*}{100} & \multirow{2}{*}{94} & 5 & 67 & 44 & 65.7\% & 493.6 & 56 & 83.6\% & 686.0 & 1.2$\times$ \\
 &  &  &  &  & 6 & 62 & 17 & 27.4\% & 1140.6 & 22 & 35.5\% & 1070.8 & 0.8$\times$ \\
\cmidrule(lr){2-14}
 & \multirow{2}{*}{bcv3} & \multirow{2}{*}{4504} & \multirow{2}{*}{100} & \multirow{2}{*}{99} & 5 & 76 & 25 & 32.9\% & 873.3 & 17 & 22.4\% & 1004.5 & 1.9$\times$ \\
 &  &  &  &  & 6 & 68 & 3 & 4.4\% & 1149.0 & 3 & 4.4\% & 1729.9 & 1.5$\times$ \\
\cmidrule(lr){2-14}
 & \multirow{1}{*}{bcv4} & \multirow{1}{*}{34688} & \multirow{1}{*}{10} & \multirow{1}{*}{9} & 3 & 9 & 7 & 77.8\% & 1446.4 & 2 & 22.2\% & 3245.1 & 4.1$\times$ \\
\midrule
\multirow{4}{*}{F-MNIST} & \multirow{2}{*}{bcv6} & \multirow{2}{*}{3604} & \multirow{2}{*}{100} & \multirow{2}{*}{83} & 5 & 49 & 7 & 14.3\% & 1098.5 & 14 & 28.6\% & 1196.7 & 1.0$\times$ \\
 &  &  &  &  & 6 & 45 & 1 & 2.2\% & 1199.2 & 2 & 4.4\% & 1246.6 & 0.8$\times$ \\
\cmidrule(lr){2-14}
 & \multirow{2}{*}{bcv7} & \multirow{2}{*}{4504} & \multirow{2}{*}{100} & \multirow{2}{*}{86} & 5 & 54 & 1 & 1.9\% & 722.7 & 1 & 1.9\% & 1013.1 & 1.4$\times$ \\
 &  &  &  &  & 6 & 47 & 0 & 0.0\% & -- & 0 & 0.0\% & -- & -- \\
\midrule
\multirow{1}{*}{CIFAR-10} & \multirow{1}{*}{bcv5} & \multirow{1}{*}{4852} & \multirow{1}{*}{10} & \multirow{1}{*}{9} & 3 & 5 & 0 & 0.0\% & -- & 0 & 0.0\% & -- & -- \\
\bottomrule
\end{tabular}
}
\end{table}